\documentclass[10pt,journal,compsoc]{IEEEtran}
\usepackage{graphicx}
\usepackage{subcaption}
\usepackage{url}

\usepackage{acronym}
\acrodef{AAL}{Ambient Assisted Living}
\acrodef{CHALET}{Cornell House Agent Learning Environment}
\acrodef{CNN}{Convolutional Neural Network}
\acrodef{FoV}{Field of View}
\acrodef{HoME}{Household Multimodal Environment}
\acrodef{HUD}{Head-Up Display}
\acrodef{MINOS}{Multimodal Indoor Simulator}
\acrodef{MSAA}{Multi-Sample Anti-Aliasing}
\acrodef{UE4}{Unreal Engine 4}
\acrodef{URDF}{Unified Robot Description File}
\acrodef{VR}{Virtual Reality}
\acrodef{THOR}{THe House of inteRactions}

\ifCLASSOPTIONcompsoc
  \usepackage[nocompress]{cite}
\else
  \usepackage{cite}
\fi
\ifCLASSINFOpdf
\else
\fi
\begin{document}
\title{UnrealROX\\ An eXtremely Photorealistic Virtual Reality Environment for Robotics Simulations and Synthetic Data Generation}
\author{Pablo~Martinez-Gonzalez,
        Sergiu~Oprea,
        Alberto~Garcia-Garcia,
				Alvaro~Jover-Alvarez,
				Sergio~Orts-Escolano,
				and~Jose~Garcia-Rodriguez% <-this % stops a space
\IEEEcompsocitemizethanks{\IEEEcompsocthanksitem P. Martinez-Gonzalez, S. Oprea, A. Garcia-Garcia, A. Jover-Alvarez, S. Orts-Escolano, and J. Garcia-Rodriguez are with the 3D Perception Lab (http://labs.iuii.ua.es/3dperceptionlab) at the University of Alicante .\protect\\
E-mail: pmartinez@dtic.ua.es, soprea@dtic.ua.es, agarcia@dtic.ua.es, ajover@dtic.ua.es, sorts@ua.es, jgarcia@dtic.ua.es}% <-this % stops an unwanted space
\thanks{}}


\markboth{}%
{}



\IEEEtitleabstractindextext{%
\begin{abstract}
Data-driven algorithms have surpassed traditional techniques in almost every aspect in robotic vision problems. Such algorithms need vast amounts of quality data to be able to work properly after their training process. Gathering and annotating that sheer amount of data in the real world is a time-consuming and error-prone task. Those problems limit scale and quality. Synthetic data generation has become increasingly popular since it is faster to generate and automatic to annotate. However, most of the current datasets and environments lack realism, interactions, and details from the real world. UnrealROX is an environment built over Unreal Engine 4 which aims to reduce that reality gap by leveraging hyperrealistic indoor scenes that are explored by robot agents which also interact with objects in a visually realistic manner in that simulated world. Photorealistic scenes and robots are rendered by Unreal Engine into a virtual reality headset which captures gaze so that a human operator can move the robot and use controllers for the robotic hands; scene information is dumped on a per-frame basis so that it can be reproduced offline to generate raw data and ground truth annotations. This virtual reality environment enables robotic vision researchers to generate realistic and visually plausible data with full ground truth for a wide variety of problems such as class and instance semantic segmentation, object detection, depth estimation, visual grasping, and navigation.
\end{abstract}

}

\maketitle

\IEEEdisplaynontitleabstractindextext

% \fi
\IEEEpeerreviewmaketitle

\IEEEraisesectionheading{\section{Introduction}\label{sec:introduction}}

\IEEEPARstart{V}{ision-based} robotics tasks have made a huge leap forward mainly due to the development of machine learning techniques (e.g. deep architectures \cite{LeCun2015} such as Convolutional Neural Networks or Recurrent Neural Networks) which are continuously rising the performance bar for various problems such as semantic segmentation \cite{Long2015}\cite{He2018}, depth estimation \cite{Eigen2014}\cite{Ummenhofer2017}, and visual grasping \cite{Lenz2015}\cite{Levine2018} among others. Those data-driven methods are in need of vast amounts of annotated samples to achieve those exceptional results. Gathering that sheer quantity of images with ground truth is a tedious, expensive, and sometimes nearby impossible task in the real world. On the contrary, synthetic environments streamline the data generation process and are usually able to automatically provide annotations for various tasks. Because of that, simulated environments are becoming increasingly popular and widely used to train those models.

Learning on virtual or simulated worlds allows faster, low-cost, and more scalable data collection. However, synthetic environments face a huge obstacle to be actually useful despite their inherent advantages: models trained in that simulated domain must also be able to perform properly on real-world test scenarios which often feature numerous discrepancies between them and their synthetic counterparts. That set of differences is widely known as the reality gap. In most cases, this gap is big enough so that transferring knowledge from one domain to another is an extremely difficult task either because renderers are not able to produce images like real-world sensors (due to the implicit noise or the richness of the scene) or either the physical behavior of the scene elements and sensors is not as accurate as it should be.

In order to address this reality gap, two methods have been proven to be effective: extreme realism and domain randomization. On the one hand, extreme realism refers to the process of making the simulation as similar as the real-world environment in which the robot will be deployed as possible \cite{McCormac2016}\cite{Gaidon2016}. That can be achieved through a combination of various techniques, e.g., photorealistic rendering (which implies realistic geometry, textures, lighting and also simulating camera-specific noise, distortion and other parameters) and accurate physics (complex collisions with high-fidelity calculations). On the other hand, domain randomization is a kind of domain adaptation technique that aims for exposing the model to a huge range of simulated environments at training time instead of just to a single synthetic one \cite{Bousmalis2017}\cite{Tobin2017}\cite{Tobin2017a}. By doing that, and if the variability is enough, the model will be able to identify the real world as just another variation thus being able to generalize \cite{Tremblay2018}.

In this work, we propose an extremely photorealistic virtual reality environment for generating synthetic data for various robotic vision tasks. In such environment, a human operator can be embodied, in virtual reality, as a robot agent inside a scene to freely navigate and interact with objects as if it was a real-world robot.  Our environment is built on top of \ac{UE4} to take advantage of its advanced \ac{VR}, rendering, and physics capabilities. Our system provides the following features: (1) a visually plausible grasping system for robot manipulation which is modular enough to be applied to various finger configurations, (2) routines for controlling robotic hands and bodies with commercial \ac{VR} setups such as Oculus Rift and HTC Vive Pro, (3) a sequence recorder component to store all the information about the scene, robot, and cameras while the human operator is embodied as a robot, (4) a sequence playback component to reproduce the previously recorded sequence offline to generate raw data such as RGB, depth, normals, or instance segmentation images, (5) a multi-camera component to ease the camera placement process and enable the user to attach them to specific robot joints and configure their parameters (resolution, noise model, field of view), and (6) open-source code, assets, and tutorials for all those components and other subsystems that tie them together.

This paper is organized as follows. Section \ref{sec:related_works} analyzes already existing environments for synthetic data generation and puts our proposal in context. Next, Section \ref{sec:system} describes our proposal and provides in-depth details for each one of its components. After that, we briefly discuss application scenarios for our environment in Section \ref{sec:applications}. At last, in Section \ref{sec:conclusion}, we draw conclusions about this work and in Section \ref{sec:limitations} we go over current limitations of our work and propose future works to improve it.

\section{Related Works}
\label{sec:related_works}

Synthetic environments have been used for a long time to benchmark vision and robotic algorithms \cite{Butler2012}. Recently, their importance has been highlighted for training and evaluating machine learning models for robotic vision problems \cite{Brodeur2017} \cite{Ros2016} \cite{Mahler2017}. Due to the increasing need for samples to train such data-driven architectures, there exists an increasing number of synthetic datasets, environments, and simulation platforms to generate data for indoor robotic tasks and evaluate those learned models. In this section, we briefly review the most relevant ones according to the scope of our proposal. We describe both the most important features and main flaws for the following works: \acs{CHALET}, \acs{HoME}, AI2-\acs{THOR}, and \acs{MINOS}. In addition, we also describe two other related tools such as UnrealCV, Gazebo, and NVIDIA's Isaac Sim which are not strictly similar but relevant enough to be mentioned. At last, we put our proposal in context taking into account all the analyzed strong points and weaknesses.

\ac{CHALET} \cite{Yan2018} is a 3D house simulator for manipulation and navigation learning. It is built upon Unity 3D so it supports physics and interactions with objects and the scene itself thanks to its built-in physics engine. CHALET features three modes of operation: standalone (navigate with keyboard and mouse input), replay (reproduce the trajectory generated on standalone mode), and client (use the framework's API to control the agent and obtain information). On the other hand, CHALET presents various weak points such as its lack of realism, the absence of a robot's body or mesh, and the limitation in the number of cameras.

\ac{HoME} \cite{Brodeur2017} is a multimodal household environment for AI learning from visual, auditive, and physical information within realistic synthetic environments sourced from SUNCG. HoME provides RGB, depth, and semantic maps based on 3D renderings produced by Panda3D, acoustic renderings based on EVERT, language descriptions of objects, and physics simulations based on Bullet. It also provides a Python framework compatible with OpenAI gym. However, HoME is not anywhere close to photorealism, there is no phyisical representation of the robot itself, and interactions are discrete.

AI2-\ac{THOR} \cite{Kolve2017} is a framework for visual AI research which consists of near-photorealistic synthetic 3D indoor scenes in which agents can navigate and change the state of actionable objects. It is built over Unity so it also integrates a physics engine which enables modeling complex physical interactions. The framework also provides a Python interface to communicate with the engine through HTTP commands to control the agent and obtain visual information and annotations. Some of the weaknesses of this environment are the lack of a 3D robot model and hands, only a first-person view camera, and the discrete nature of its actions with binary states.

\ac{MINOS} \cite{Savva2017} is a simulator for navigation in complex indoor environments. An agent, represented by a cylinder proxy geometry, is able to navigate (in a discrete or continuous way) on scenes sourced from existing synthetic and reconstructed datasets of indoor scenes such as SUNCG and Matterport respectively. Such agent can obtain information from multimodal sensory inputs: RGB, depth, surface normals, contact forces, semantic segmentation, and various egocentric measurements such as velocity and acceleration. The simulator provides both Python and web client APIs to control the agent and set the parameters of the scene. However, this simulator lacks some features such as a fully 3D robot model instead of a geometry proxy, photorealism, configurable cameras and points of view, and interactions with the scene.

\subsection{Other Tools and Environments}

Although not strictly related, we would like to remark a couple of tools from which we drew inspiration to shape our proposal: UnrealCV, Gazebo, and NVIDIA's Isaac Sim.

On the one hand, UnrealCV \cite{Qiu2016}\cite{Qiu2017} is a project that extends \ac{UE4} to create virtual worlds and ease communication with computer vision applications. UnrealCV consists of two parts: server and client. The server is a plugin that runs embedded into an \ac{UE4} game. It uses sockets to listen to high-level UnrealCV commands issued by a client and communicates with UE4 through its C++ API to provide advanced functionality for each command, e.g., rendering per-instance segmentation masks. The client is a Python API which communicates with the server using plain text protocol. It just sends those available commands to the server and waits for a response. A detailed list of commands can be consulted in the official documentation of the plugin. We took the main concept and design behind UnrealCV and implemented the whole pipeline inside \ac{UE4} itself to be more efficient and customizable. Another framework which helped us design our environment was Gazebo \footnote{\url{http://http://gazebosim.org/}} a well-known robot simulator that enables accurate and efficient simulation of robots in indoor and outdoor environments. It integrates a robust physics engine (Bullet, ODE, Simbody, and DART), advanced 3D graphics (using OGRE), and sensors and noise modelling. On the other hand, NVIDIA's Isaac Sim is a yet to be released virtual simulator for robotics that lets developers train and test their robot software using highly realistic virtual simulation environments. However, its software development kit is still in early access at the time this work was carried out.

\subsection{Our Proposal in Context}

After analyzing the strong points and weaknesses of the most popular indoor robotic environments, we aimed to combine the strengths of all of them while addressing their weaknesses and introducing new features. In this regard, our work focuses on simulating a wide range of common indoor robot actions, both in terms of poses and object interactions, by leveraging a human operator to generate plausible trajectories and grasps in virtual reality. To the best of our knowledge, this is the first extremely photorealistic environment for robotic vision in which interactions and movements can be realistically simulated in virtual reality. Furthermore, we make possible the generation of raw data (RGB-D/3D/Stereo) and ground truth (2D/3D class and instance segmentation, 6D poses, and 2D/3D bounding boxes) for many vision problems. Although UnrealCV is fairly similar to our work, since both aim to connect Unreal Engine and computer vision/robotics, we took radically different design decisions: while its architecture is a Python client/server, ours is contained entirely inside \ac{UE4} in C++. That architecture allows us to place objects, cameras, and skeletons, and generate images in a more efficient way than other frameworks. Finally, the whole pipeline and tools are released as open-source software with extensive documentation\footnote{\url{https://github.com/3dperceptionlab/unrealrox}}.

\section{System}
\label{sec:system}

\begin{figure}
	\centering
	\includegraphics[width=0.45\textwidth]{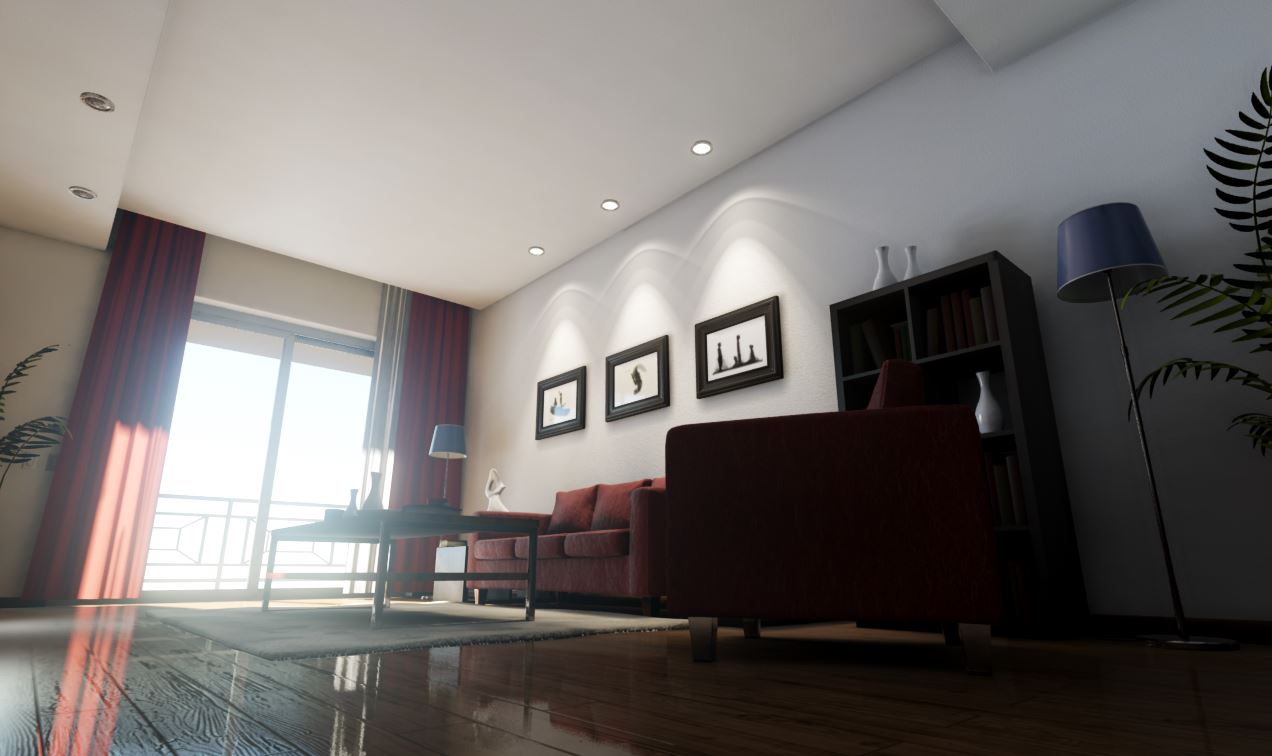}
	\\
	\includegraphics[width=0.45\textwidth]{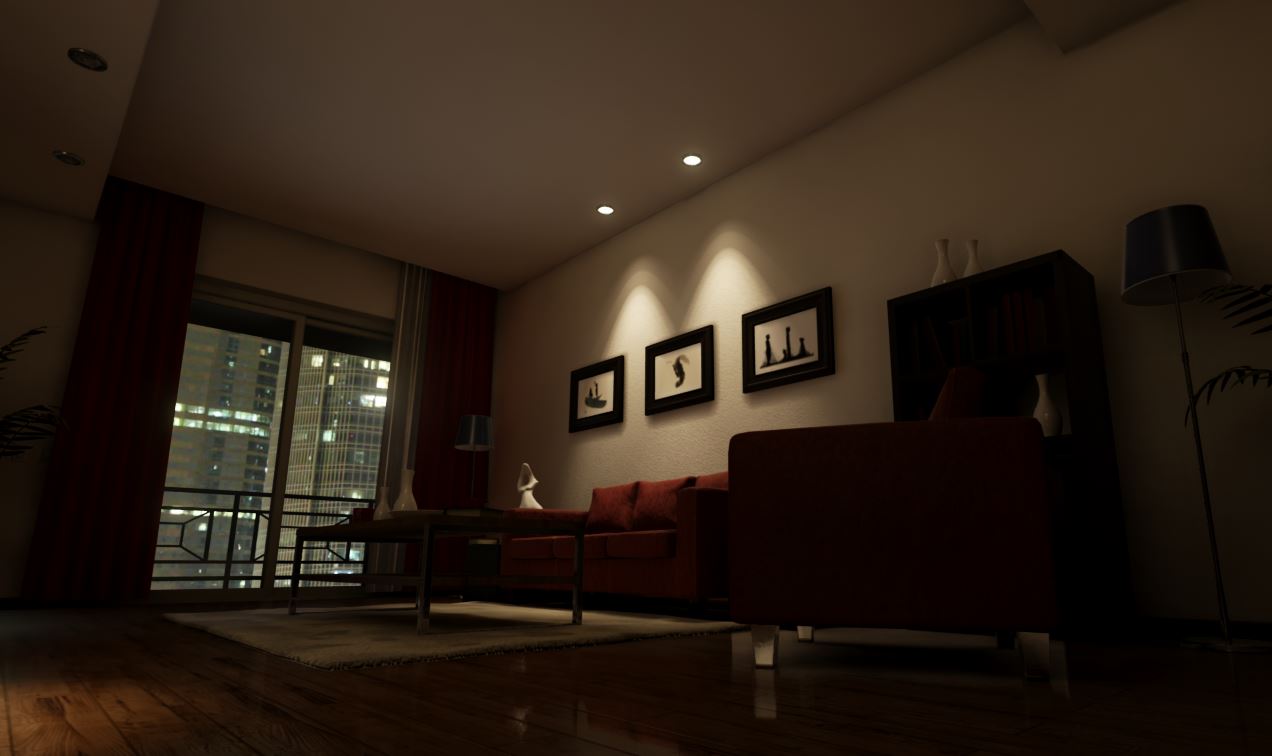}
	\caption{Snapshots of the daylight and night room setup for the \emph{Realistic Rendering} released by Epic Games to showcase the realistic rendering capabilities of \ac{UE4}.}
	\label{fig:realistic_rendering}
\end{figure}

The rendering engine we chose to generate photorealistic RGB images and immerse the agent in \acs{VR} is \acf{UE4}. The reasons for this choice are the following ones: (1) it is arguably one of the best game engines able to produce extremely realistic renderings, (2) beyond gaming, it has become widely adopted by \acl{VR} developers and indoor/architectural visualization experts so a whole lot of tools, examples, documentation, and assets are available; (3) due to its impact across various communities, many hardware solutions offer plugins for \acs{UE4} that make them work out-of-the-box; and (4) Epic Games provides the full C++ source code and updates to it so the full suite can be used and easily modified for free. Arguably, the most attractive feature of \acs{UE4} that made us take that decision is its capability to render photorealistic scenes like the one shown in Figure \ref{fig:realistic_rendering}. Some UE4 features that enable this realism are: physically-based materials, pre-calculated bounce light via Lightmass, stationary lights, post-processing, and reflections.

It is also important to remark that we do have strict real-time constraints for rendering since we need to immerse a human agent in virtual reality, i.e., we require extremely realistic and complex scenes rendered at very high framerates (usually more than 80 FPS). By design, \acs{UE4} is engineered for virtual reality so it provides a specific rendering solution for it named Forward Renderer. That renderer is able to generate images that meet our quality standards at 90 FPS thanks to high-quality lighting features, \ac{MSAA}, and instanced stereo rendering.

\begin{figure}
	\centering
	\includegraphics[width=\linewidth]{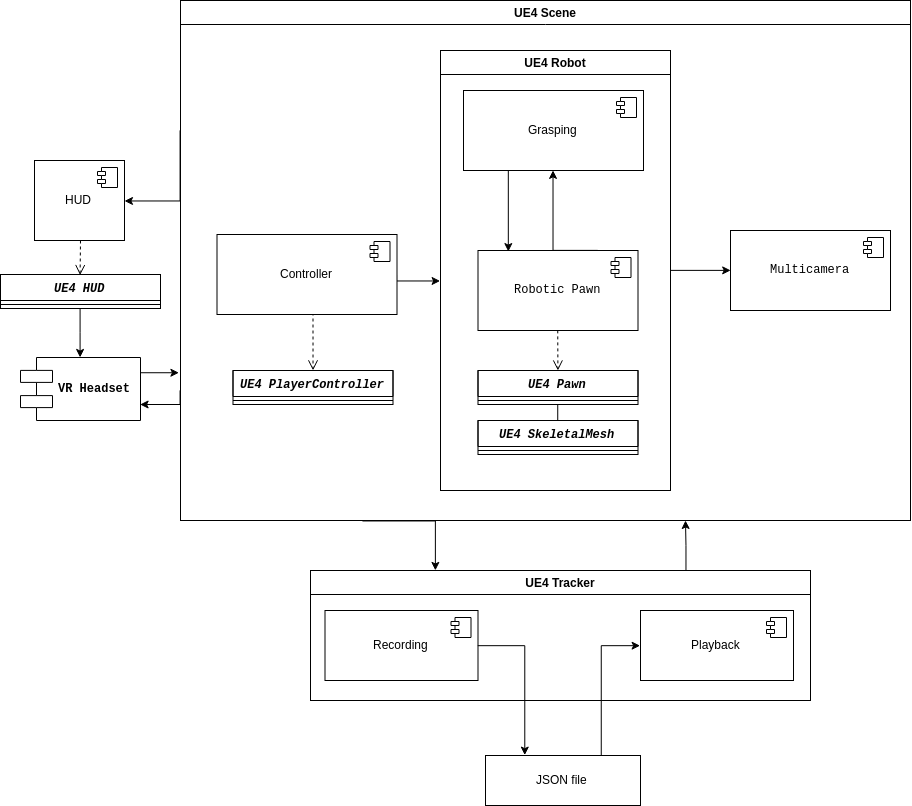}
	\caption{System diagram showing the various subsystems and their abstract relationships: Robotic Pawn, Controller, \acs{HUD}, Multi-camera, Grasping, Recording, and Playback.}
	\label{fig:diagram}
\end{figure}

The whole system is built over \acs{UE4} taking advantage of various existing features, extending certain ones with to suit our specific needs, and implementing others from scratch to devise a more efficient and cleaner project that abides to software design principles. A general overview of our proposal is shown in Figure \ref{fig:diagram}. In this section we describe each one of the subsystems that our proposal is composed of: robotic pawns, controller, \acs{HUD}, grasping, multi-camera, recording, and playback.

\subsection{Robotic Pawns}

\begin{figure}
	\centering
	\includegraphics[width=0.49\linewidth]{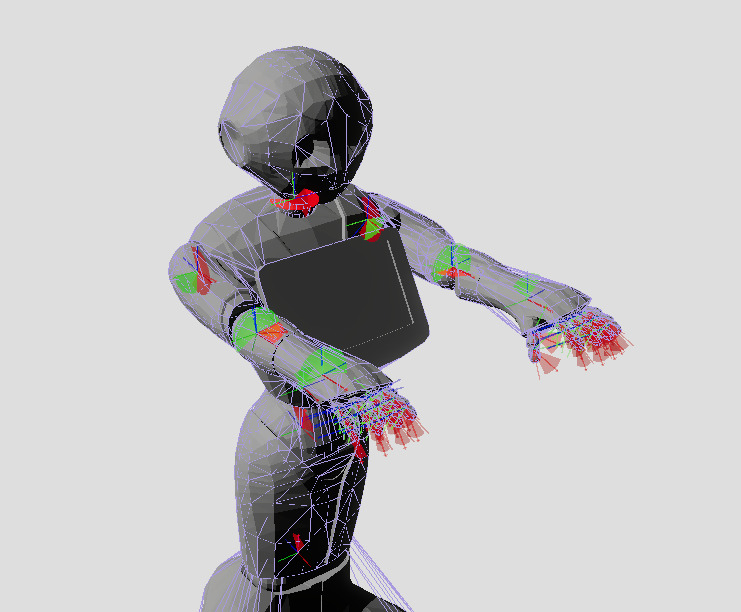}
	\includegraphics[width=0.49\linewidth]{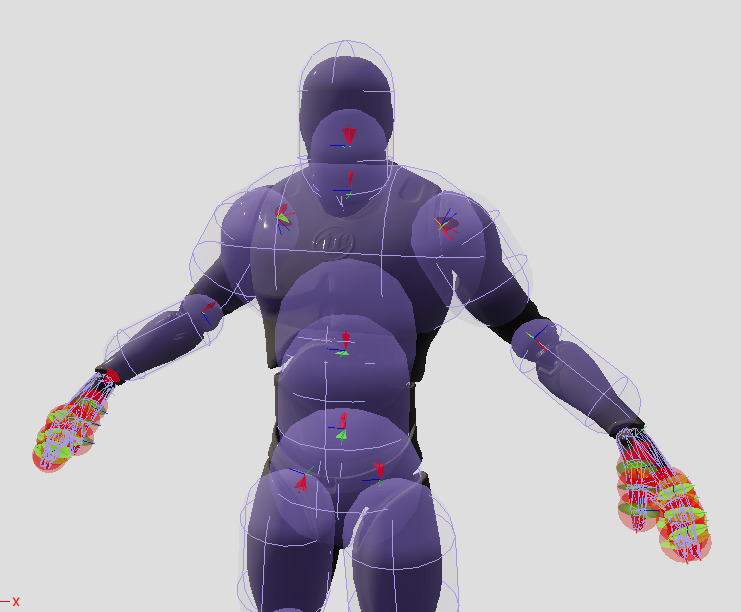}
	\caption{Pepper and Mannequin integrated with colliders and constraints.}
	\label{fig:robot_pawns}
\end{figure}

One of the most important parts of the system is the representation of the robots in the virtual environment. Robots are represented by the mesh that models them, the control and movement logic, the animations that it triggers, and the grasping system (explained later in its corresponding section). To encapsulate all this, we have created a base class that contains all the common behavior that any robot would have in our system, which can then be extended by child classes that implement specific things such as the mesh or the configuration of the fingers for the grasping system. Using that encapsulation, we introduced two sample robots in our environment: \ac{UE4}'s mannequin and Aldebaran's Pepper (see Figure \ref{fig:robot_pawns}).

In \ac{UE4} there is a hierarchy of predefined classes ready to work together that should be used properly in order to take advantage of the facilities offered by the engine. For example, any element that we want to place in a scene must extend the \emph{Actor} class, and at the same time, an \emph{Actor} that is supposed to receive inputs from the user must extend the \emph{Pawn} class (and optionally can have a \emph{Controller} class to abstract input events, as we will see in the next section). This means that our base class that represents the common behavior of robots must extend \emph{Pawn} class.

The meshes that model characters with joints like our robots are called \emph{SkeletalMesh} in \ac{UE4}. In addition to the mesh that defines their geometry, they incorporate a skeleton that defines how that geometry will be deformed according to the relative position and rotation of its bones. An \emph{SkeletalMesh} is added as a component to our class (actually, an instance of \emph{SkeletalMeshComponent}).

There are two types of inputs to which our robots must react to, those that come from pressing buttons or axes, and those that come from moving the \ac{VR} motion controllers. The latter is managed by an \ac{UE4} component that must be added to our \emph{Pawn} class and that will modify its position according to the real-world motion controllers movement. We will be able to access the position of these components from the animation class, associate it with the hand bones of the robot \emph{SkeletalMesh}, and move the whole arm by inverse kinematics.

The animation class is created from the \emph{SkeletalMesh}, so it is separate from the \emph{Pawn} class, although the first has to access information from the second. Specifically, our animation classes handles the hand closing animation for the grasping system, and, in the case of robots with legs, it also takes control of the displacement speed to execute the walking animation at different speeds. Finally, the animation class is also used by the playback system (described below) to recover the \emph{SkeletalMesh} pose for a single frame, since it is from where the position and rotation of each joint of the \emph{SkeletalMesh} is accessible for modification.

\subsection{Controller Subsystem}

\begin{figure}[hptb]
		\includegraphics[width=0.48\linewidth]{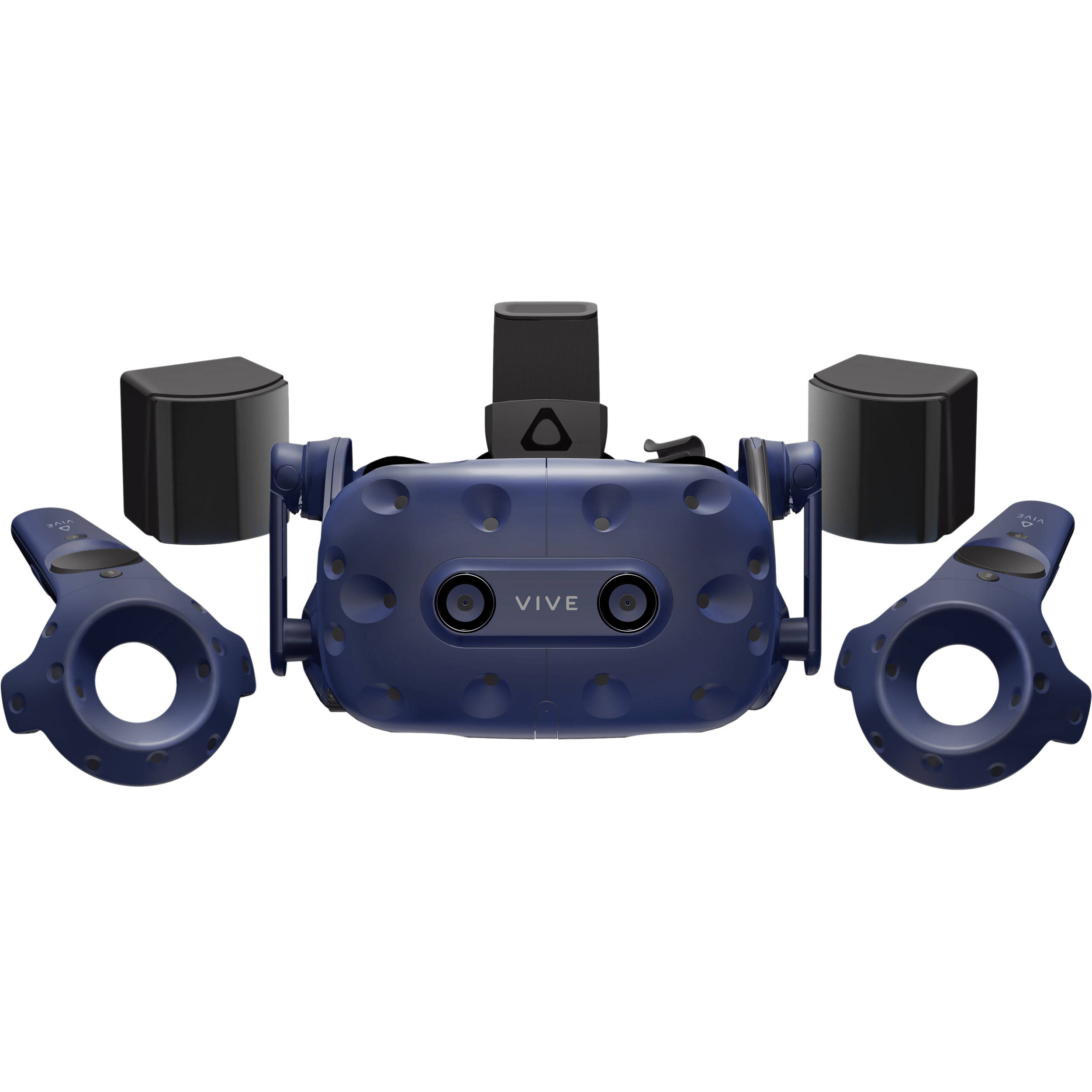}
		\includegraphics[width=0.48\linewidth]{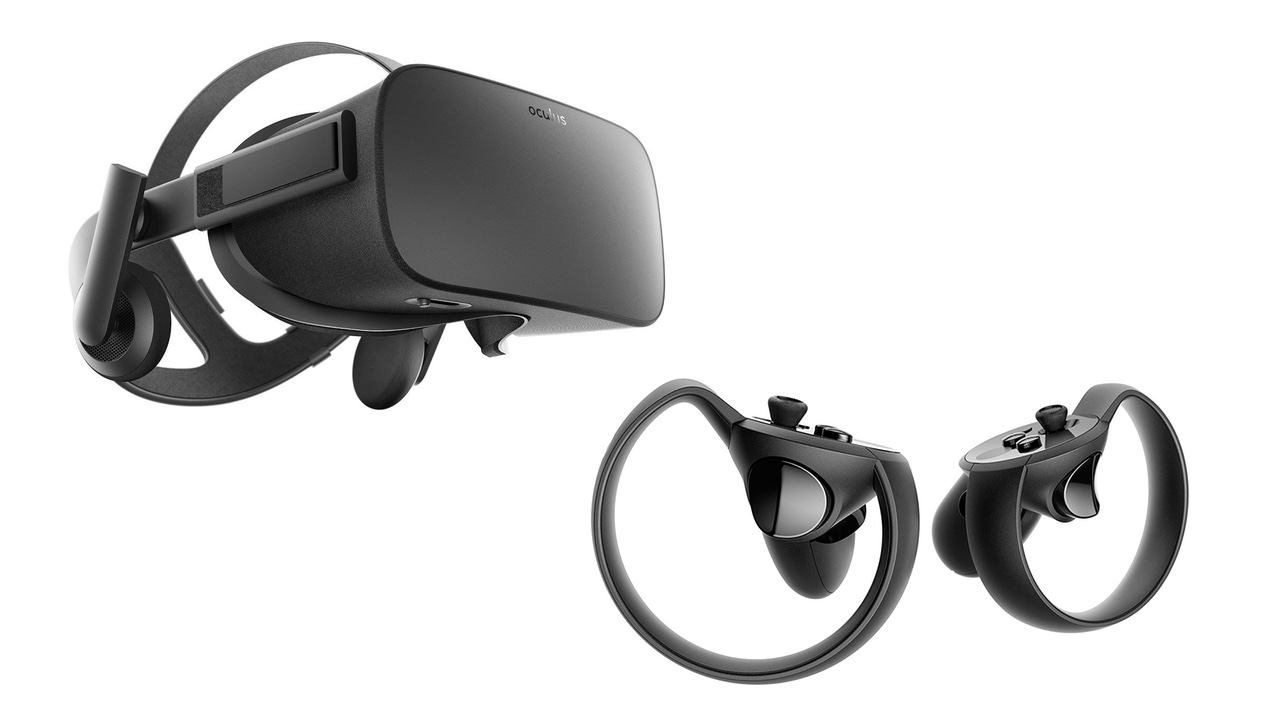}
	\caption{Seamlessly supported \ac{VR} headsets thanks to the decoupled controller subsystem: HTC Vive Pro and Oculus Rift.}
	\label{fig:headsets}
\end{figure}

We would like our system to seamlessly support a wide range of \acl{VR} setups to reach a potentially higher number of users. In this regard, it is important to decouple the controller system from the rest of the environment so that we can use any device (such as the Oculus Rift and the HTC Vive Pro shown in Figure \ref{fig:headsets}) without excessive effort. To that end, it is common to have a class that handles all the inputs from the user (in an event-driven way) and then distributes the execution to other classes depending on that input. The very same \ac{UE4} provides the base class for this purpose, 
namely \emph{PlayerController}. Many of these user inputs are focused on controlling the movement and behavior of a character in the scene, usually represented in \acl{UE4} as a \emph{Pawn} class. This means that the \emph{PlayerController} class is closely related to the \emph{Pawn} one. Decoupling input management from functionality is useful as it allows us switching among different controllers for the same \emph{Pawn} (different control types for example), or use the same controller for several ones (if they have the same behavior for inputs).

Our controller system extends the base class \emph{PlayerController} and handles all kind of user inputs, both from keyboard and \ac{VR} controllers (we have tested our system with Oculus Rift and HTC Vive). This is configured in the \ac{UE4} editor, more specifically in the \emph{Input Project Preferences} panel, where several keys, buttons, or axes can be associated with an event name, which is later binded to a handler function in the custom \emph{PlayerController} class. The controller calls the movement and grasping functionalities from the pawn, and also global system functions as toggling the recording system, restarting the scene, and resetting the \ac{VR} headset position. It also controls the \ac{HUD} system for showing input debugging feedback.

\subsection{HUD Subsystem}

\begin{figure}[!h]
	\centering
	\includegraphics[width=\linewidth]{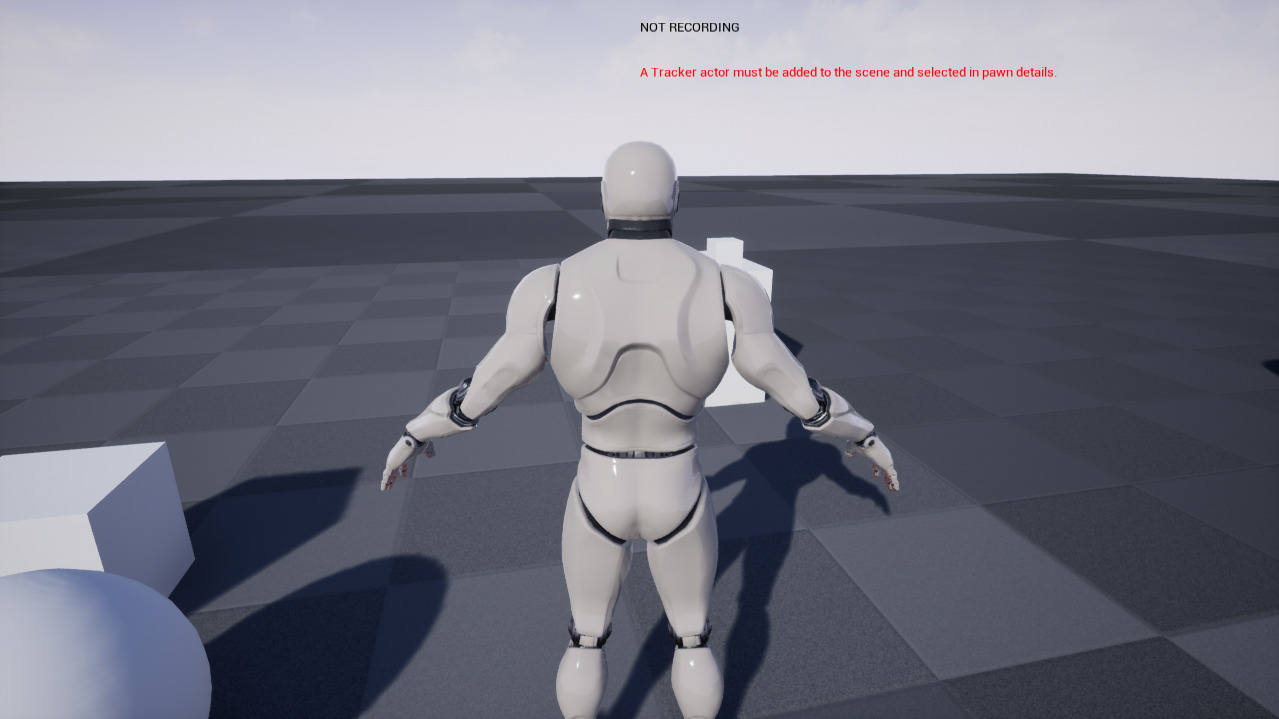}
	\caption{Information and error messages shown in the \ac{HUD}.}
	\label{fig:notrackererror}
\end{figure}

It is convenient for any system to feature a debug system that provides feedback about the application state to the user. In UnrealROX, we offer an interface to show information at various levels to the user if requested. This information is presented in a \ac{HUD} which can be turned off or on to the user's will. It can even be completely decoupled from the system as a whole for maximum performance. The main information modalities provided by the \ac{HUD} are the following ones:

\begin{itemize}
	\item Recording: A line of text with the recording state is always shown in the HUD in order to let the user know if his movements through the scene are being recorded.
	\item States: Notifies the user with a message on the screen of the relevant buttons pressed, the joints in contact with an object, the profiling being activated, etc. The amount of seconds these messages last on screen can be established independently. Most of them are printed for 5 seconds.
	\item Error: Prints a red message indicating an error that lasts in screen for 30 seconds (or until another error occurs). An example of this would be trying to record without the tracker on the scene (as seen in Figure \ref{fig:notrackererror}).
	\item Scene Capture: It allows us to establish a debugging point of view so that we can see our robot from a different point of view than the first person camera.
\end{itemize}

\begin{figure}[h]
	\centering
	\includegraphics[width=\linewidth]{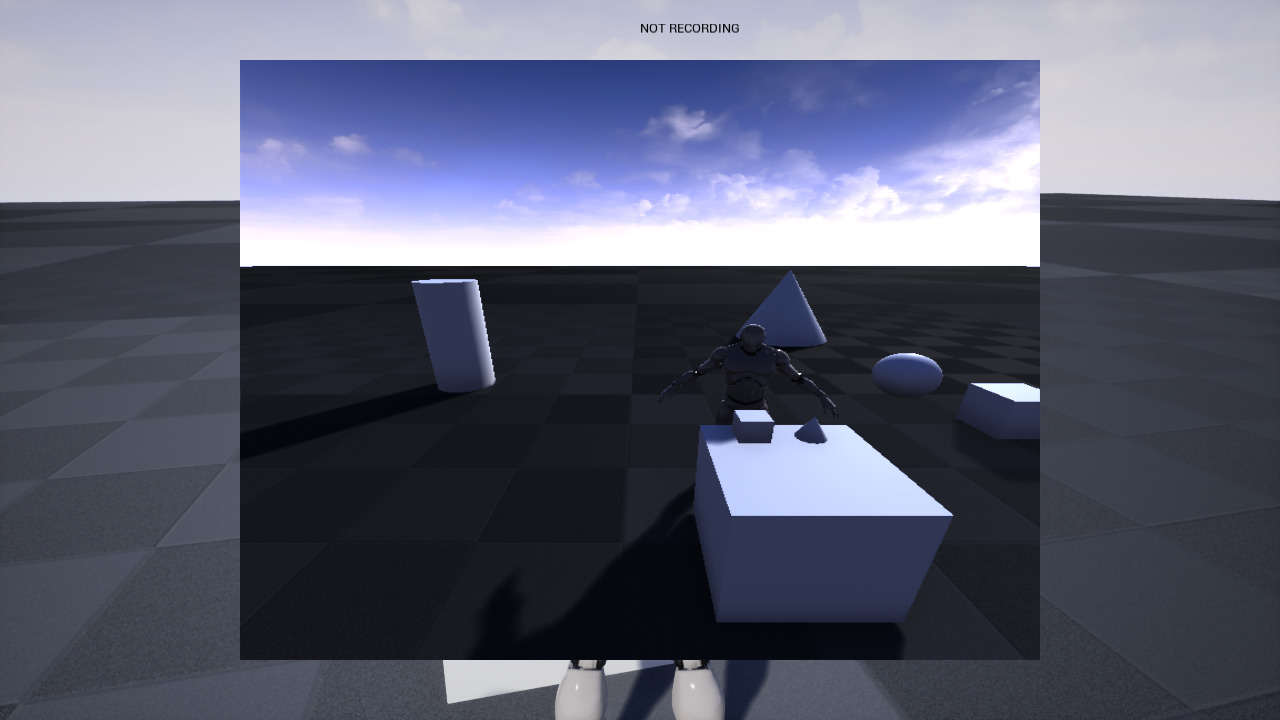}
	\caption{Scene Capture drawn in the viewport.}
	\label{fig:scenecapture}
\end{figure}

We have implemented this functionality extending the \ac{HUD} class that \ac{UE4} provides, and we also made it fully decoupled from the rest of the system in a simple way by implementing an interface\footnote{\url{https://docs.unrealengine.com/en-US/Programming/UnrealArchitecture/Reference/Interfaces}}. Classes that inherite from HUD class have a canvas and a debug canvas on which primitive shapes can be drawn. It provides some simple methods for rendering text, textures, rectangles, and materials which can also be accessed from blueprints. An example of texture drawing in practice in our project is the Scene Capture, which consists in drawing a texture in the viewport captured from an arbitrary camera (as shown in Figure \ref{fig:scenecapture}). This will be useful for the user to see if the animations are being played correctly in a \acl{VR} environment.

\subsection{Grasping Subsystem}

Grasping subsystem is considered one of the core components of UnrealROX. We have focused on providing a realistic grasping, both in the way the robot grasp an object and in the movements it makes. When grasping an object we need to simulate a real robot behaviour, thus smooth and plausible movements are needed. The grasping action is fully controlled by the user through the controls, naturally limited to the degrees of freedom of the human body. In this way, we achieve a good representation of a humanoid robot interacting in a realistic home environment, also known as assistive robots which are the current trend in the field of social robotics. 

Current approaches for grasping in \ac{VR} environments are animation-driven, and based on predefined movements \cite{OculusDistanceGrab2017}\cite{VRTemplate2016}\cite{OculusFirstExperience}. This will restrict the system to only a few pre-defined object geometries hindering user's interaction with the environment resulting also in a unrealistic grasping. In contrast with these approaches, the main idea of our grasping subsystem consists in manipulating and interacting with different objects, regardless of their geometry and pose. In this way, the user can freely decide which object to interact with without restrictions. The robot can manipulate an object with each hand, and change an object from a hand to the other. It can also manipulate two different objects at the same time, drop them freely or throw them around the scene.

\begin{figure}[h]
	\centering
	\includegraphics[width=\linewidth]{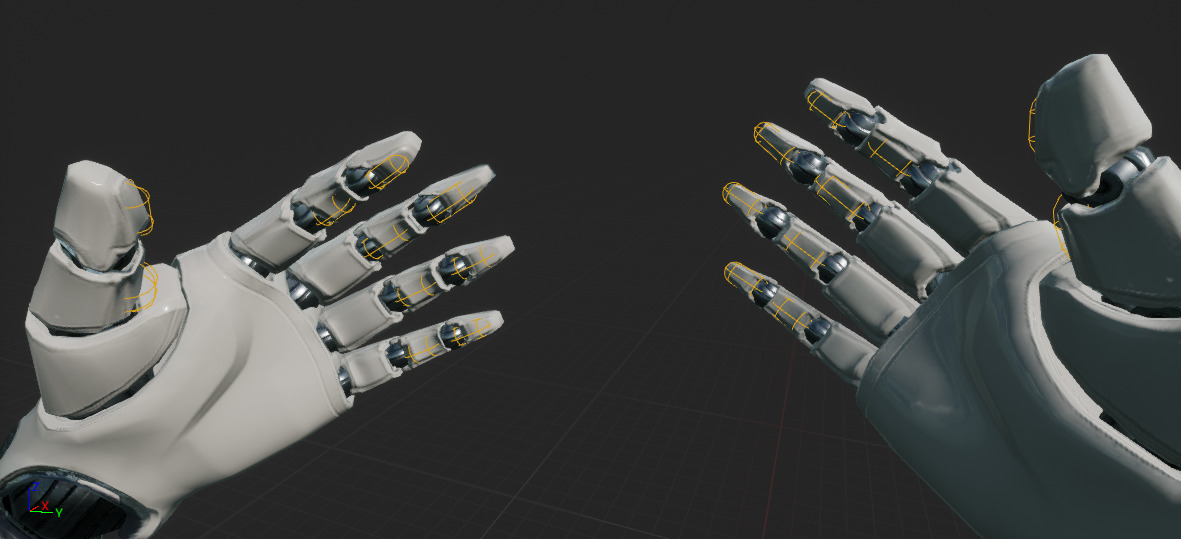}
	\caption{\emph{Sphere trigger volumes} placed on finger phalanges of both hands represented in yellow.}
	\label{fig:handtriggers}
\end{figure}

At the implementation level of this subsystem, we make use of \acs{UE4}'s \emph{trigger volumes} placed on each one of the finger phalanges as we can see in Figure \ref{fig:handtriggers}. These \emph{triggers} act as sensors that will determine if we are manipulating an object in order to grasp it. With the controllers we are able to close robot's hands limiting individually each finger according to the \emph{triggers}. We also implement a logic for determine when to grasp or release an object based on the \emph{triggers} state. Fingers position change smoothly in order to replicate a real robot hand behaviour and to avoid pass through an object.

\begin{figure}[!htb]
	\centering
	\hfill
	\begin{subfigure}{0.49\linewidth}
		\centering
		\includegraphics[width=\linewidth]{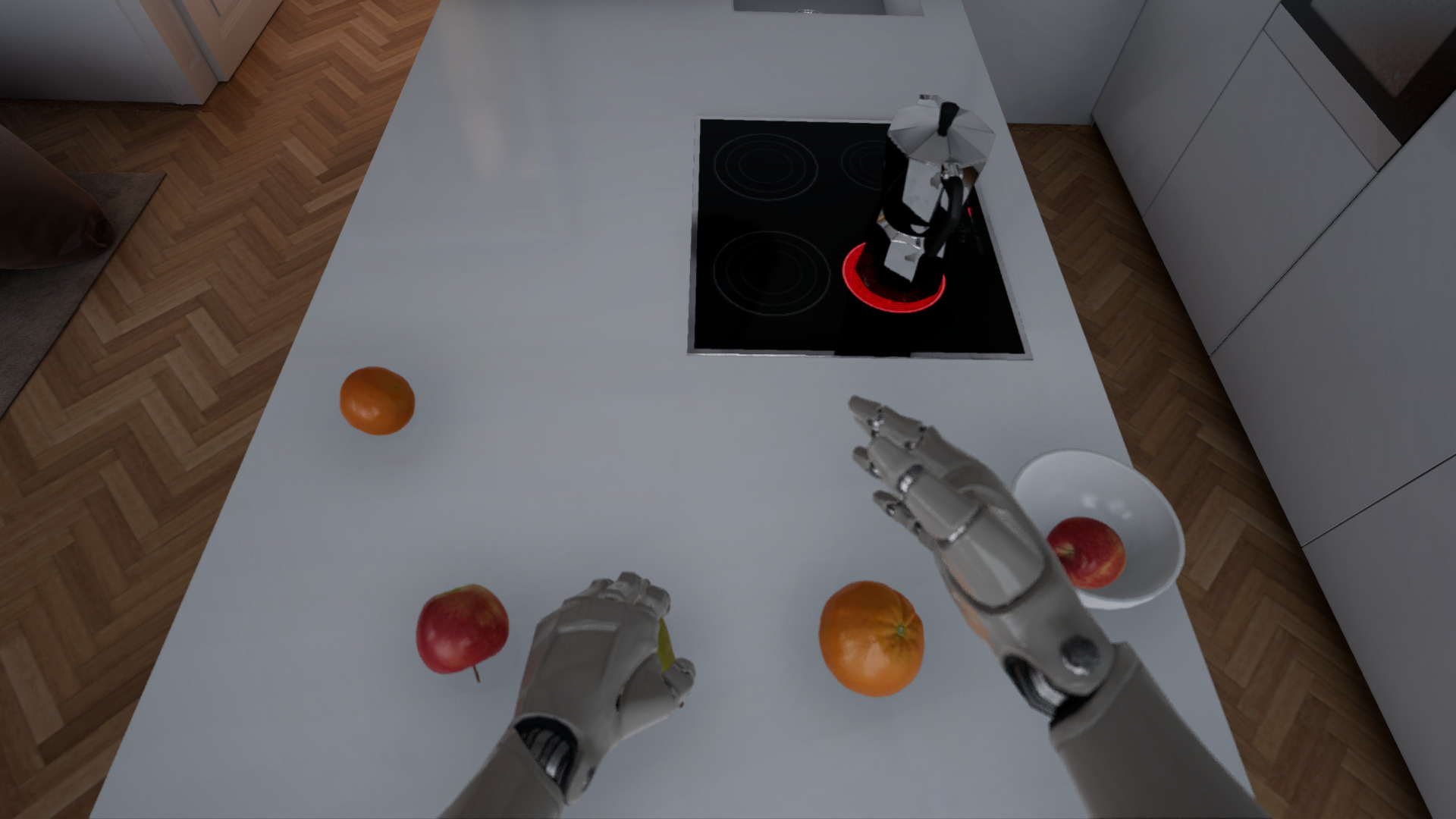}
		\caption{Frame 0}
	\end{subfigure}
	\hfill
	\begin{subfigure}{0.49\linewidth}
		\centering
		\includegraphics[width=\linewidth]{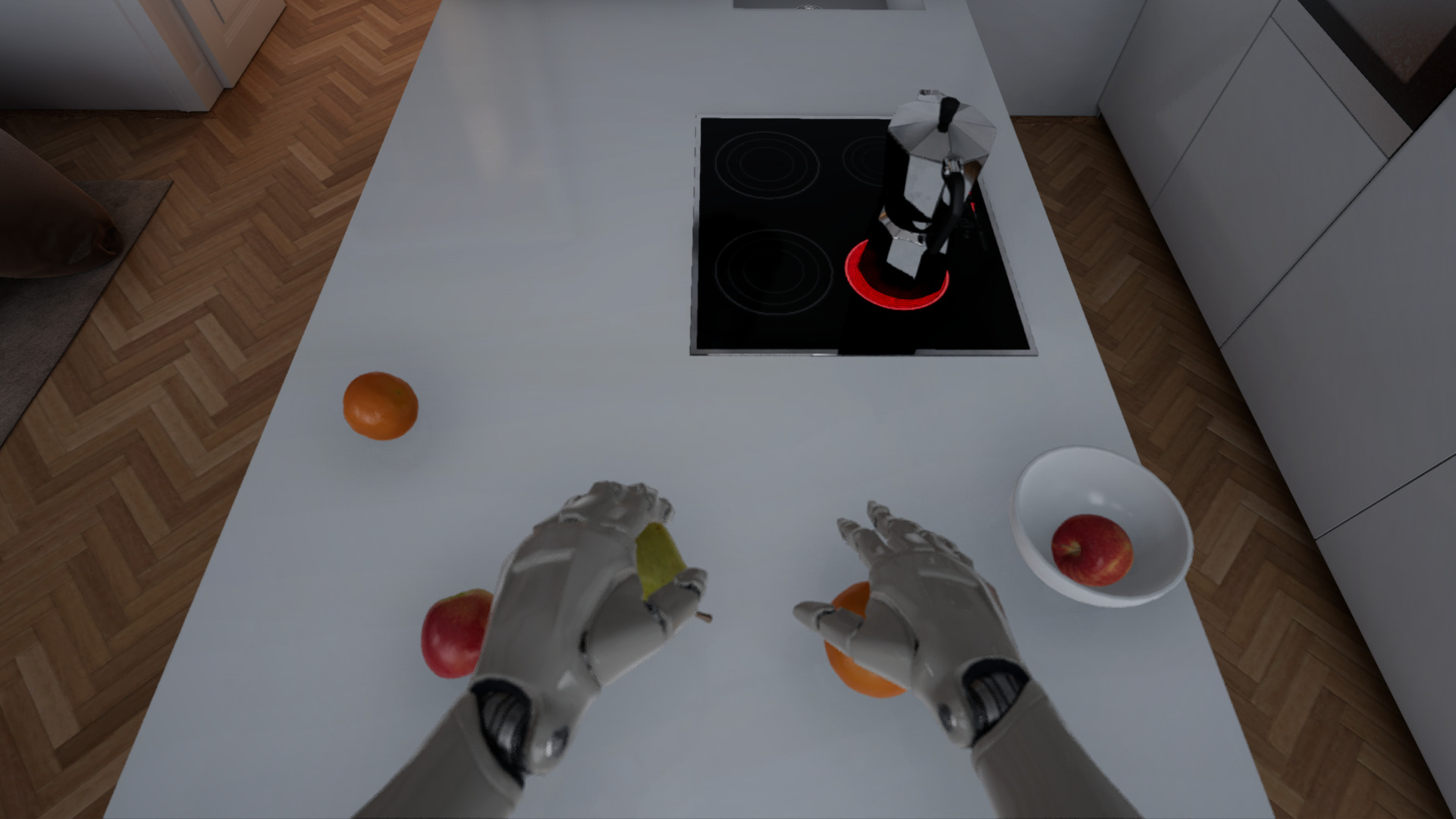}
		\caption{Frame 35}
	\end{subfigure}
	\hfill
	\\
	\hfill
	\begin{subfigure}{0.49\linewidth}
		\centering
		\includegraphics[width=\linewidth]{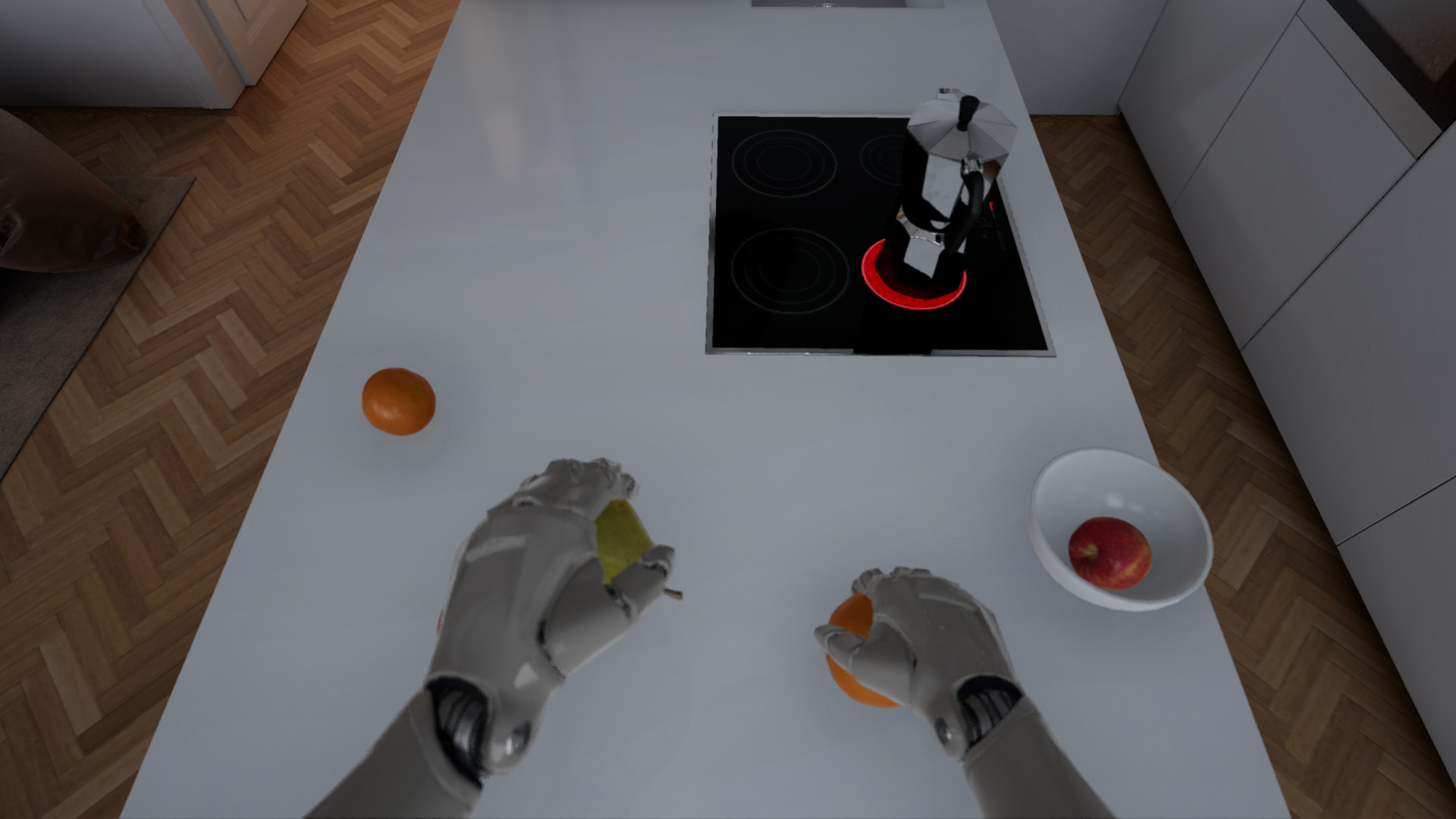}
		\caption{Frame 70}
	\end{subfigure}
	\hfill
	\begin{subfigure}{0.49\linewidth}
		\centering
		\includegraphics[width=\linewidth]{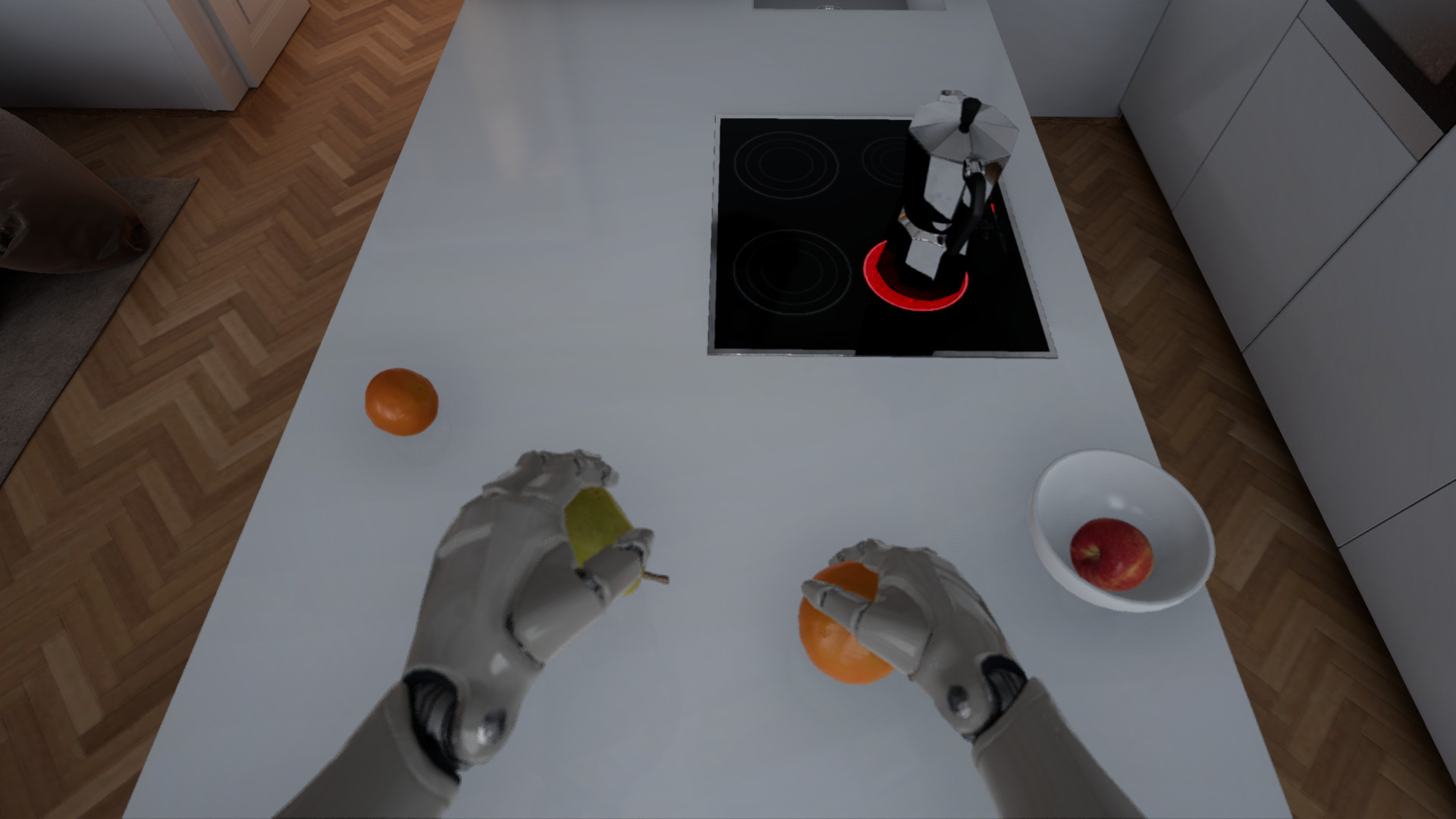}
		\caption{Frame 75}
	\end{subfigure}
	\hfill
	\begin{subfigure}{0.49\linewidth}
		\centering
		\includegraphics[width=\linewidth]{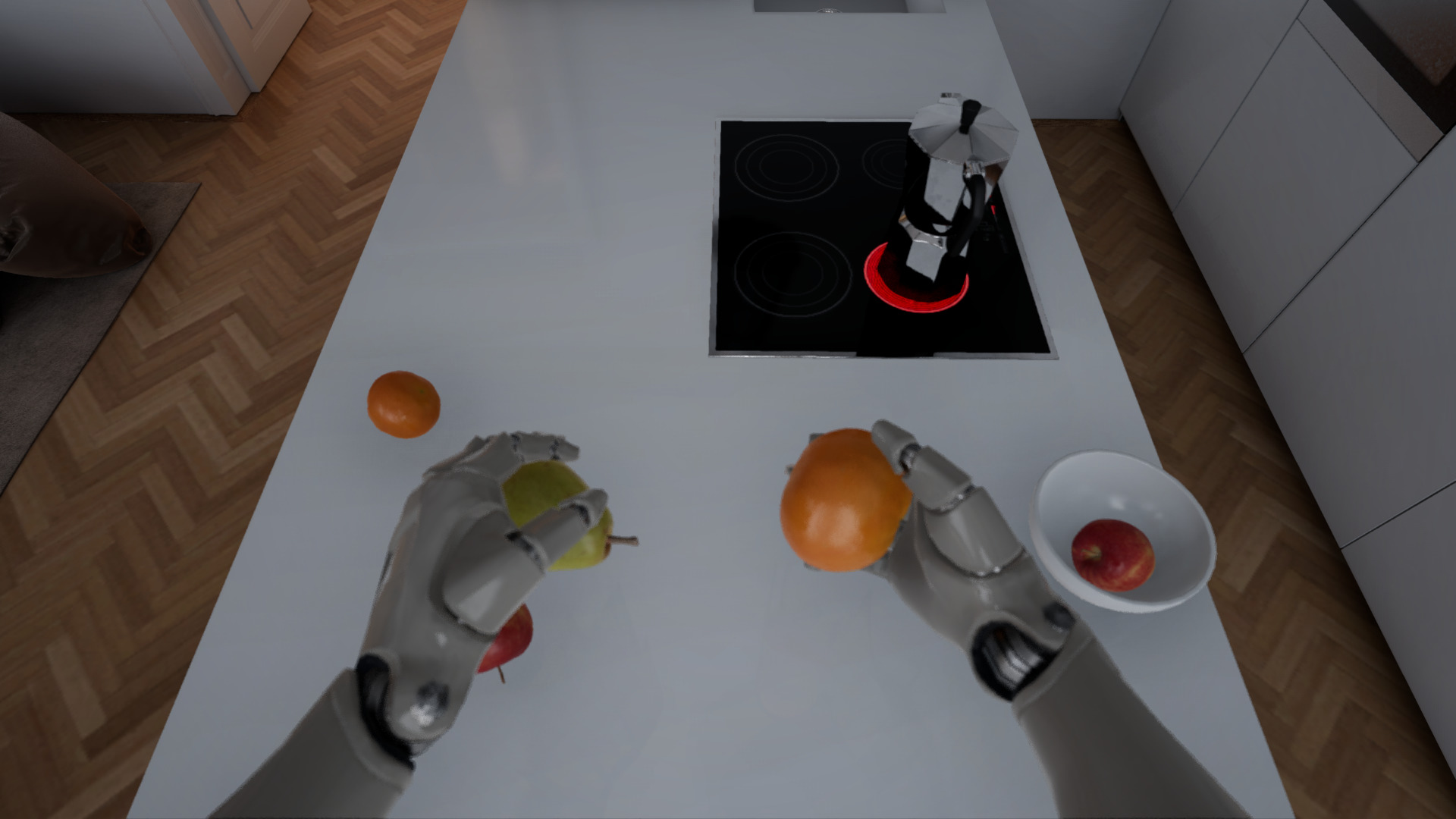}
		\caption{Frame 80}
	\end{subfigure}
	\hfill
	\begin{subfigure}{0.49\linewidth}
		\centering
		\includegraphics[width=\linewidth]{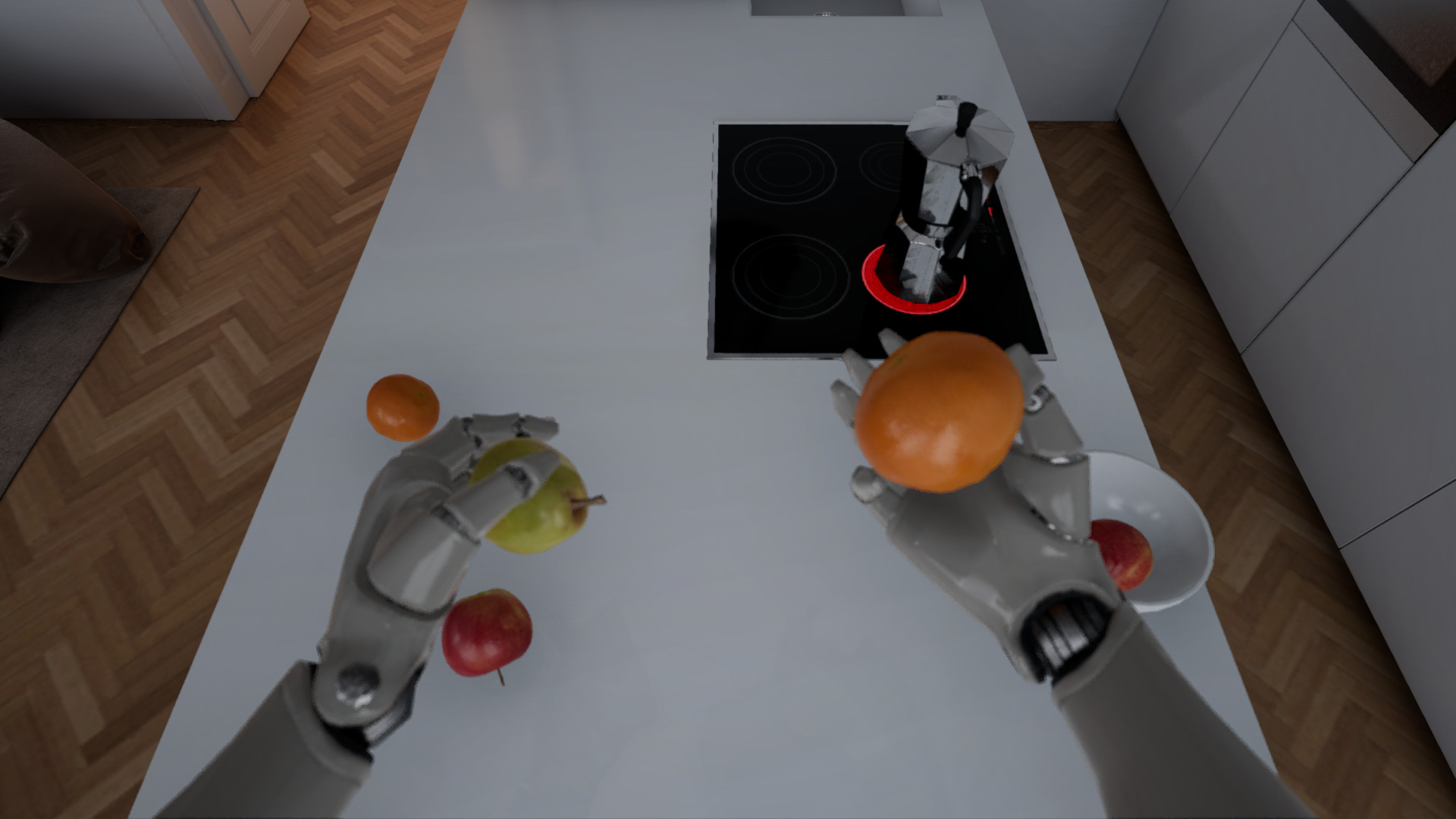}
		\caption{Frame 85}
	\end{subfigure}
	\hfill
	\caption{Sequence of 6 frames ($Seq= {F_0, F_{35}, F_{70}, F_{75}, F_{80}, F_{85}}$) representing a grasping action with right hand meanwhile holding a fruit with the left hand. Frames are left-right and top-down ordered}
	\label{fig:grasping}
\end{figure}

A sequence example grasping two objects with our custom system is shown in Figure \ref{fig:grasping}.

\iffalse
The most common issues of existing grasping systems on virtual reality environments are:
\begin{itemize}
  \item Animation-driven grasping with predefined movements
  \item Impossibility to deal with new object geometries
  \item Unrealistic grasping
\end{itemize}

Through our grasping subsystem we will try to deal with above limitations by providing a realistic grasping, independent of the object geometries and without predefined movements. This will allow us to provide infinite grasping situations depending only on the user's ability of grasping an object or interact with the scene. 
\fi

\subsection{Multi-camera Subsystem}

Most robots in the public market (such as Pepper\footnotetext{\url{https://www.softbankrobotics.com/emea/en/robots/pepper}} or Baxter\footnotetext{\url{https://www.rethinkrobotics.com/baxter/}}) integrate multiple cameras in different parts of their bodies. In addition, external cameras are usually added to the system to provide data from different points of view, e.g., ambient assisted living environments tend to feature various camera feeds for different rooms to provide the robot with information that it is not able to perceive directly. In UnrealROX, we want to simulate the ability to add multiple cameras in a synthetic environment with the goal in mind of having the same or more amount of data that we would have in a real environment. For instance, in order to train a data-driven grasping algorithm it would be needed to generate synthetic images from a certain point of view: the wrist of the robot. To simulate this situation in our synthetic scenario, we give the user the ability to place cameras attached to sockets in the robot’s body, e.g., the wrist itself or the end-effector (eye-in-hand). Furthermore, we also provide the functionality to add static cameras over the scene.

To implement this subsystem, we make use of \acs{UE4}'s \emph{CameraActor} as the camera class and the \emph{Pawn} class as the entity to which we will attach them. By default, \acs{UE4} does not allow us to precisely attach components in the editor so it is necessary to define a socket-camera relationship in the \emph{Pawn} class. This is due to the fact that it has direct access to the skeleton to which we will be attaching specific cameras. 

The objective of the \emph{CameraActor} class is to render any scene from a specific point of view. This actor can be placed and rotated at the user’s discretion in the viewport, which makes them ideal for recording any type of scene from any desired point of view. The \emph{CameraActor} is represented in \acs{UE4} by a 3D camera model and like any other actor, it can be moved and rotated in the viewport. Apart from handling attached and static cameras, UnrealROX exposes the most demanded camera settings through its interface (projection mode, \ac{FoV}, color grading, tone mapping, lens, and various rendering effects), as well as providing additional features such as creating stereo-vision setups.

\begin{figure}[hptb]
	\begin{subfigure}{.49\textwidth}
		\includegraphics[width=\textwidth]{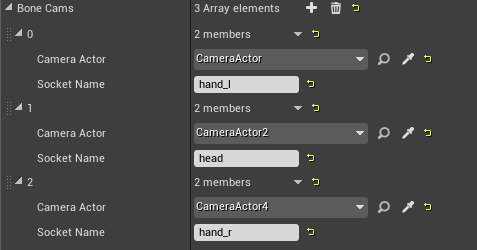}
		\caption{Representation of the array.}
		\label{fig:g1}
	\end{subfigure}%
	\hfill
	\begin{subfigure}{.49\textwidth}
		\includegraphics[width=\textwidth]{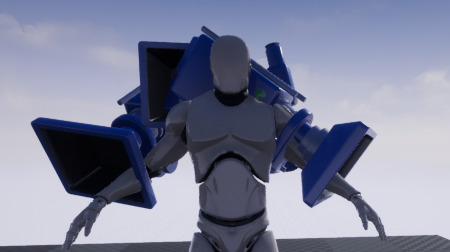}
		\caption{Pawn Actor with cameras.}
		\label{fig:g2}
	\end{subfigure}
	\caption{In-engine representation of the array of structs. In (\subref{fig:g1}) we can see the representation of the array struct in the instance of the object, while in (\subref{fig:g2}) we see its visual     representation in the engine.}
	\label{fig:viewportstruct}
\end{figure}

To implement the camera attachment functionality we make extensive use of the \emph{AttachToActor} function provided by \acs{UE4}, which is in charge of parenting one actor with another following some attachment rules. We can specify the socket to which we want to attach the object. This means that when the selected socket changes its transform, the attached object will change it too according to the specified \emph{AttachmentRules}. These rules dictate how this new attached actor will behave when the socket it is linked moves or rotates. The \emph{AttachmentRules} can be defined separately for location, rotation, and scale. This lead us to define an implicit relationship between the \emph{CameraActor} and the socket it is attached to. For that, the \emph{Pawn} class implements an array of \emph{USTRUCT} so that each relationship has two parameters: the camera itself and the socket name. These properties are accompanied by the \emph{EditAnywhere} meta, which makes possible the edition of the properties not only on the Class Default Object (CDO) but also on its instances. The user will be in charge of filling the array specified in Listing 3. To make the attachment process easier, we provide a friendly user interface inside the editor (see Figure \ref{fig:viewportstruct}).

\subsection{Recording Subsystem}

UnrealROX decouples the recording and data generation processes so that we can achieve high framerates when gathering data in \acf{VR} without decreasing performance due to extra processing tasks such as changing rendering modes, cameras, and writing images to disk. In this regard, the recording subsystem only acts while the agent is embodied as the robot in the virtual environment. When enabled, this mode gathers and dumps, on a per-frame basis, all the information that will be needed to replay and reconstruct the whole sequence, its data, and its ground truth. That information will be later used as input for the playback system to reproduce the sequence and generate all the requested data.

In order to implement such behavior we created a new \ac{UE4} \emph{Actor}, namely \emph{ROXTracker}, which overrides the \emph{Tick} function. This new invisible actor is included in the scene we want to record and executes its tick code for each rendered frame. That tick function loops over all cameras, objects, and robots (skeletons) in the scene and writes all the needed information to a text file in an asynchronous way. For each frame, the actor dumps the following information: recorded frame number, timestamp in milliseconds since the start of the game, the position and rotation for each camera, the position, rotation, and bounding box minimum and maximum world-coordinates for each object, and the position and rotation of each joint of the robot's skeleton.

\begin{figure}[!htb]
	\centering
	\includegraphics[width=\linewidth]{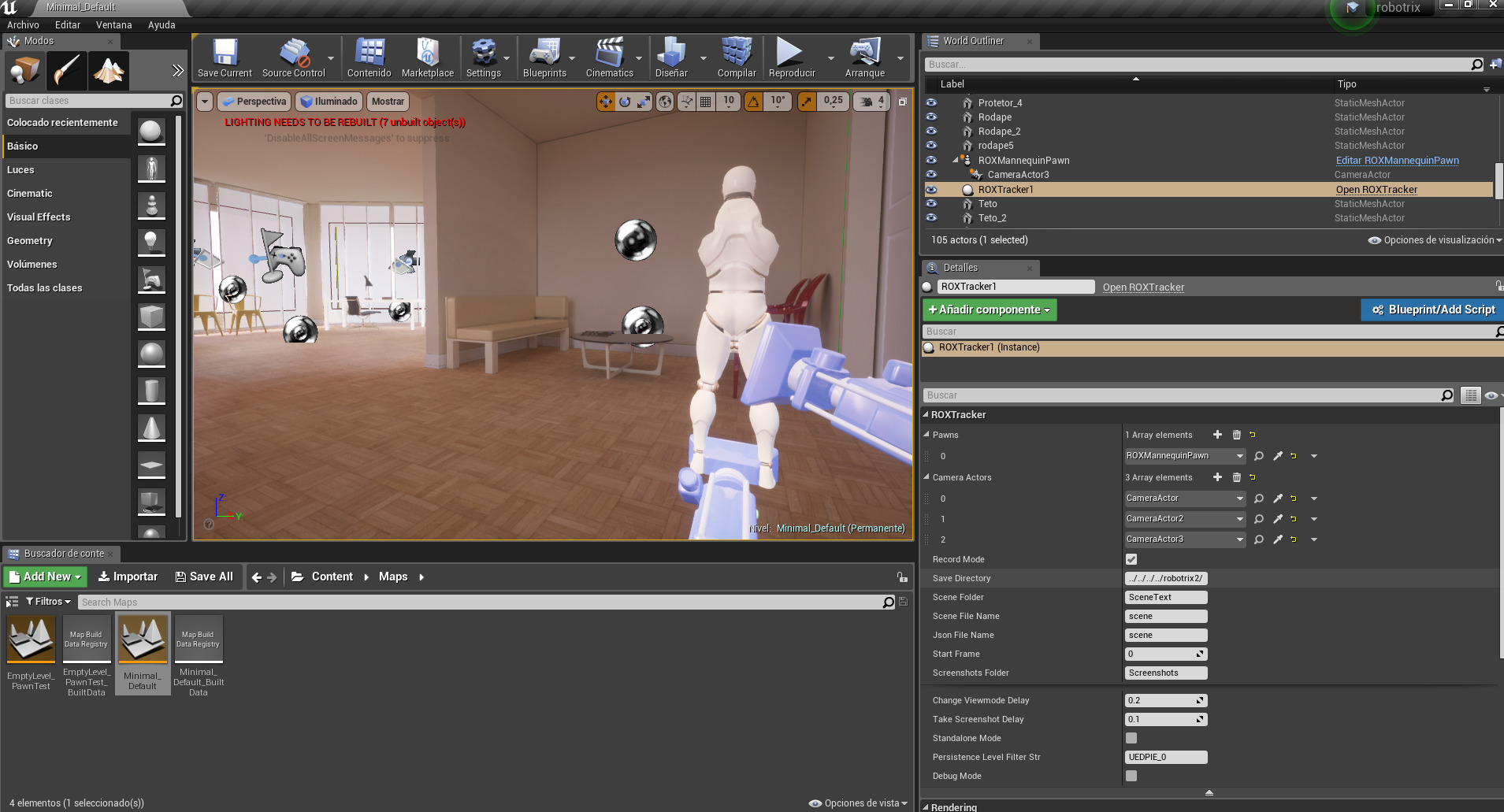}
	\caption{\emph{ROXTracker} custom interface showing the robot mannequin, multiple cameras, and various parameters to configure which pawns and cameras are tracked and other sequence details.}
	\label{fig:trackerui}
\end{figure}

The information is dumped in raw text format for efficiency, after the sequence is fully recorded, the raw text file is processed and converted into a more structured and readable JSON file so that it can be easily interpreted by the playback system.

\subsection{Playback Subsystem}

Once the scene has been recorded, we can use the custom user interface in \ac{UE4} to provide the needed data for the playback mode: the sequence description file in JSON format and an output directory. Other parameters such as frame skipping (to skip a certain amount of frames at the beginning of the sequence) and dropping (keep only a certain amount of frames) can also be customized (see Figure \ref{fig:trackerui}).

\begin{figure}[!htb]
	\centering
	\hfill
	\begin{subfigure}{0.49\linewidth}
		\centering
		\includegraphics[width=\linewidth]{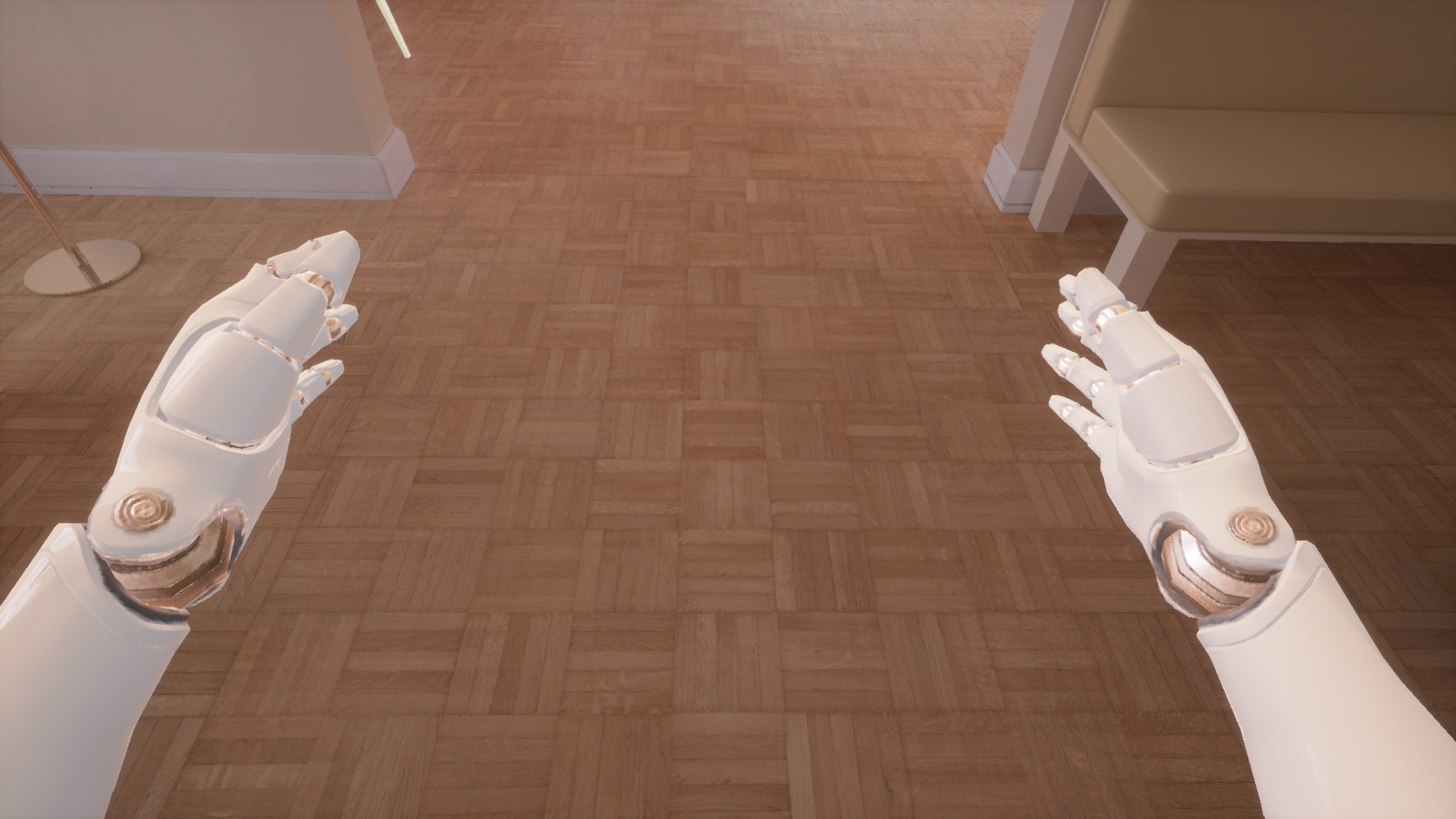}
		\caption{RGB}
	\end{subfigure}
	\hfill
	\begin{subfigure}{0.49\linewidth}
		\centering
		\includegraphics[width=\linewidth]{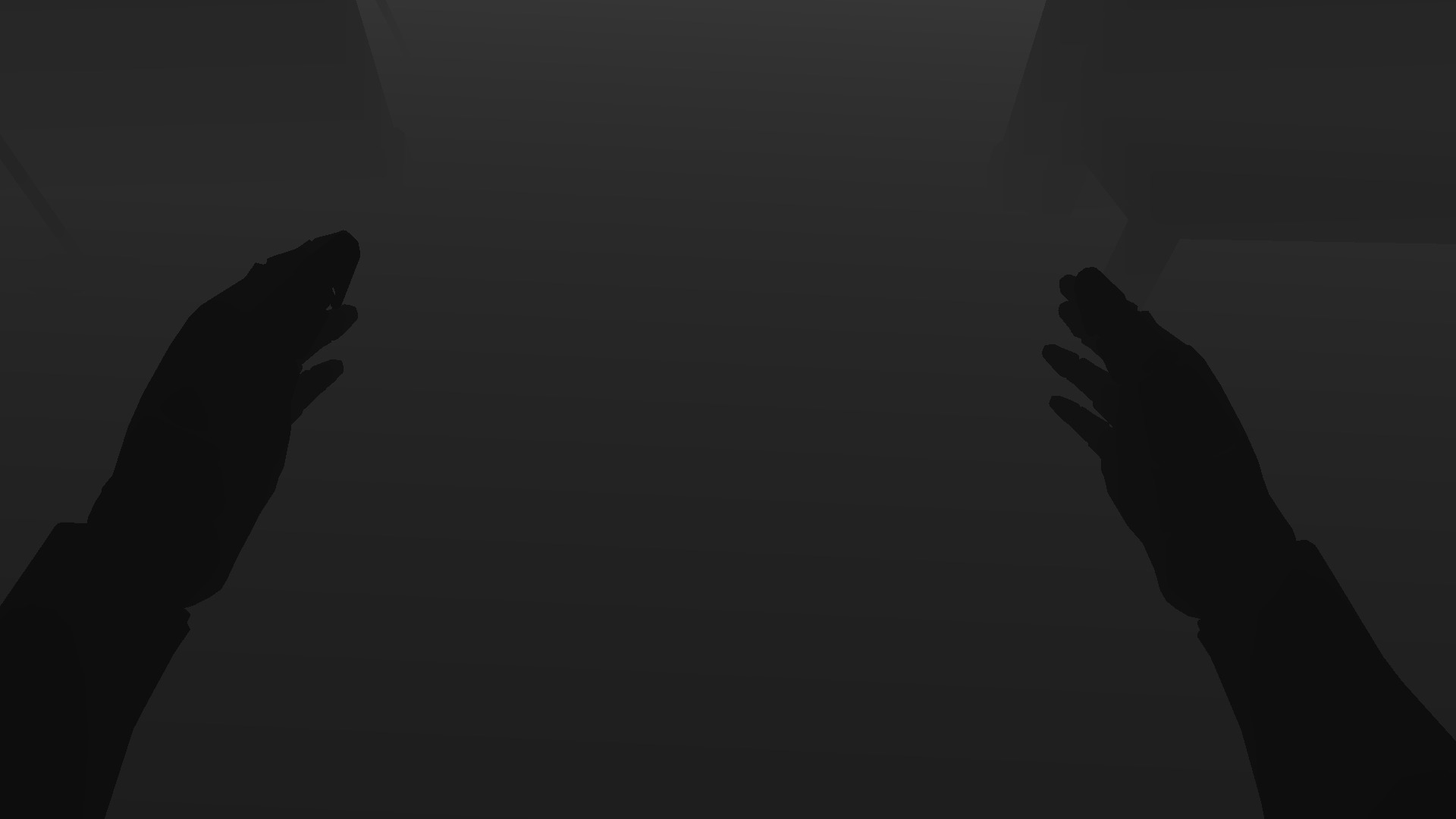}
		\caption{Depth}
	\end{subfigure}
	\hfill
	\\
	\hfill
	\begin{subfigure}{0.49\linewidth}
		\centering
		\includegraphics[width=\linewidth]{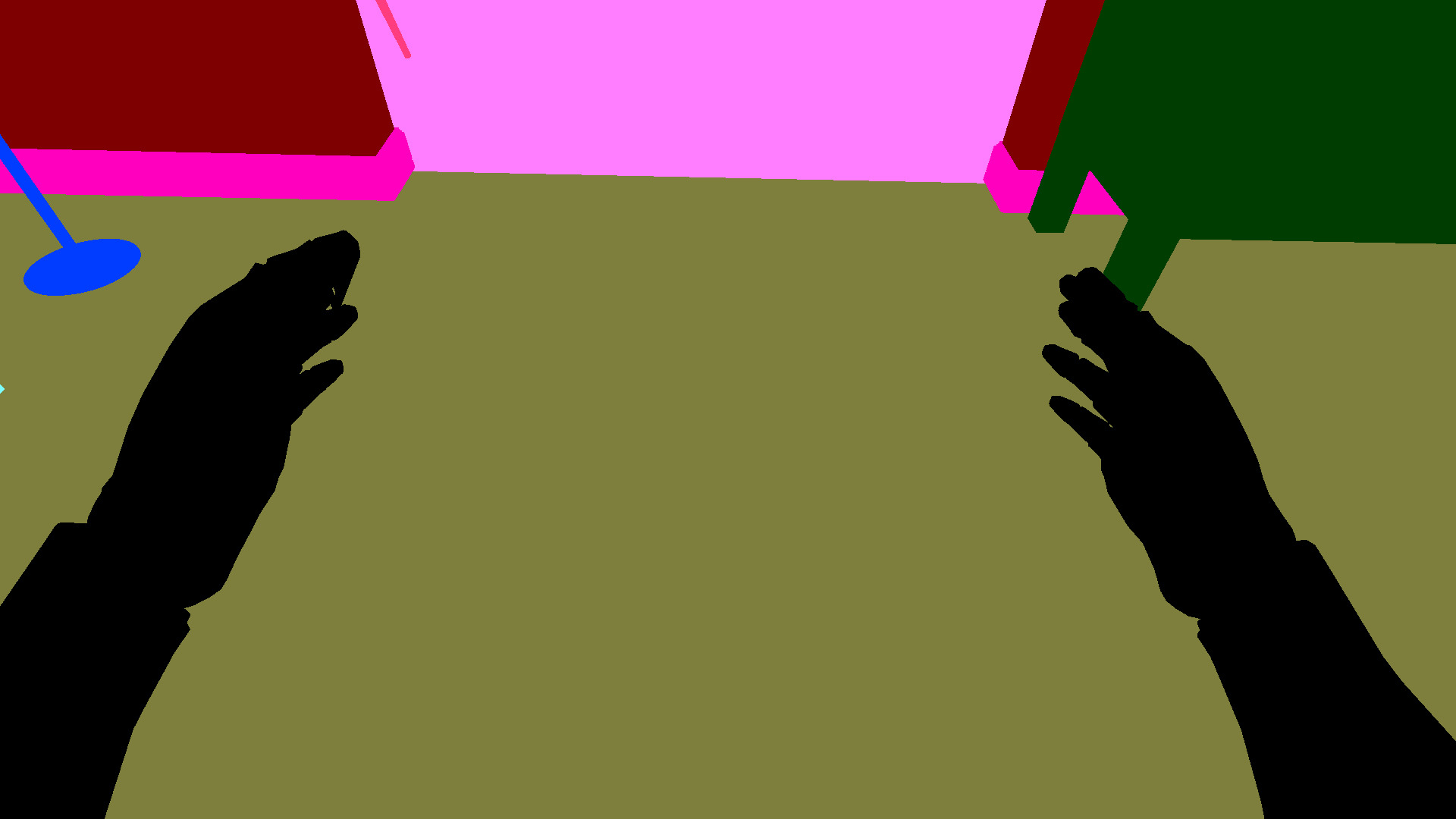}
		\caption{Mask}
	\end{subfigure}
	\hfill
	\begin{subfigure}{0.49\linewidth}
		\centering
		\includegraphics[width=\linewidth]{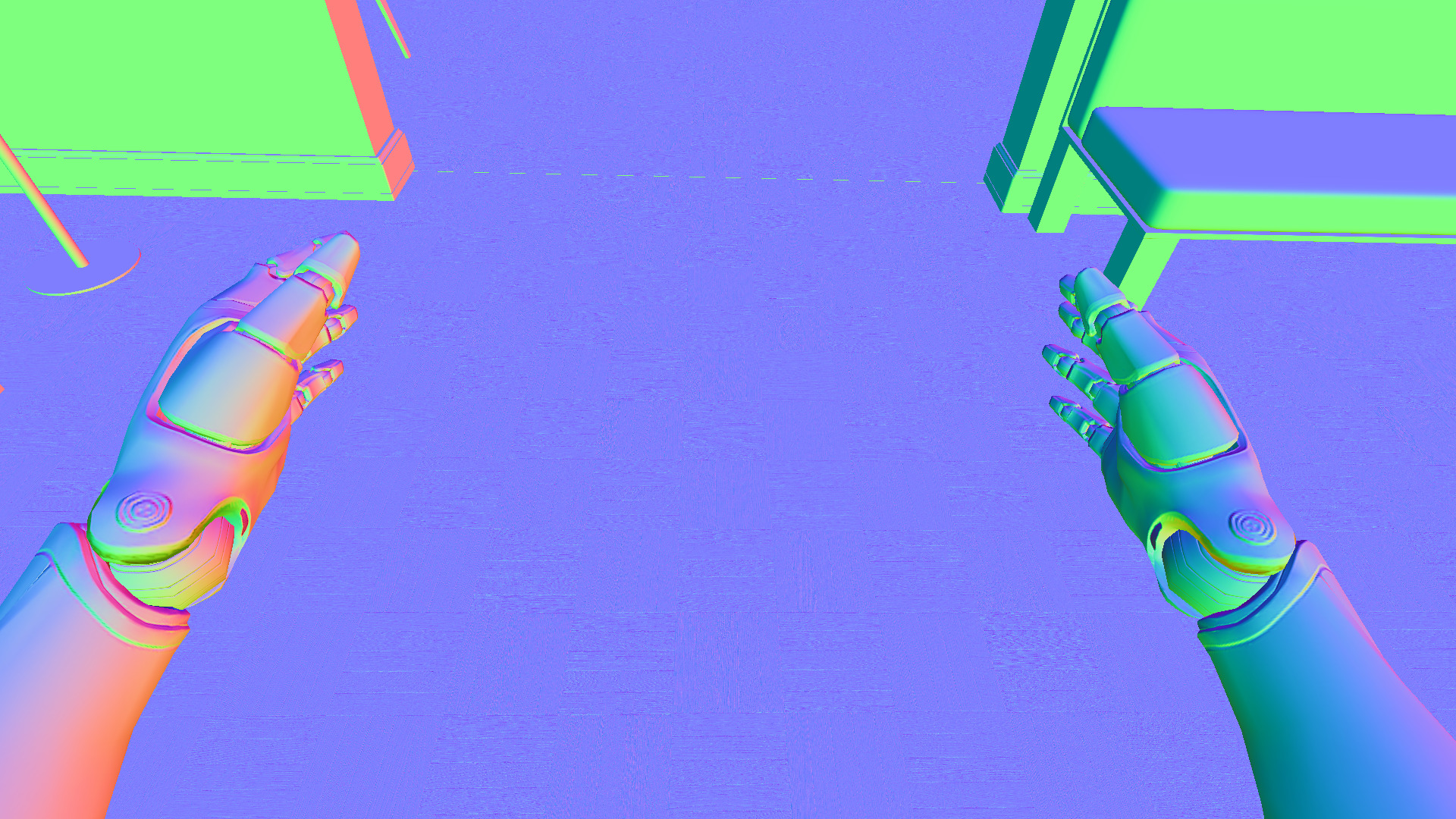}
		\caption{Normals}
	\end{subfigure}
	\caption{Rendering modes cycled by the playback mode.}
	\label{fig:playback}
\end{figure}

This mode disables any physics simulation and interactions (since object and skeleton poses will be hard-coded by the sequence description itself) and then interprets the sequence file to generate all the raw data from it: RGB images, depth maps, instance segmentation masks, and normals. For each frame, the playback mode moves every object and every robot joint to the previously recorded position and sets their rotation. Once everything is positioned, it loops through each camera. For each one of them, the aforementioned rendering modes (RGB, depth, instance, and normals) are switched and the corresponding images are generated as shown in Figure \ref{fig:playback}.

\section{Applications}
\label{sec:applications}

UnrealROX environment has multiple potential application scenarios to generate data for various robotic vision tasks. Traditional algorithms for solving such tasks can take advantage of the data but the main purpose of this environment is providing the ability to generate large-scale datasets. Having the possibility of generating vast amounts of high-quality annotated data, data-driven algorithms such as deep learning models can especially benefit from it to increase their performance, in terms of accuracy, and improve their generalization capabilities in unseen situations during training. The set of tasks and problems that can be addressed using such data ranges from low to high-level ones, covering the whole spectrum of indoor robotics. Some of the most relevant low-level tasks include:

\begin{itemize}
	\item Stereo Depth Estimation: One of the typical ways of obtaining 3D information for robotics is using a pair of displaced cameras used to obtain two different views from the same scene at the same time frame. By comparing both images, a disparity map can be obtained whose values are inversely proportional to the scene depth. Our multi-camera system allows the placement of stereo pairs at configurable baselines so that the environment is able to generate pairs of RGB images, and the corresponding depth, from calibrated cameras.

	\item Monocular Depth Estimation: Another trending way of obtaining 3D information consists of using machine learning methods to infer depth from a single RGB image instead of a stereo pair. From a practical standpoint, it is specially interesting since it requires far less hardware and avoids the need for calibration strategies. Our multi-camera system generates by default depth information for each RGB frame (see Figure \ref{fig:gt_monocular_depth}). For instance, this tool was used to generate data in \cite{Bauer2019}\cite{Bauer2019a}.

\begin{figure}[!htb]
	\centering
	\includegraphics[width=0.325\linewidth]{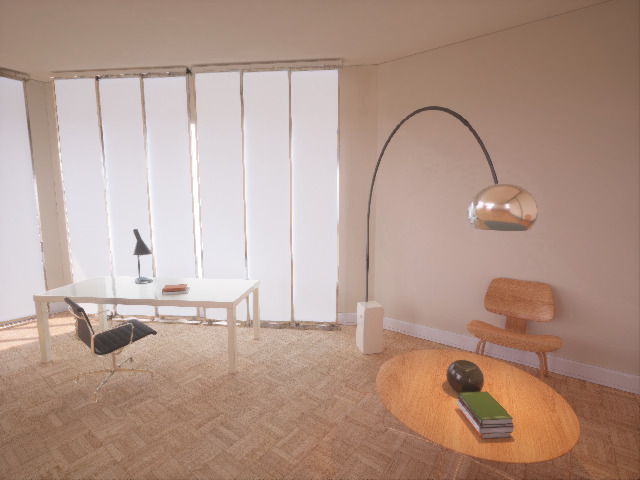}
	\includegraphics[width=0.325\linewidth]{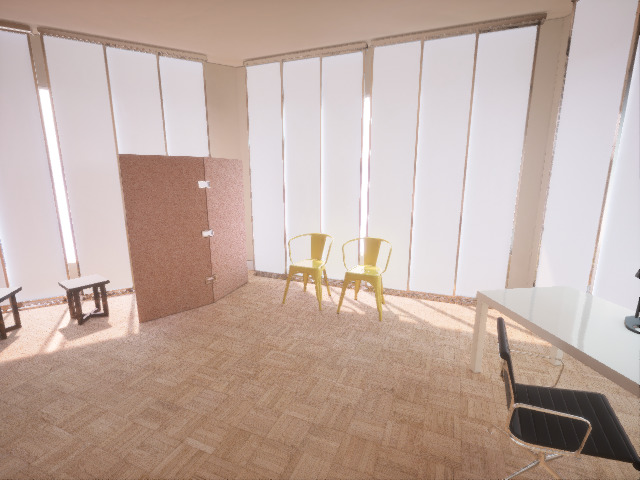}
	\includegraphics[width=0.325\linewidth]{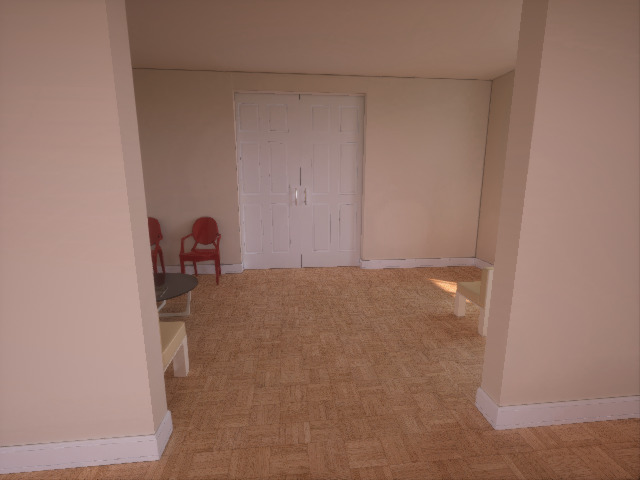}
	\smallskip
	\includegraphics[width=0.325\linewidth]{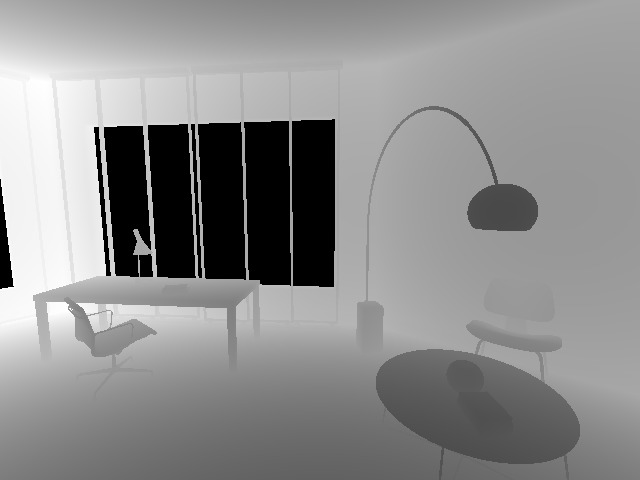}
	\includegraphics[width=0.325\linewidth]{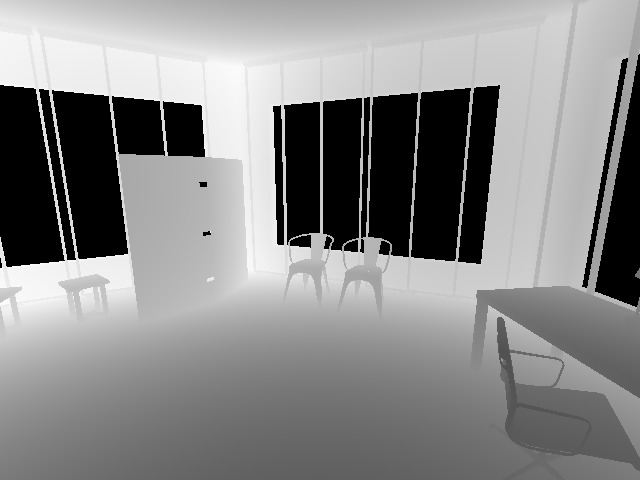}
	\includegraphics[width=0.325\linewidth]{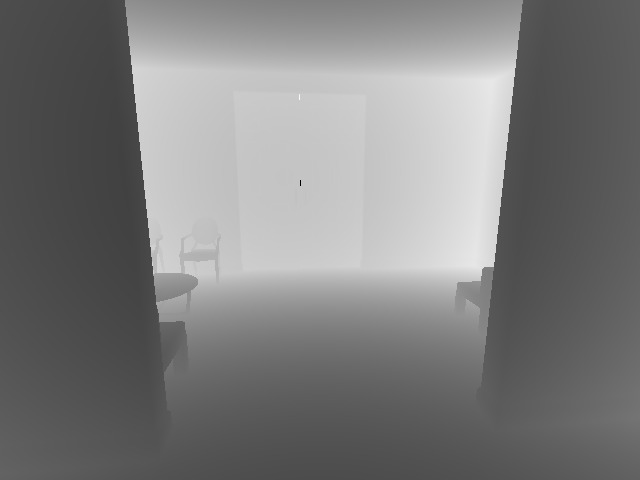}
	\caption{Sample RGB sequence for monocular depth estimation.}
	\label{fig:gt_monocular_depth}
\end{figure}

	\item Object Detection and Pose Estimation: Being able not only to identify which objects are in a given scene frame but also their estimate pose and bounding box is of utmost importance for an indoor robot. Our environment is able to produce 2D and 3D bounding boxes for each frame as ground truth. Furthermore, for each frame of a sequence, the full 6D pose of the objects is annotated too (see Figure \ref{fig:gt_bbox}).

\begin{figure}[!htb]
	\centering
	\includegraphics[width=0.325\linewidth]{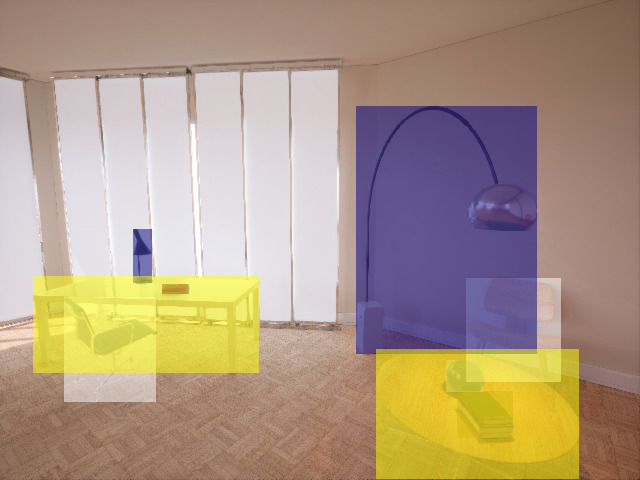}
	\includegraphics[width=0.325\linewidth]{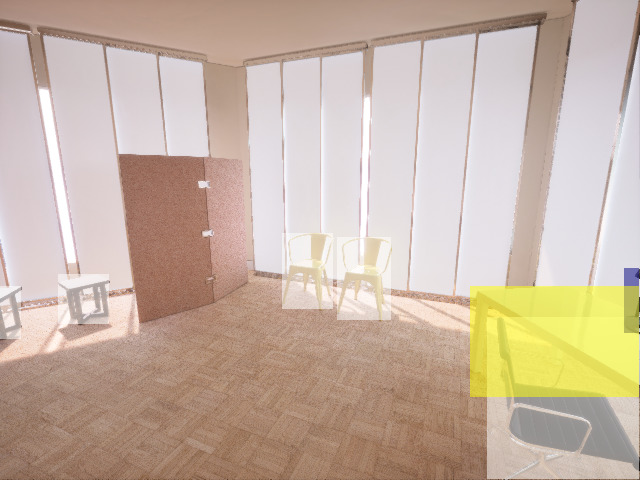}
	\includegraphics[width=0.325\linewidth]{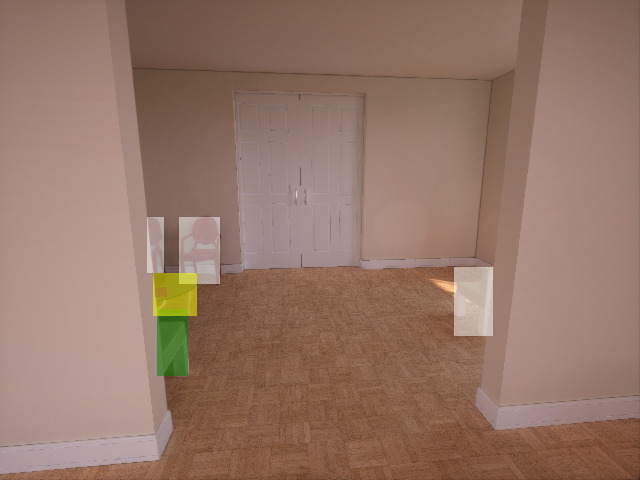}
	\caption{Sample bounding box annotations for the RGB sequence shown in Figure \ref{fig:gt_monocular_depth}.}
	\label{fig:gt_bbox}
\end{figure}

	\item Instance/Class Segmentation: For certain applications, detecting a bounding box for each object is not enough so we need to be able to pinpoint the exact boundaries of the objects. Semantic segmentation of frames provides per-pixel labels that indicate to which instance or class does a particular pixel belong. Our environment generates 2D (per-pixel) and 3D (per-point) labels for instance and class segmentation (see Figure \ref{fig:gt_mask}).

\begin{figure}[!htb]
	\centering
	\includegraphics[width=0.325\linewidth]{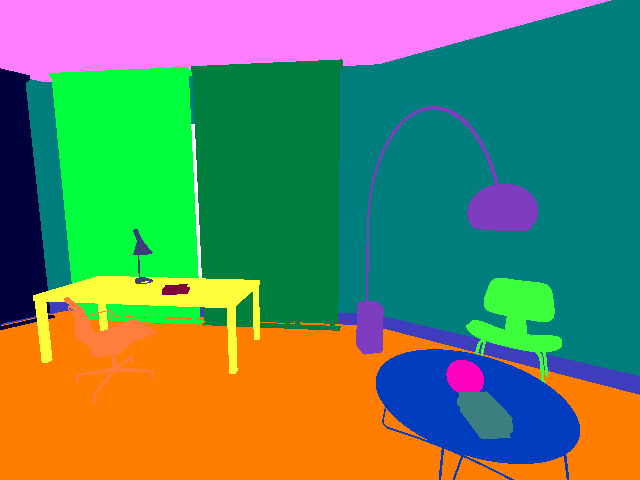}
	\includegraphics[width=0.325\linewidth]{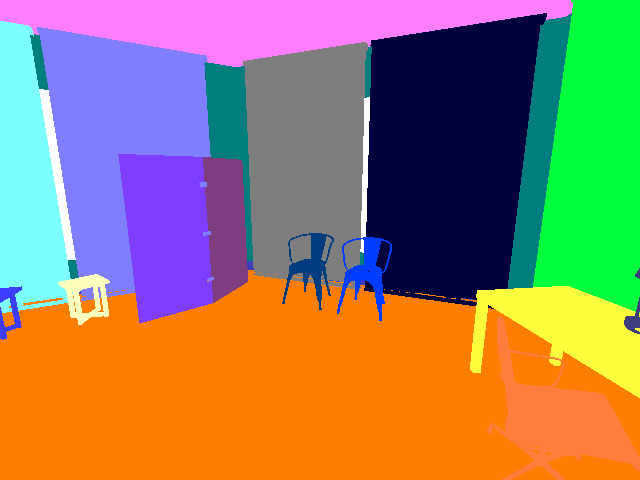}
	\includegraphics[width=0.325\linewidth]{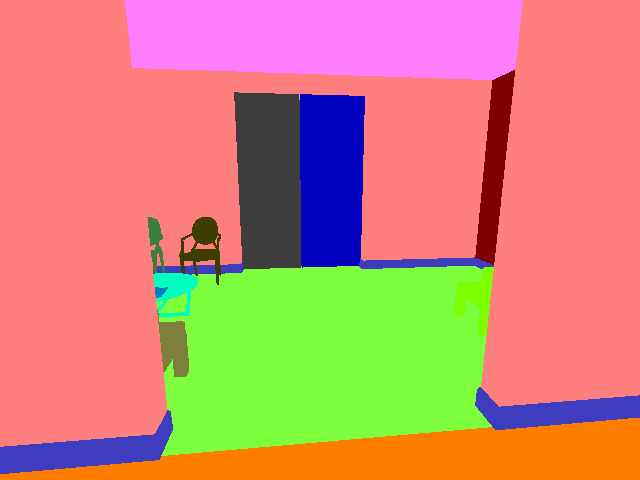}
	\caption{Sample instance segmentation sequence for the RGB images shown in Figure \ref{fig:gt_monocular_depth}.}
	\label{fig:gt_mask}
\end{figure}

	\item Normal Estimation: Estimating the normals of a given surface is an important previous step for many other tasks. For instance, certain algorithms require normal information in a point cloud to extract grasping points for a robot. UnrealROX provides per-pixel normal information.
\end{itemize}

That low-level data enables other higher-level tasks that either make use of the output of those systems or take the low-level data as input or even both possibilities:

\begin{itemize}
	\item Hand Pose Estimation: Estimating the 6D pose of each joint of the hands provides useful information for various higher-level tasks such as gesture detection, grasping or collaboration with other robots. We provide per-frame 6D pose annotations for each joint of the robot's hands.
	\item Visual Grasping and Dexterous Manipulation: Grasping objects and manipulating them while grasped with one or both hands is a high-level task which can be solved using information from various sources (RGB images, depth maps, segmentation masks, normal maps, and joint estimates to name a few). In our case, we provide sequences in which the robot interacts with objects to displace, grab, and manipulate them so that grasping algorithms can take advantage of such sequences recorded from various points of view (see Figure \ref{fig:gt_interaction}).
\begin{figure}[!htb]
	\centering
	\includegraphics[width=0.325\linewidth]{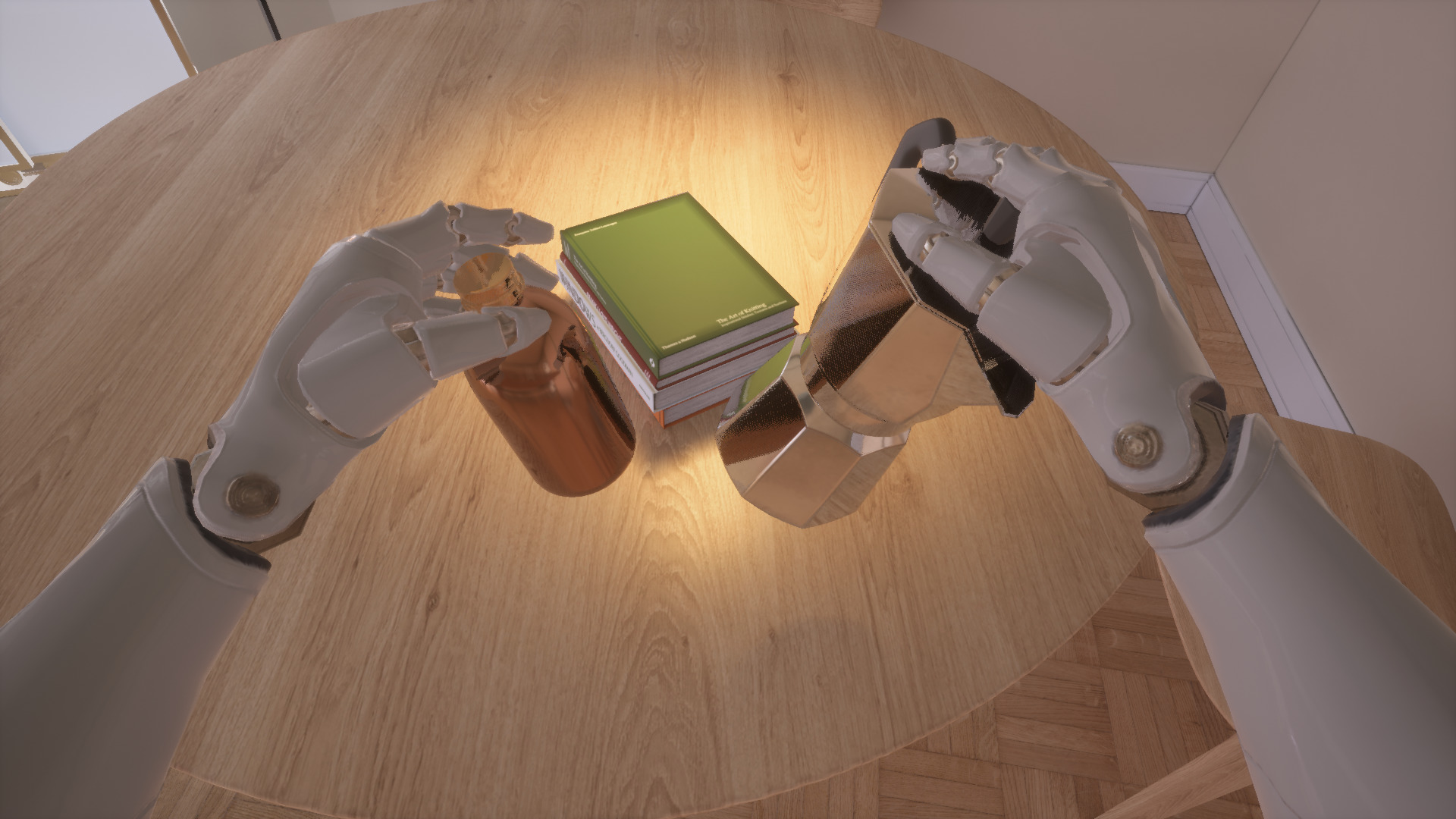}
	\includegraphics[width=0.325\linewidth]{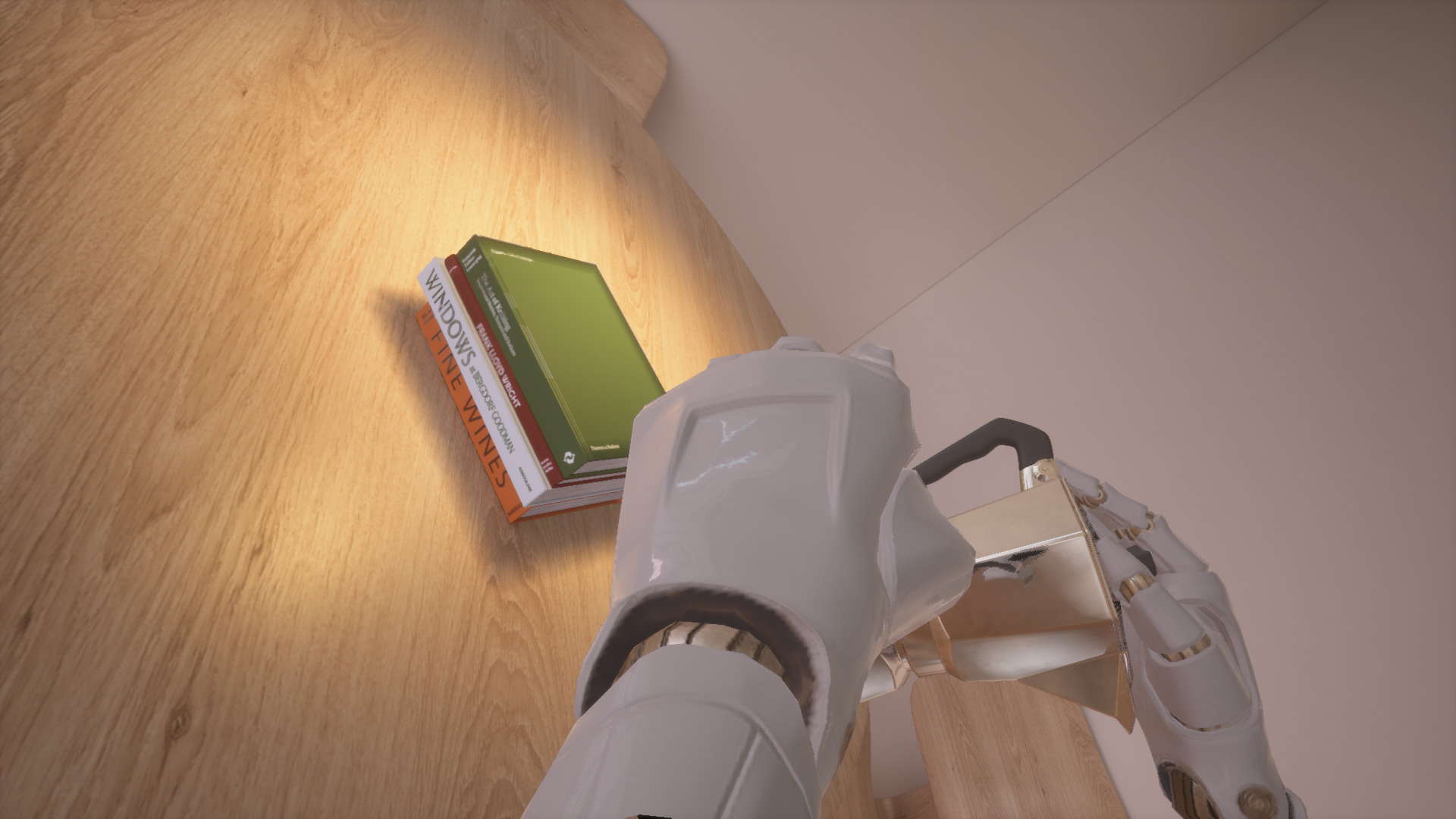}
	\includegraphics[width=0.325\linewidth]{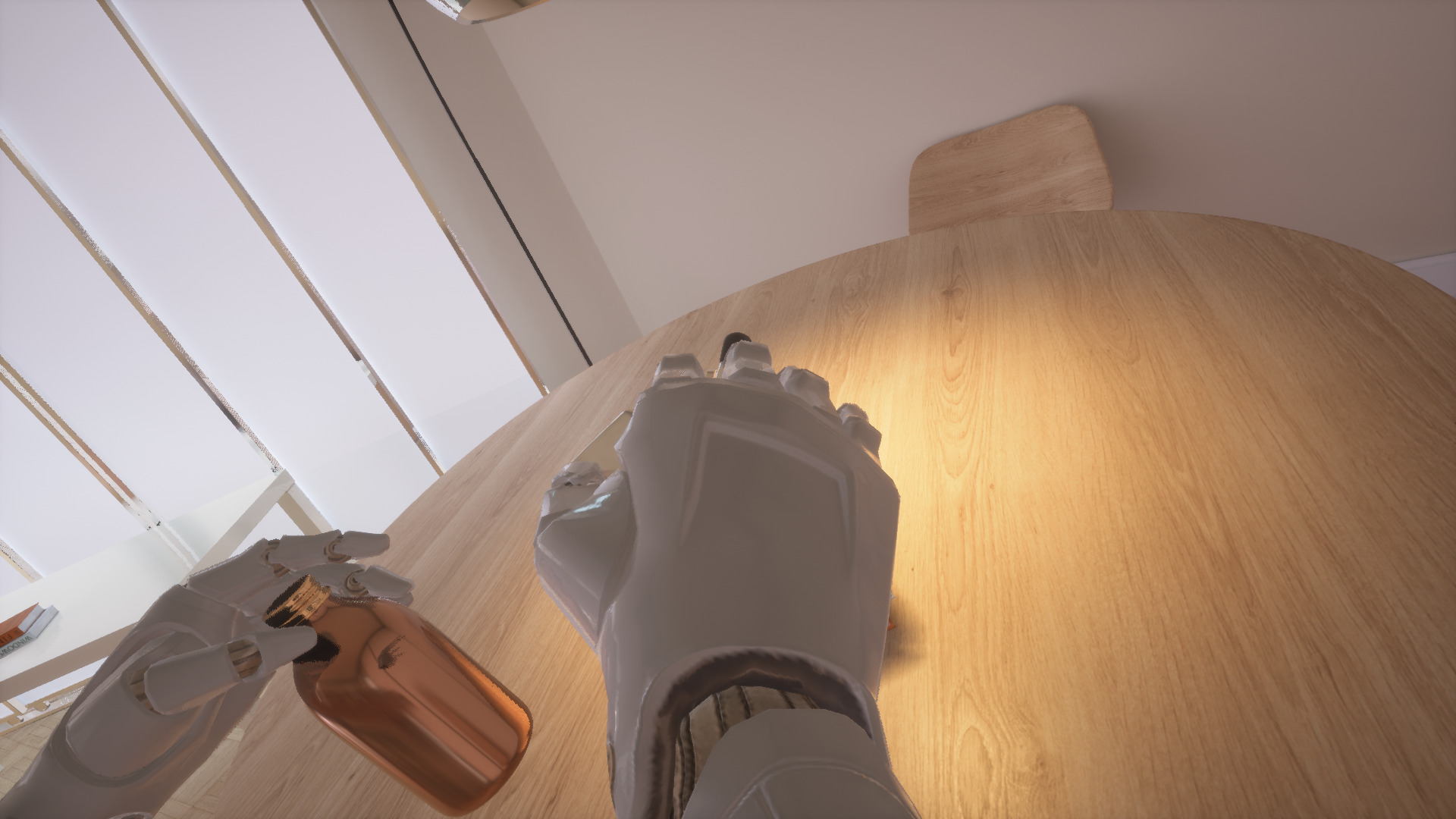}
	\smallskip
	\includegraphics[width=0.325\linewidth]{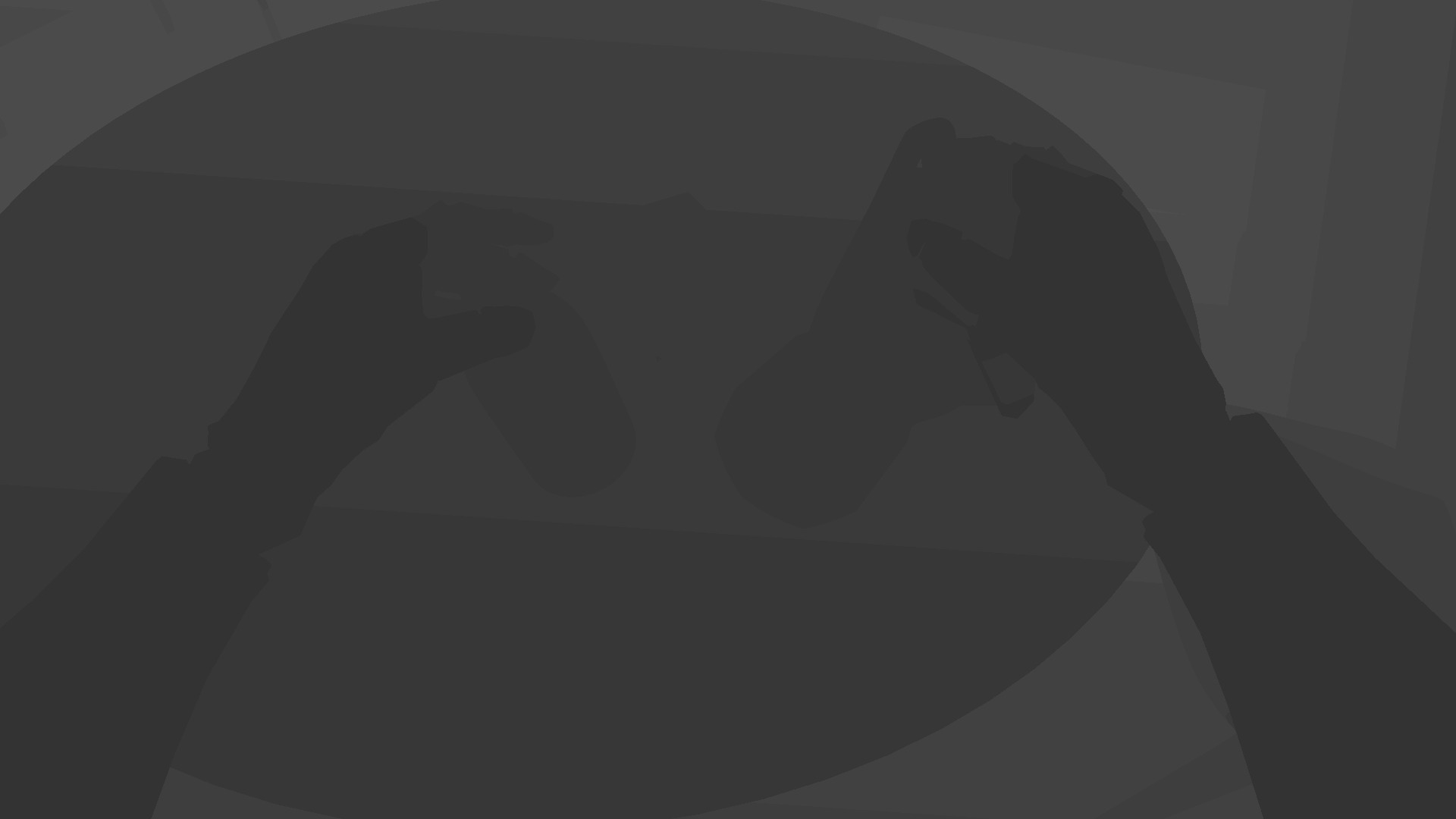}
	\includegraphics[width=0.325\linewidth]{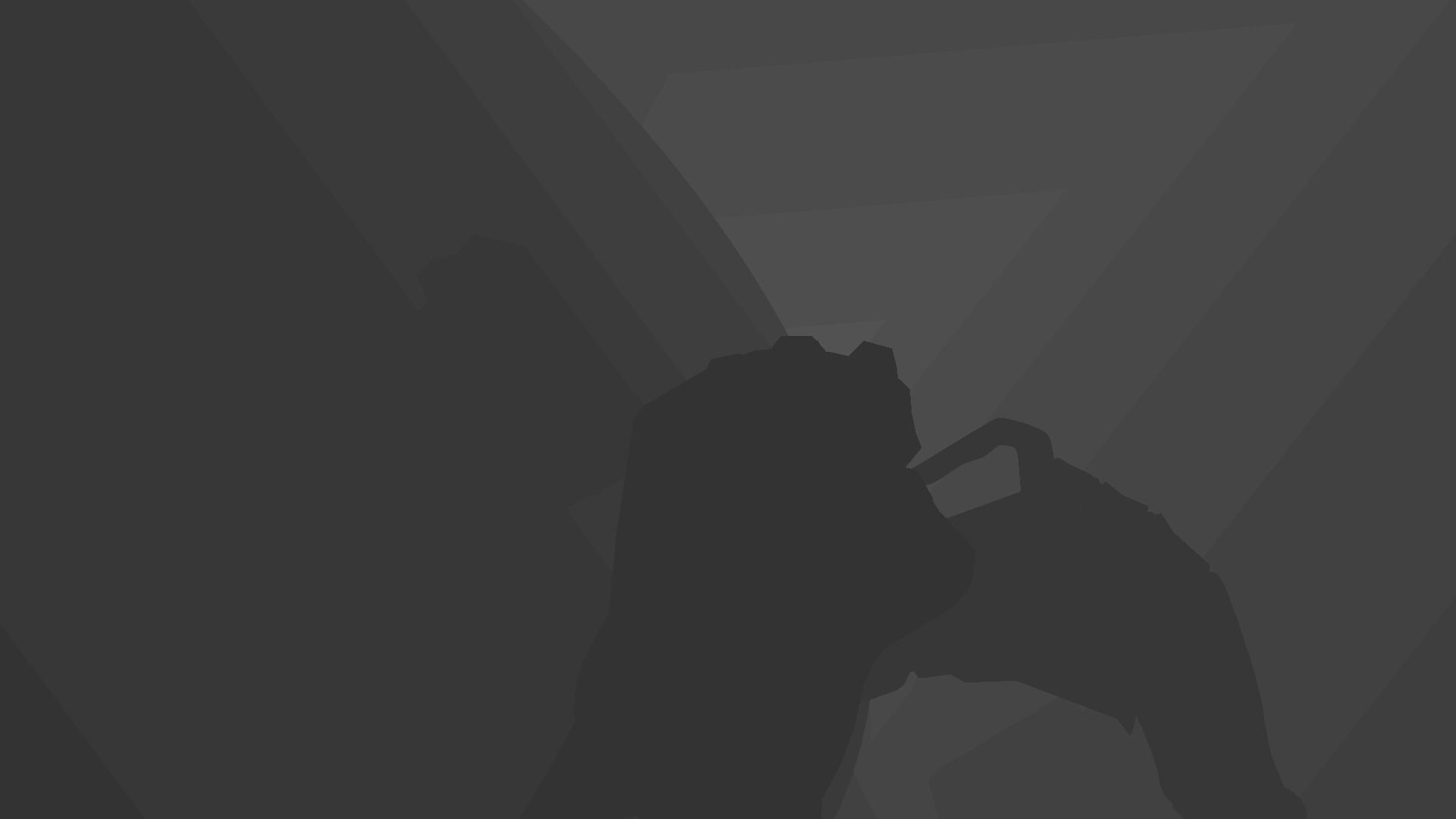}
	\includegraphics[width=0.325\linewidth]{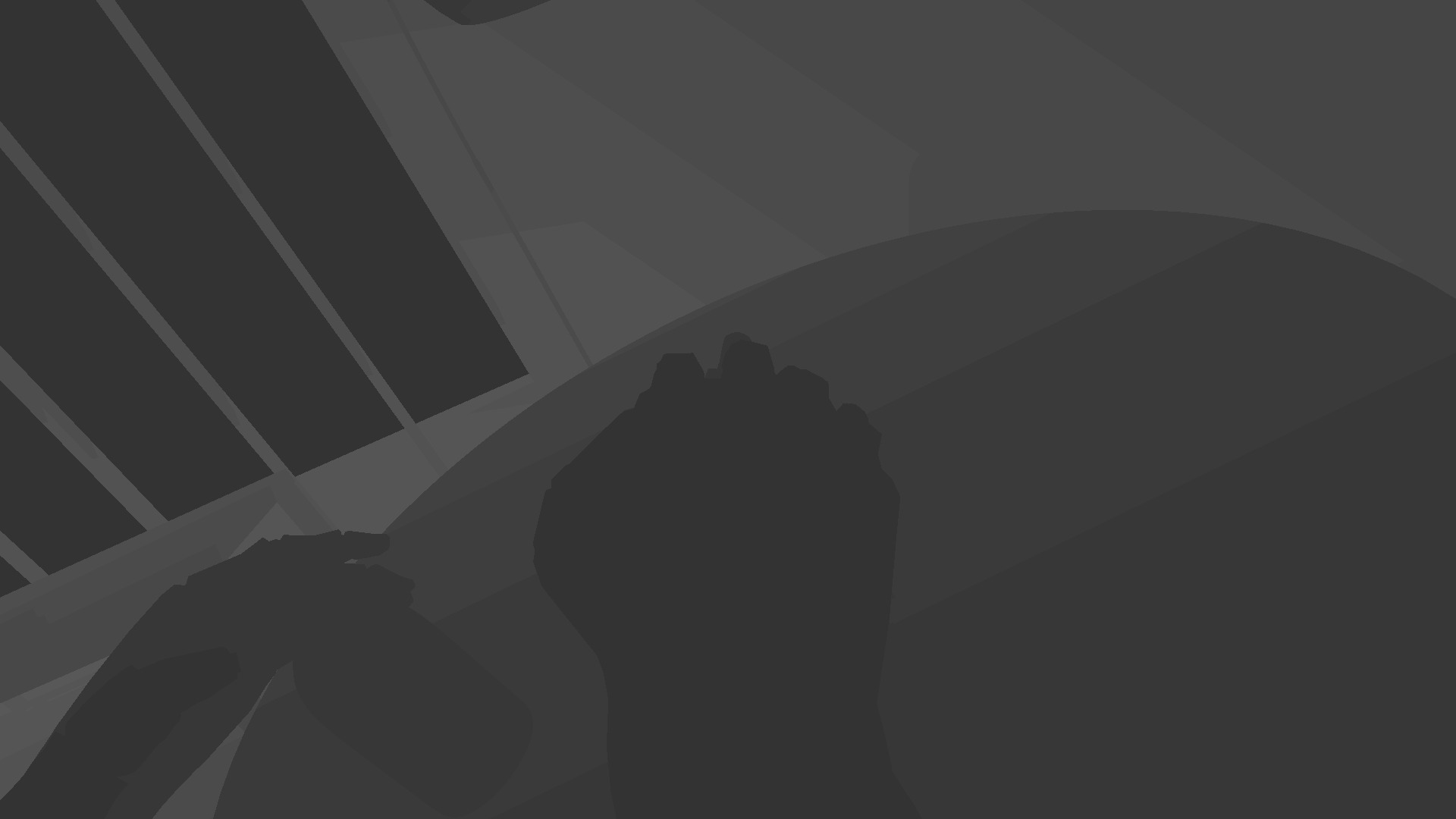}
	\smallskip
	\includegraphics[width=0.325\linewidth]{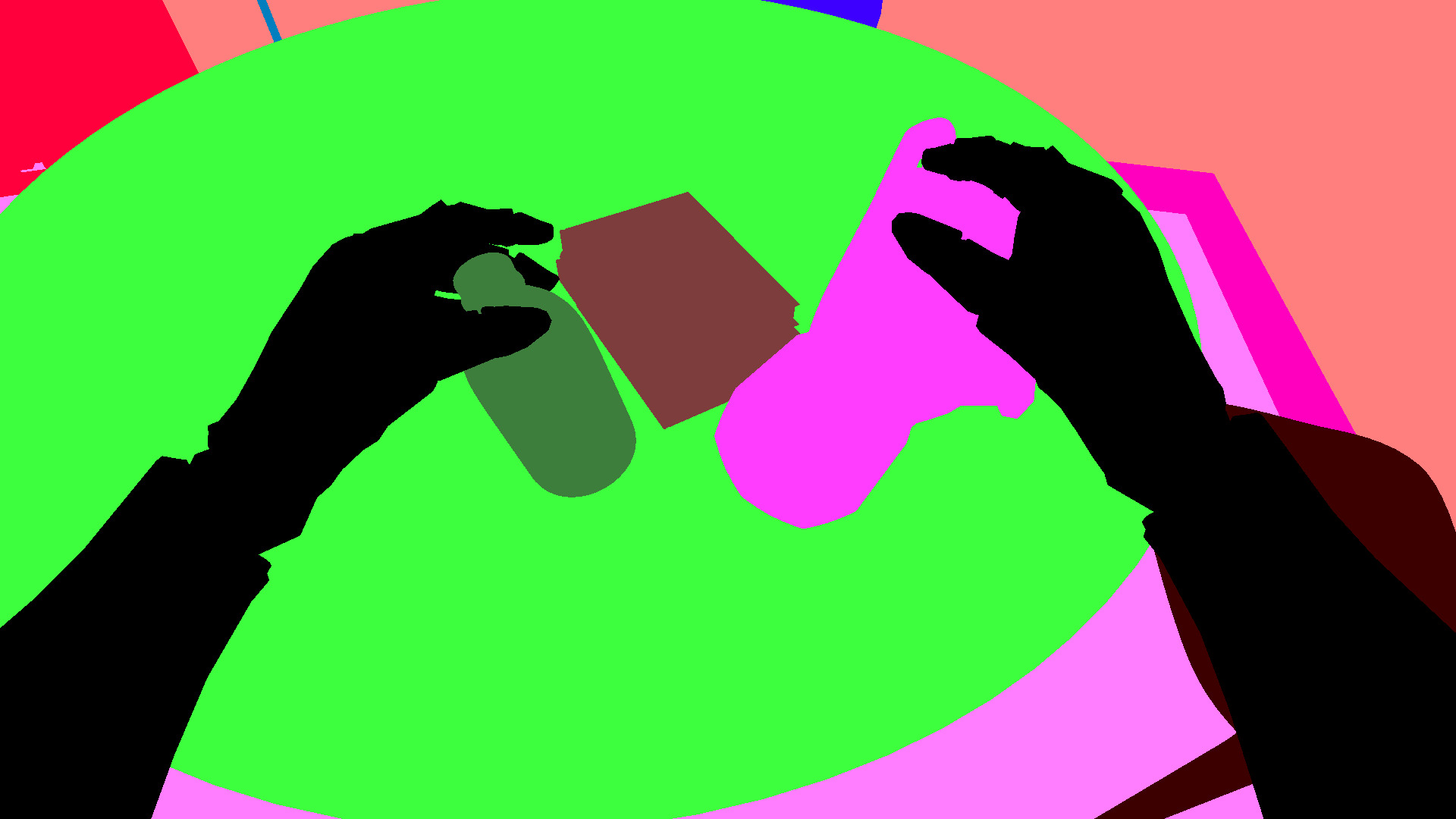}
	\includegraphics[width=0.325\linewidth]{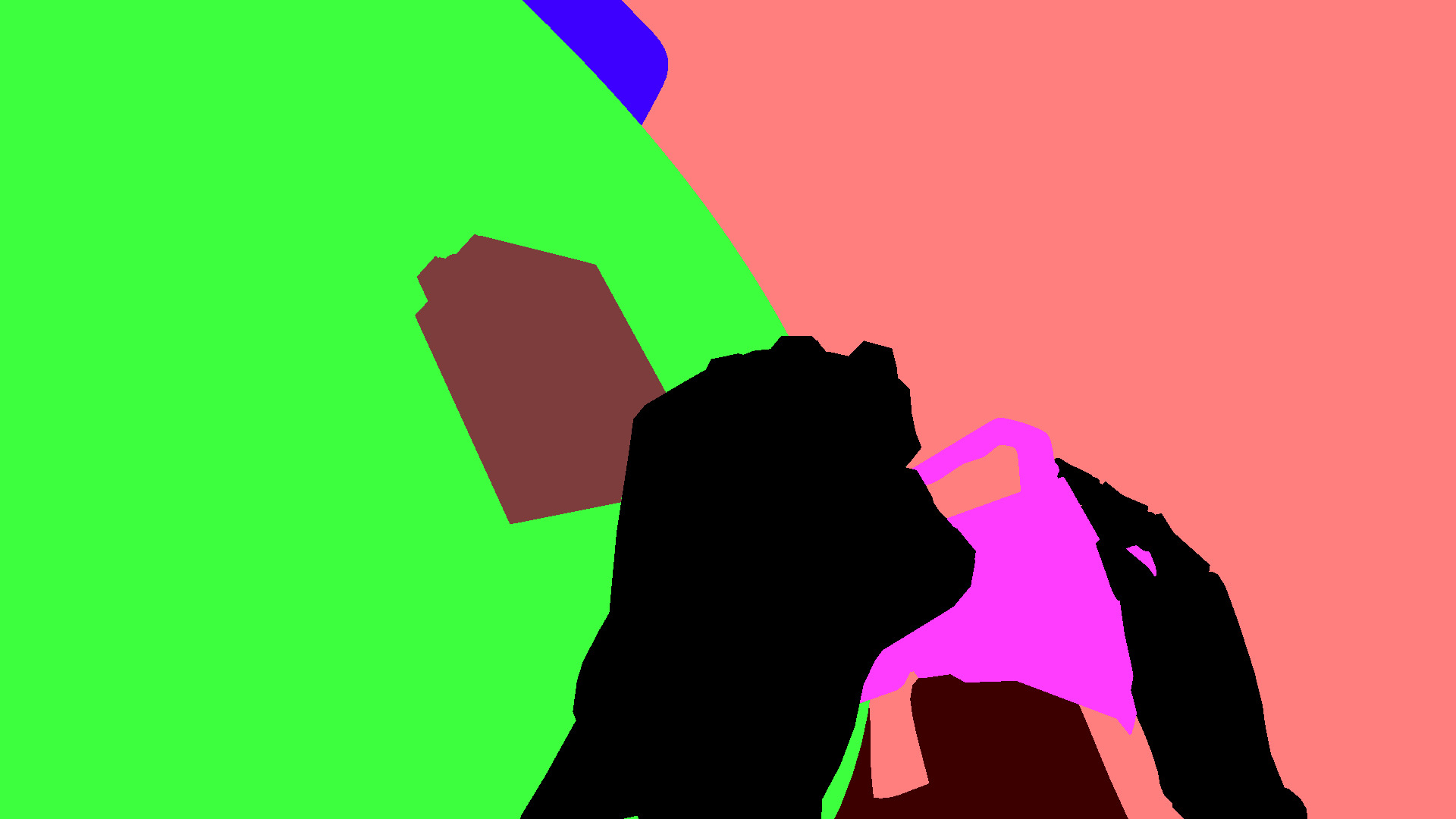}
	\includegraphics[width=0.325\linewidth]{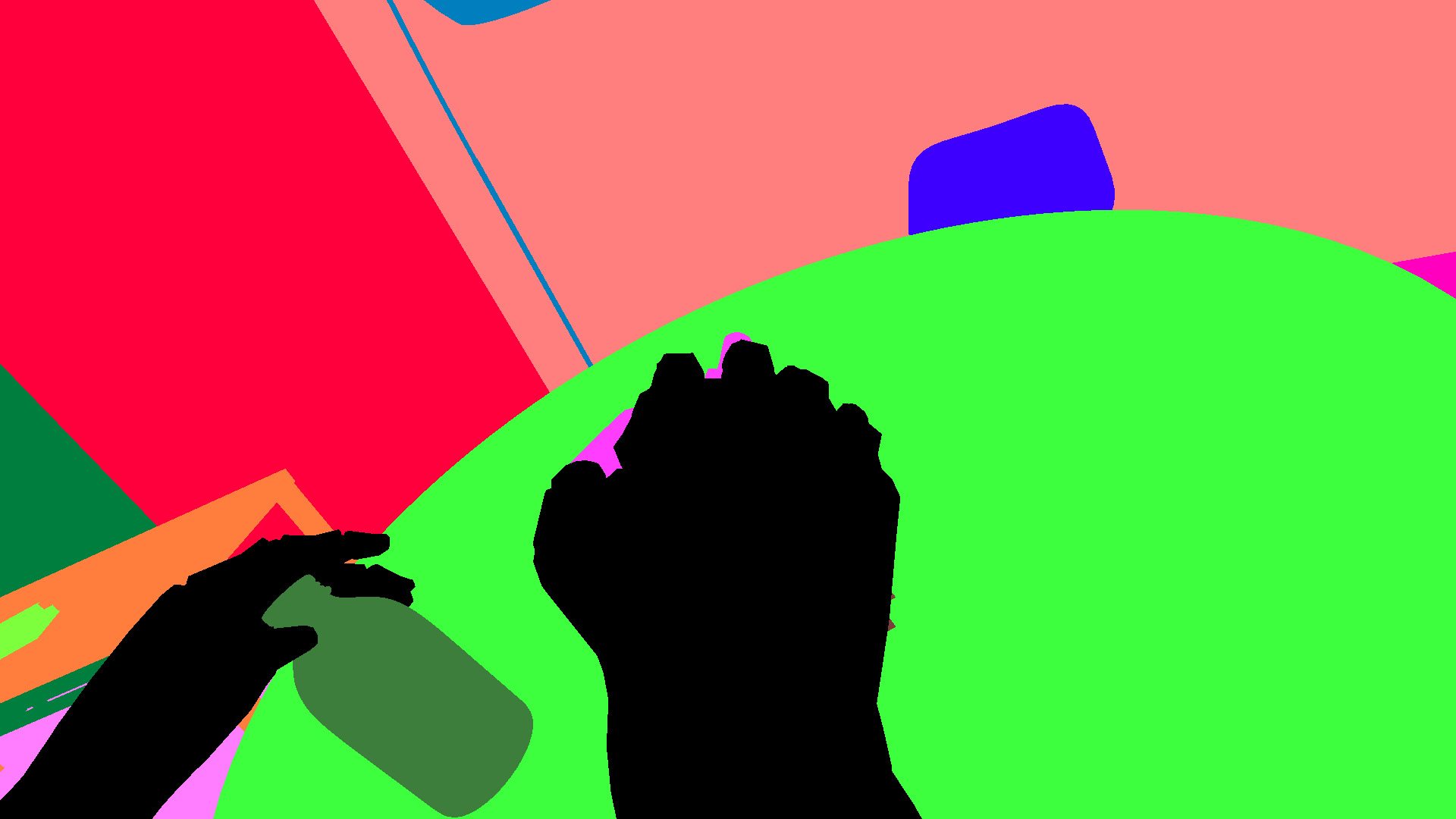}
	\smallskip
	\includegraphics[width=0.325\linewidth]{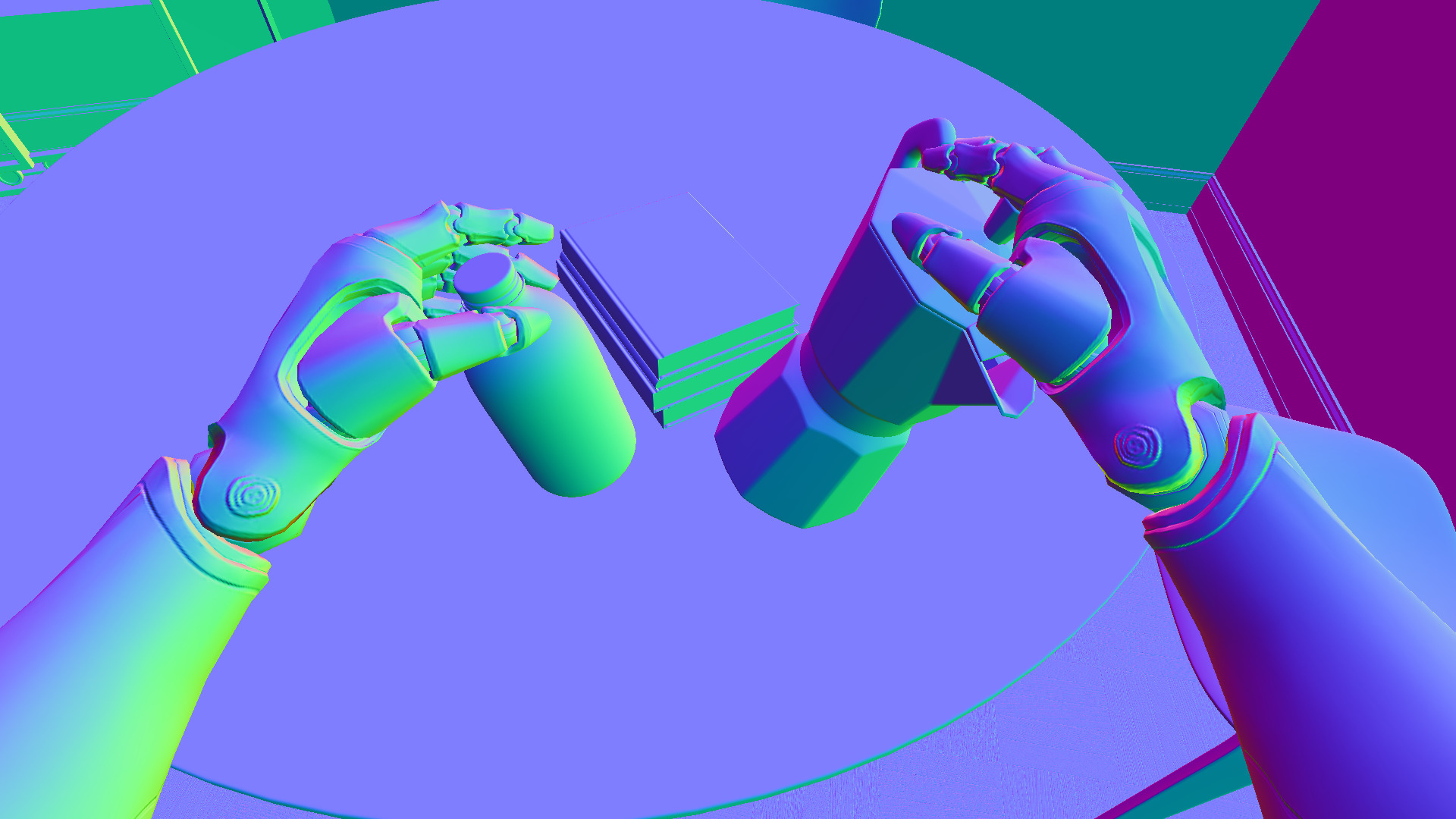}
	\includegraphics[width=0.325\linewidth]{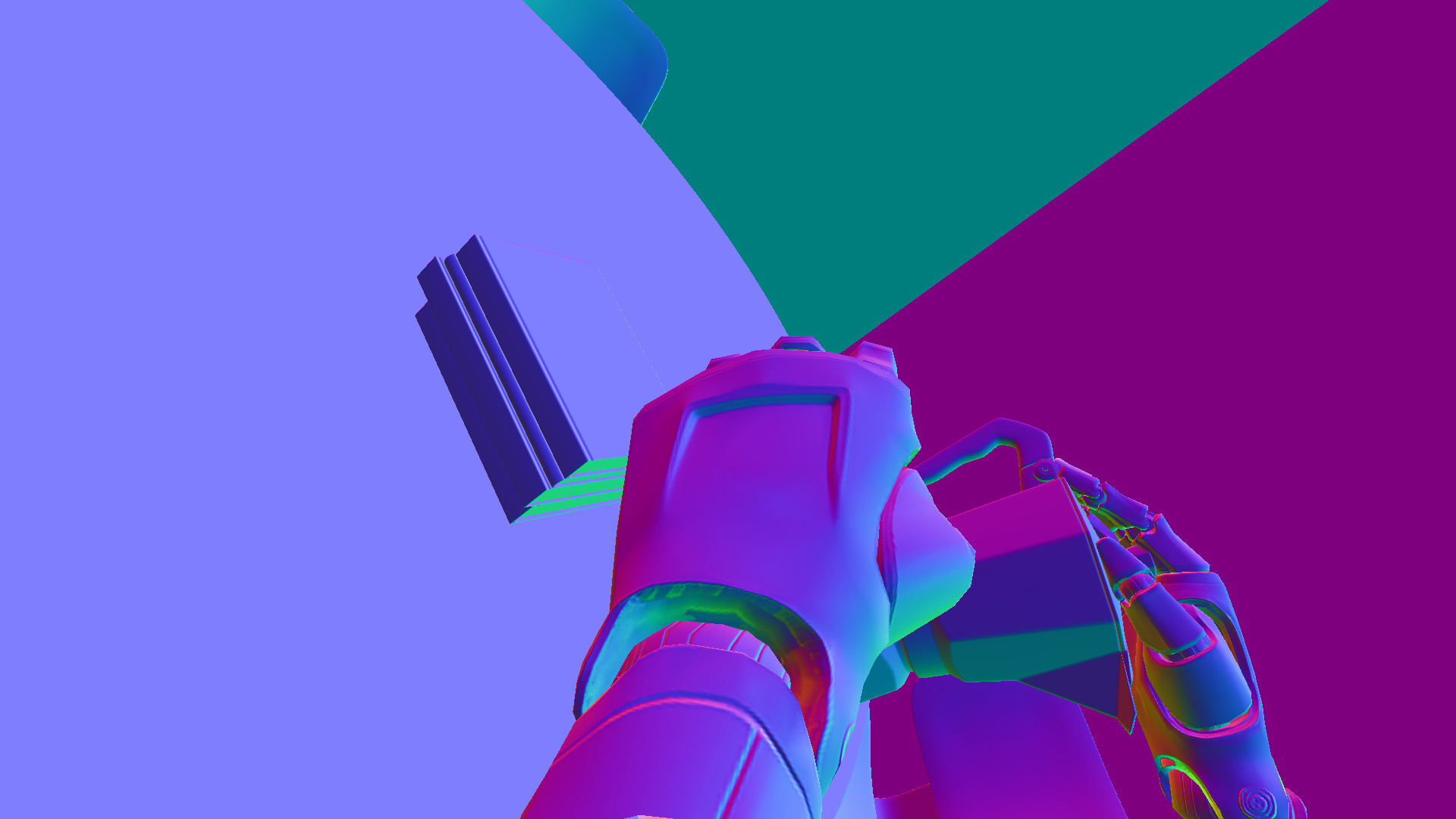}
	\includegraphics[width=0.325\linewidth]{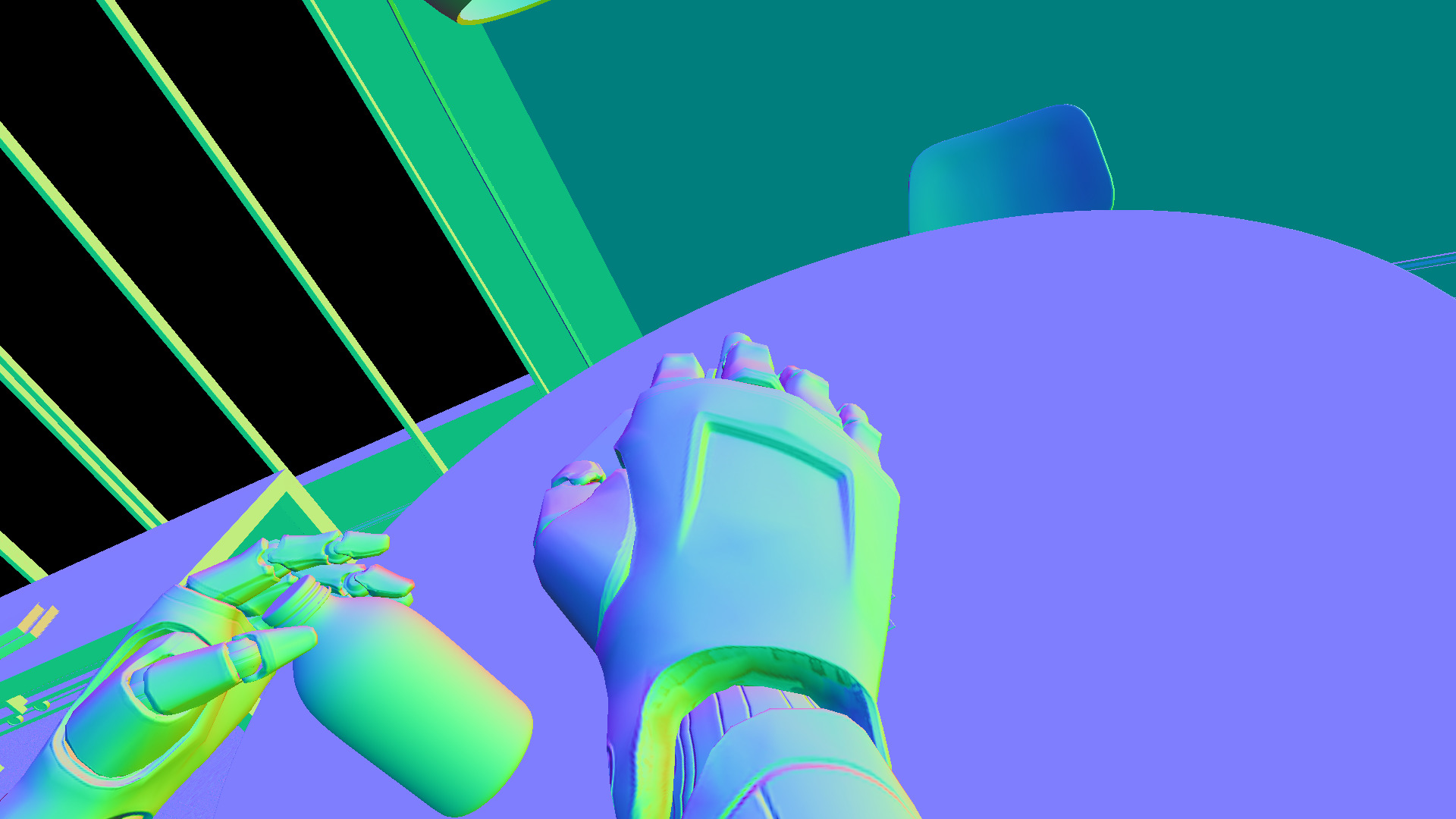}
	\caption{Sample hands interaction sequence and its associated data (from top to bottom: RGB, depth, instance masks, and normals).}
	\label{fig:gt_interaction}
\end{figure}

	\item Robot Pose Estimation: As well as providing 6D pose for hand joints, our environment also provides such information for all the joints of a robot on a per-frame basis. This allows training and testing body pose estimation algorithms which can be extremely useful in indoor environments to analyze behaviors and even collaborate with other robots too. To that end, we equipped our multi-camera system with the capability of adding room cameras that capture full bodies typical from assisted indoor living (see Figure \ref{fig:gt_outside}).

\begin{figure}[!htb]
	\centering
	\includegraphics[width=0.325\linewidth]{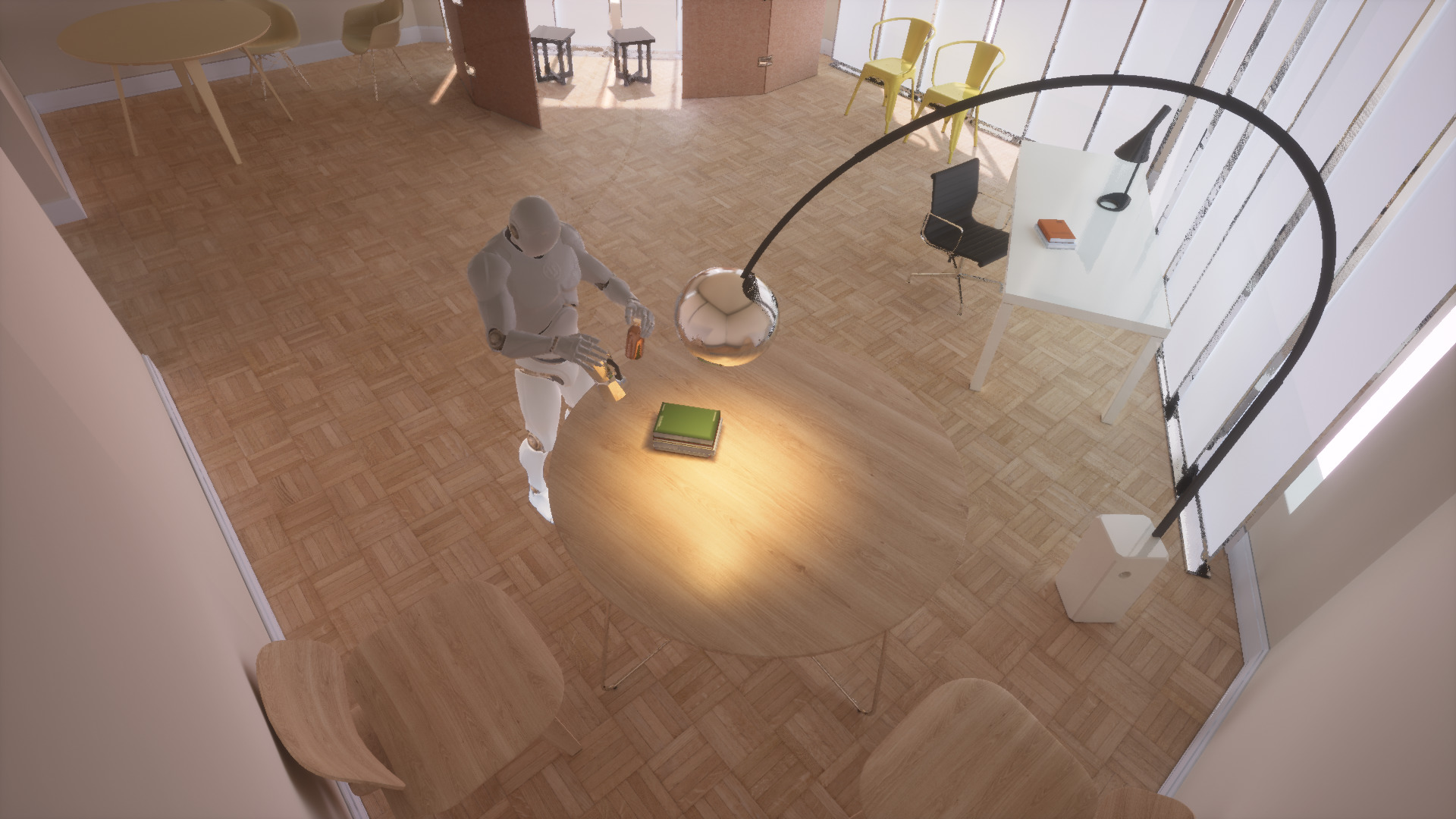}
	\includegraphics[width=0.325\linewidth]{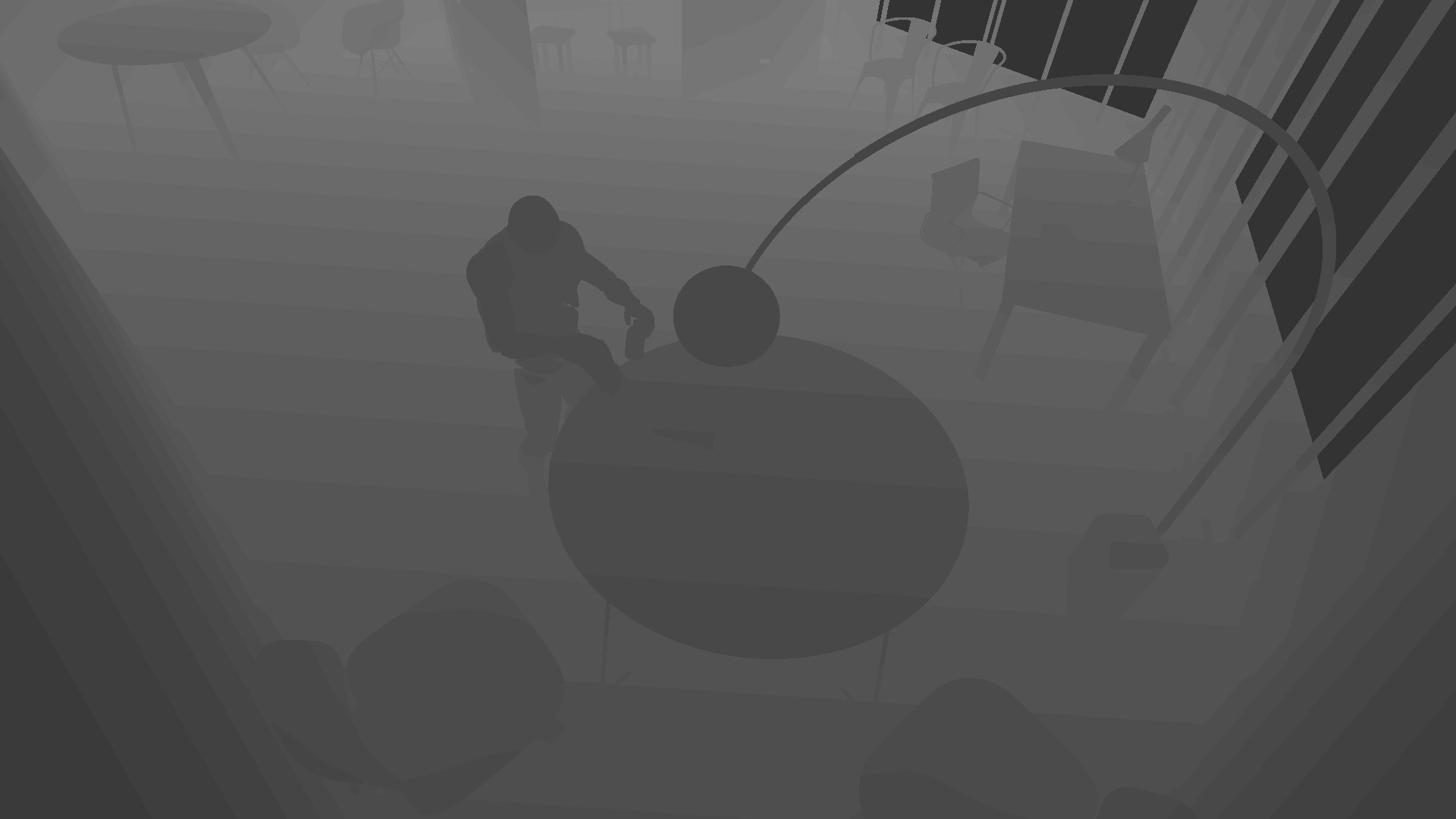}
	\includegraphics[width=0.325\linewidth]{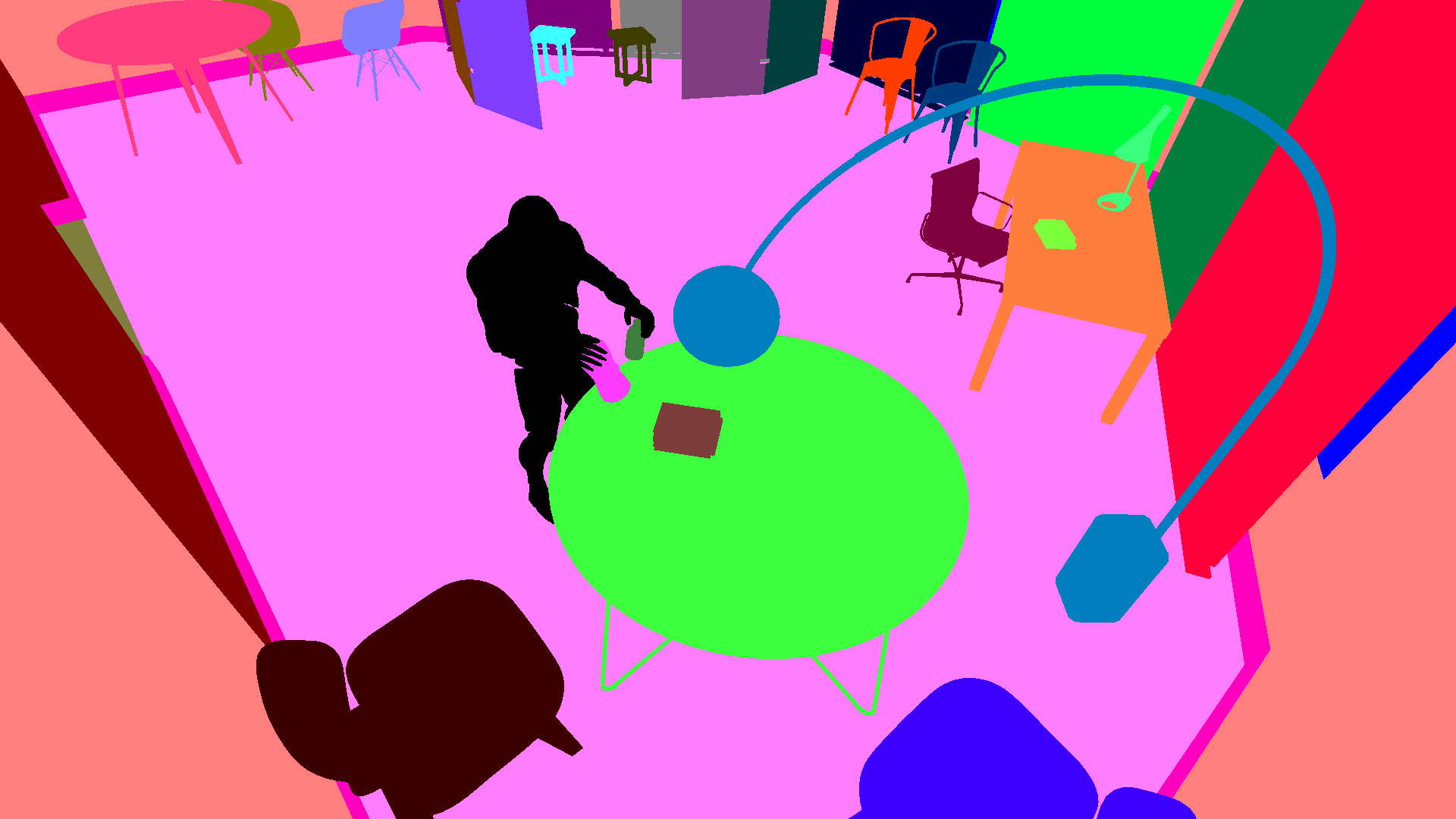}
	\caption{Sample data from an external point of view in the room with the corresponding images (RGB, depth, and instance masks).}
	\label{fig:gt_outside}
\end{figure}

	\item Obstacle Avoidance and Navigation: By leveraging various types of low-level information such as RGB images, depth maps, bounding boxes, and semantic segmentation, robots can learn to avoid obstacles (by detecting objects and estimating their distance) and even navigate in indoor environments (by building a map to localize themselves in the indoor scene while avoiding objects and walls and being able to reason semantically to move intelligently). 
\end{itemize}

As we can observe, UnrealROX is able to generate data for a significantly wide range of robotic vision applications. Most of them orbit around indoor robotics, although some of them might as well be applied to outdoor situations. In general, their purpose can be grouped into the more general application of \ac{AAL} due to the inherent goal of achieving a robotic system able to operate intelligently in an indoor scenario in an autonomous way to provide support at various social tasks such as in-house rehabilitation, elder care, or even disabled assistance.

\section{Experiments}

\label{sec:experiments}

In the previous section we showed multiple potential applications to which our data generator and the corresponding ground truth could be applied to train machine learning systems. In this section, we selected two of those applications to experiment with them in order to prove the effectiveness of our approach. Those two representative problems are: monocular depth estimation from RGB images and 6D object pose estimation.

\subsection{Monocular Depth Estimation}

As we already mentioned, estimating depth from 2D RGB images is an useful technique for many other higher-level applications such as scene reconstruction, object detection, and semantic segmentation. The problem can be formulated as follows: given a colored RGB image from any camera, the goal is to predict a dense depth map for each pixel as accurately as possible \cite{Bhoi2019}.

\begin{figure}[!htb]
    \centering
    \includegraphics[width=0.24\linewidth]{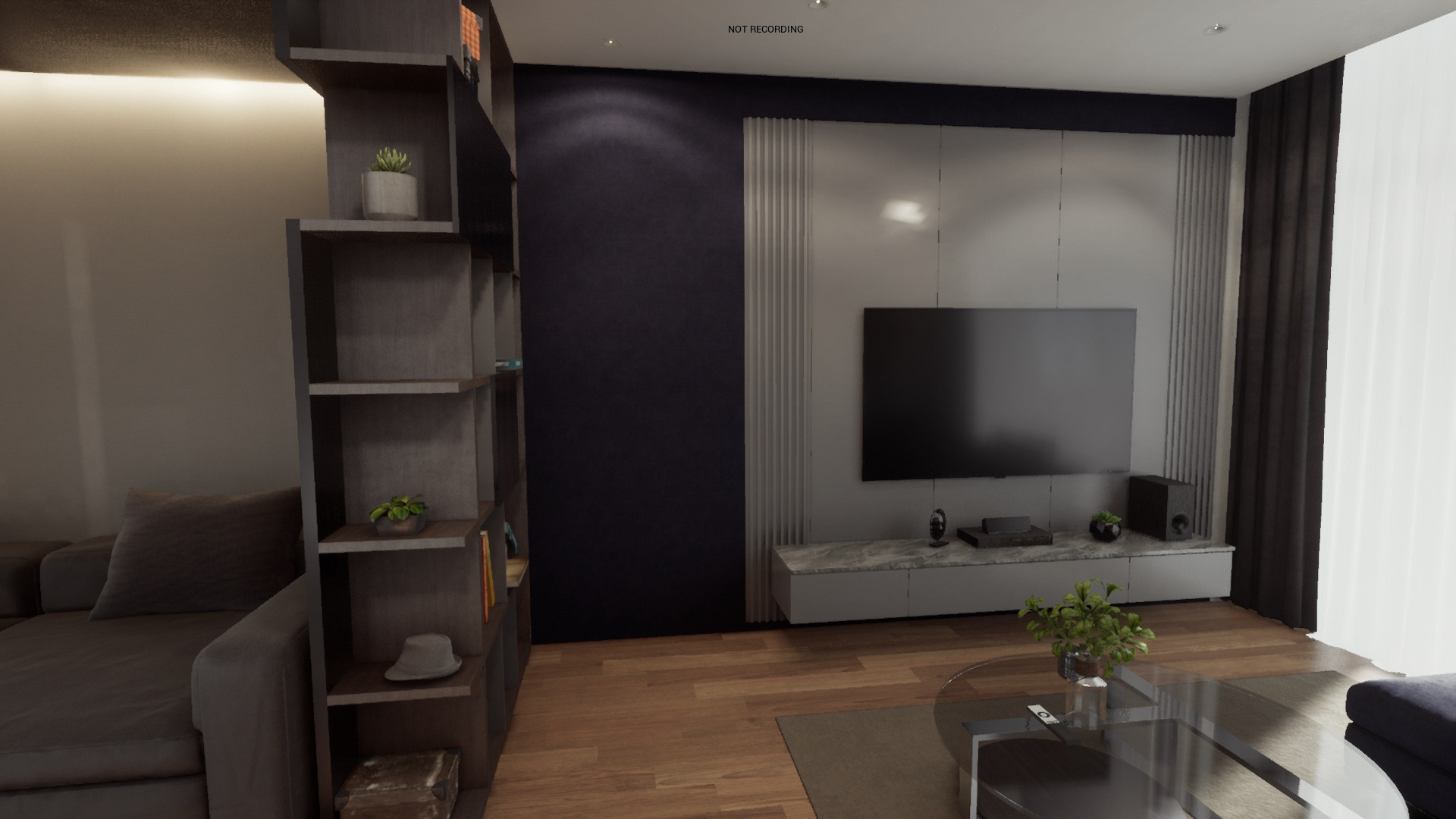}
    \includegraphics[width=0.24\linewidth]{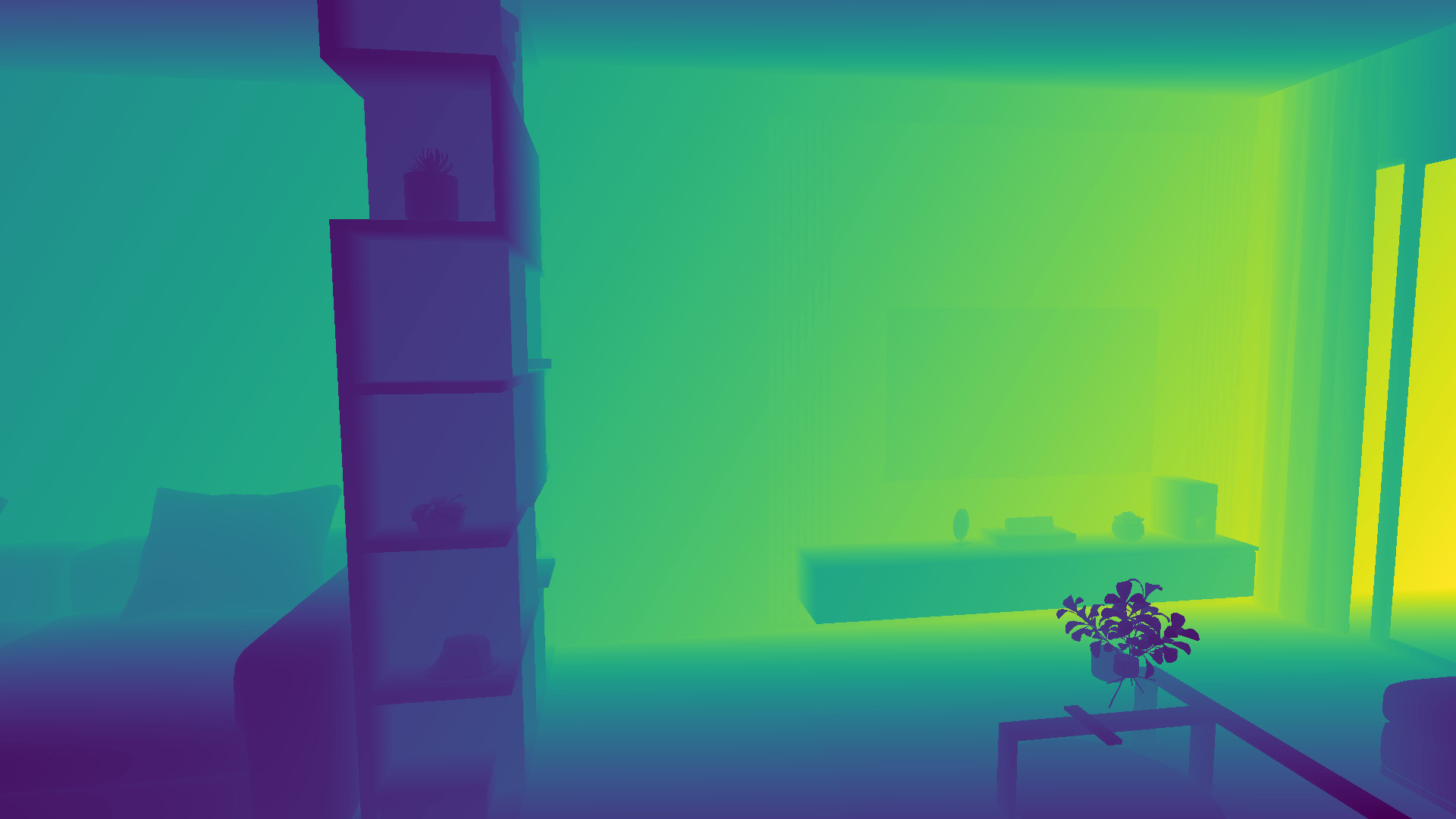}
    \includegraphics[width=0.24\linewidth]{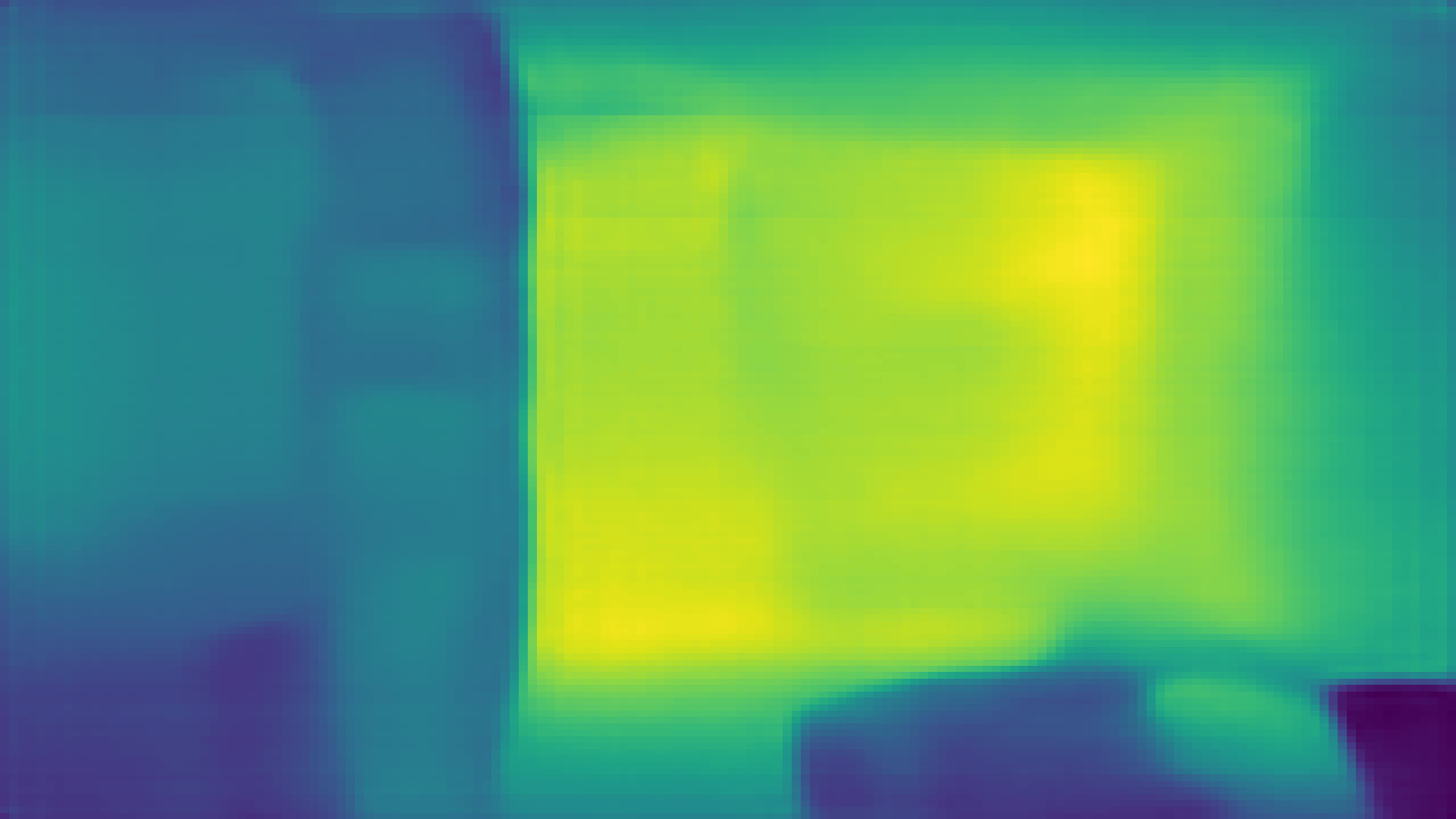}
    \includegraphics[width=0.24\linewidth]{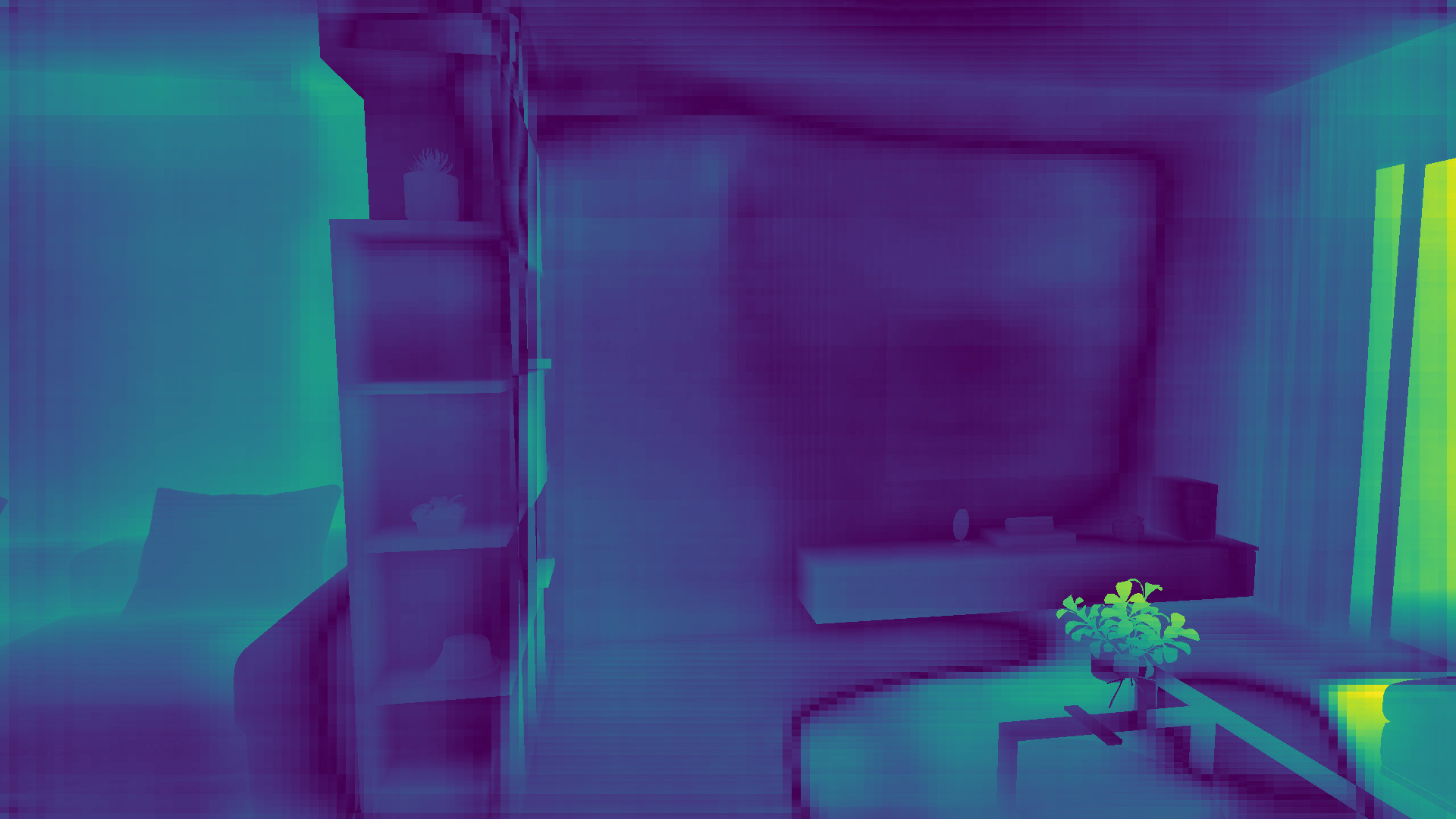}\\
    \smallskip
    \includegraphics[width=0.24\linewidth]{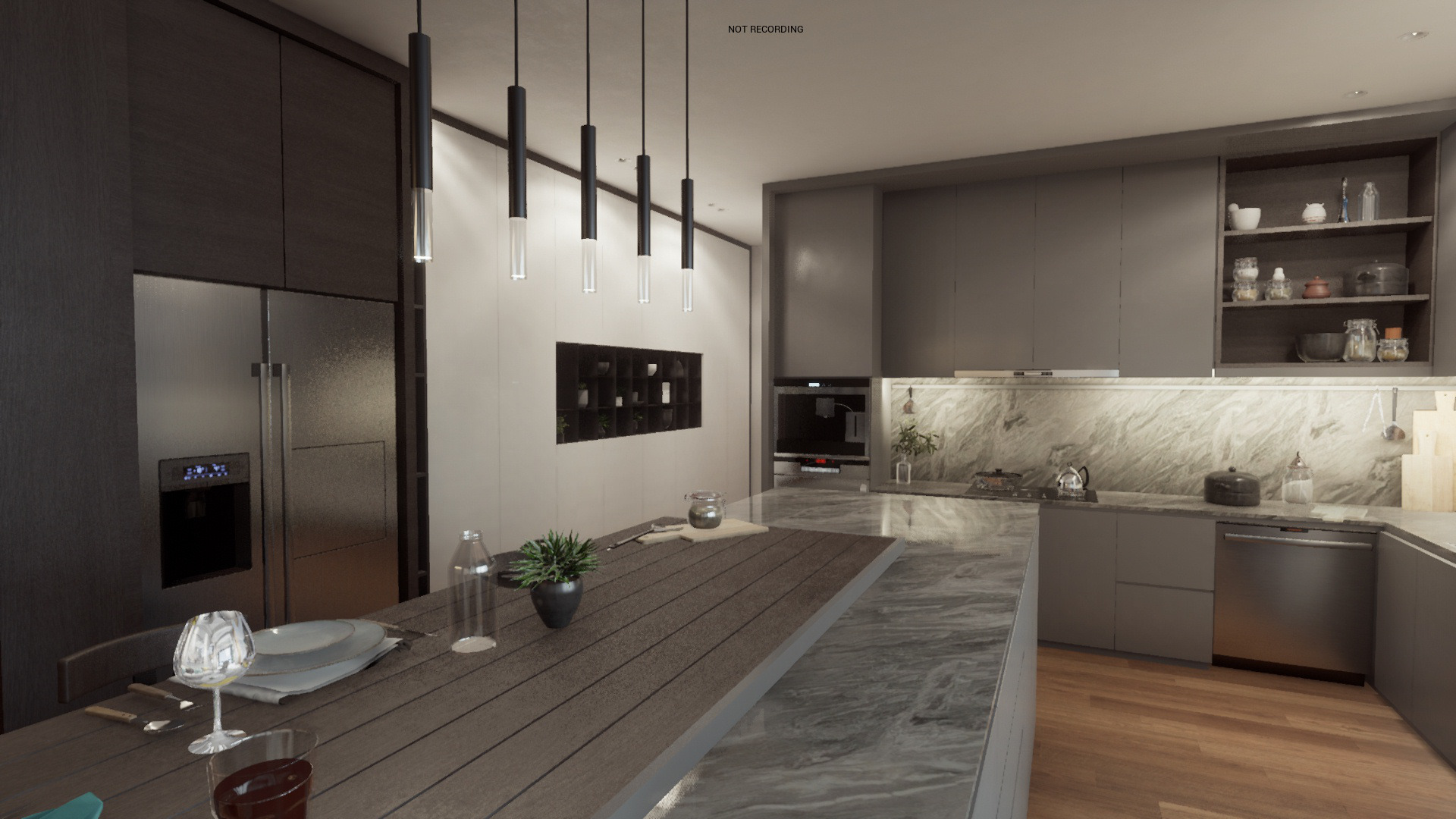}
    \includegraphics[width=0.24\linewidth]{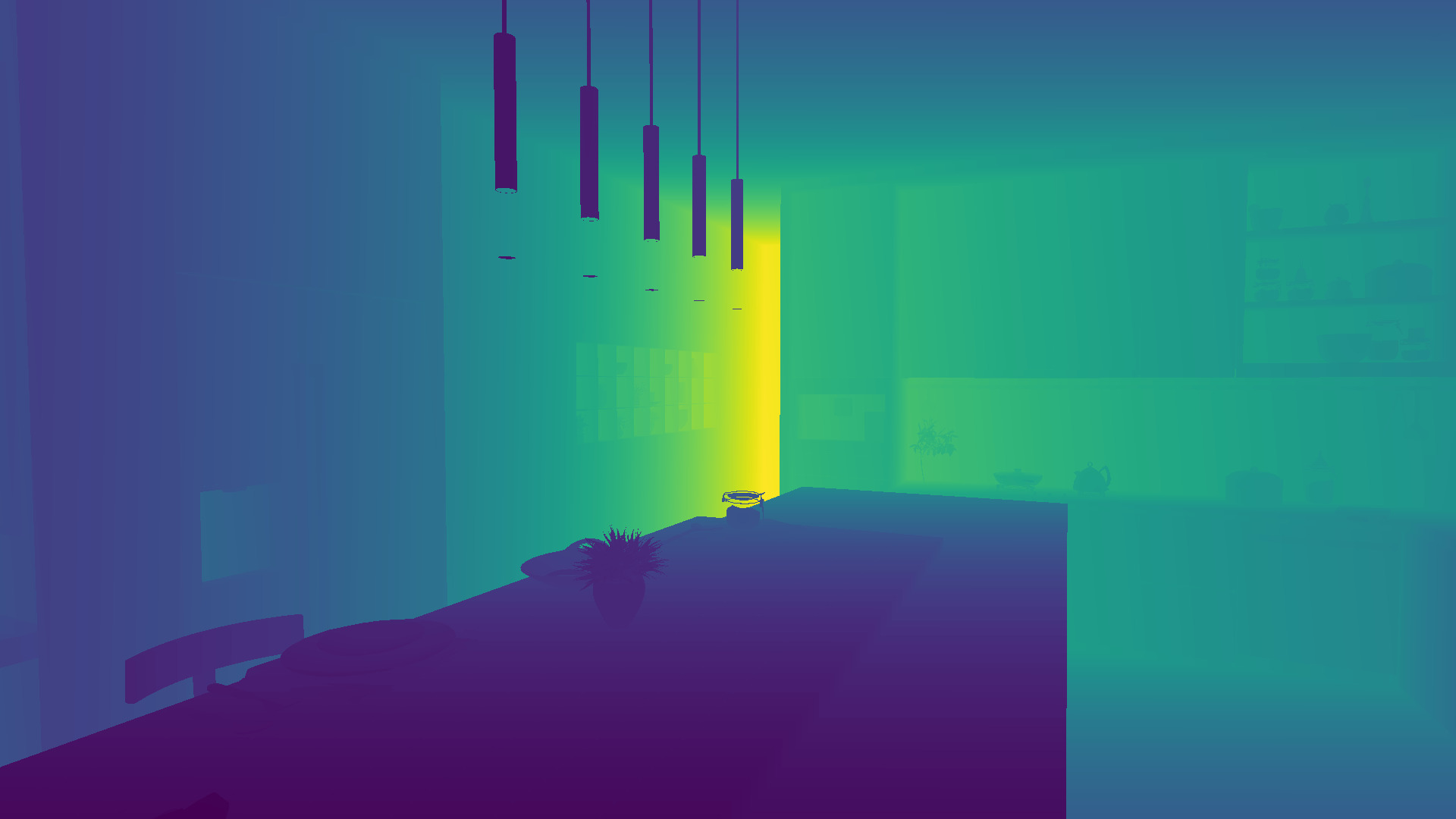}
    \includegraphics[width=0.24\linewidth]{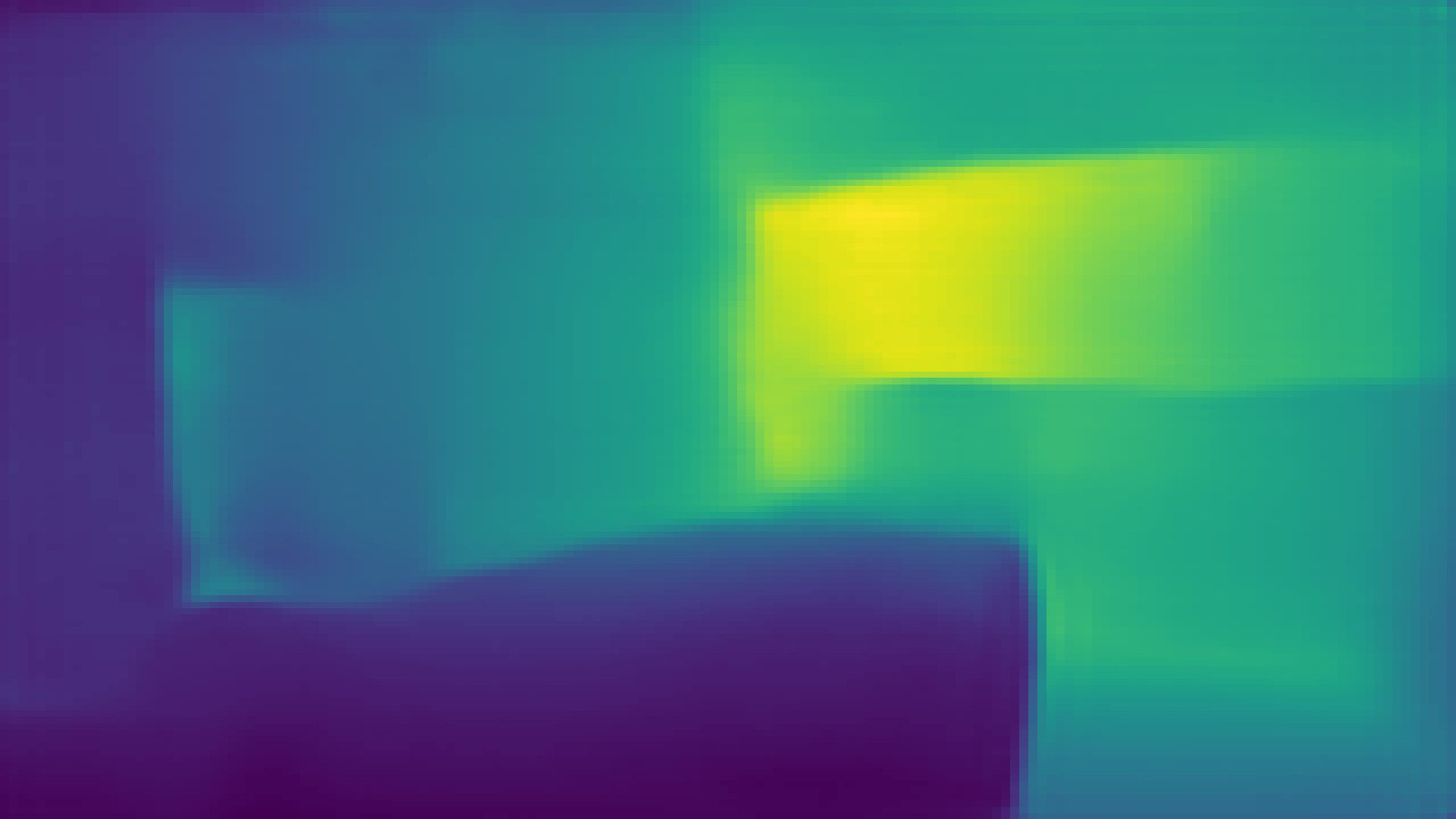}
    \includegraphics[width=0.24\linewidth]{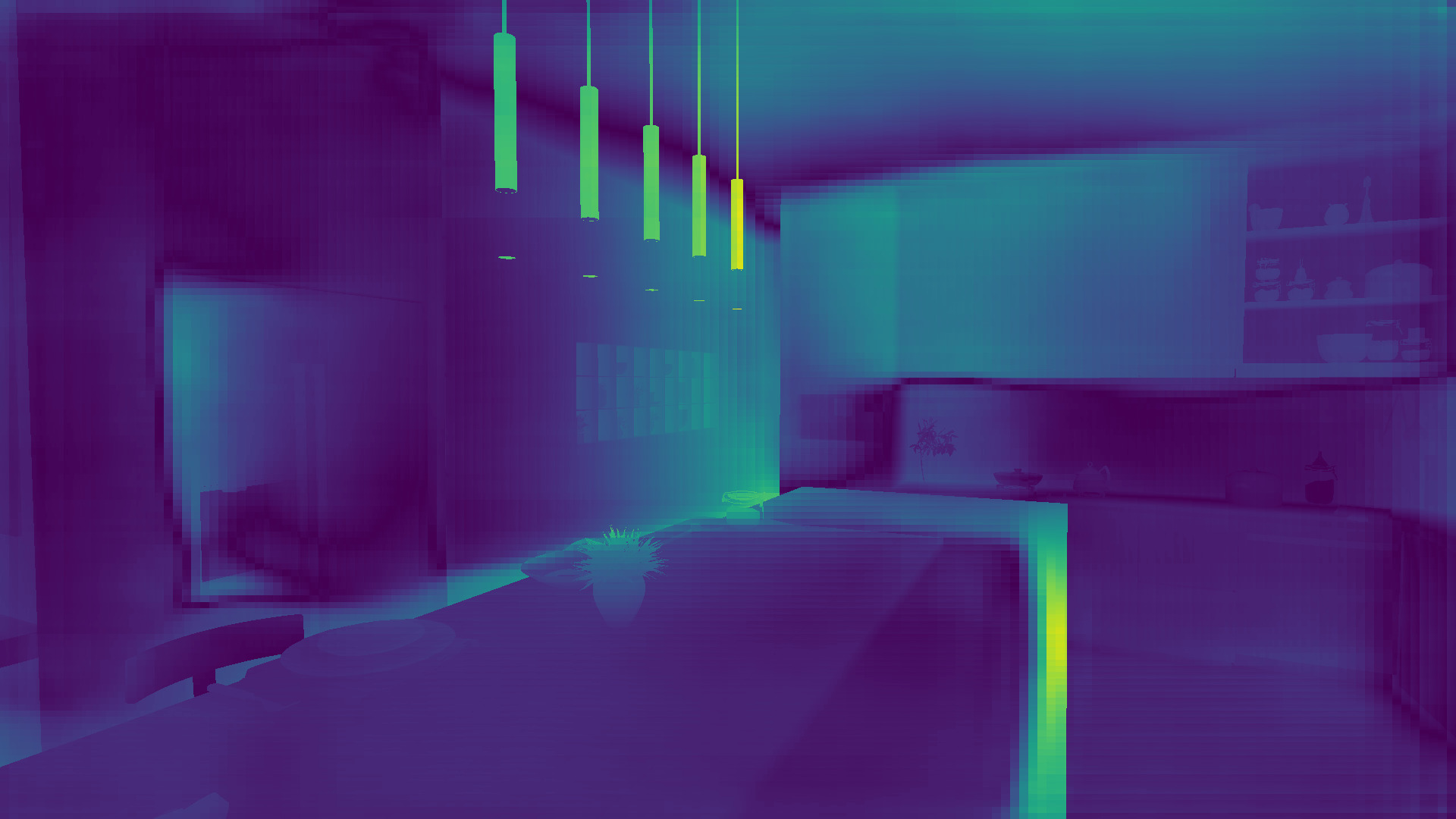}\\
    \smallskip
    \includegraphics[width=0.24\linewidth]{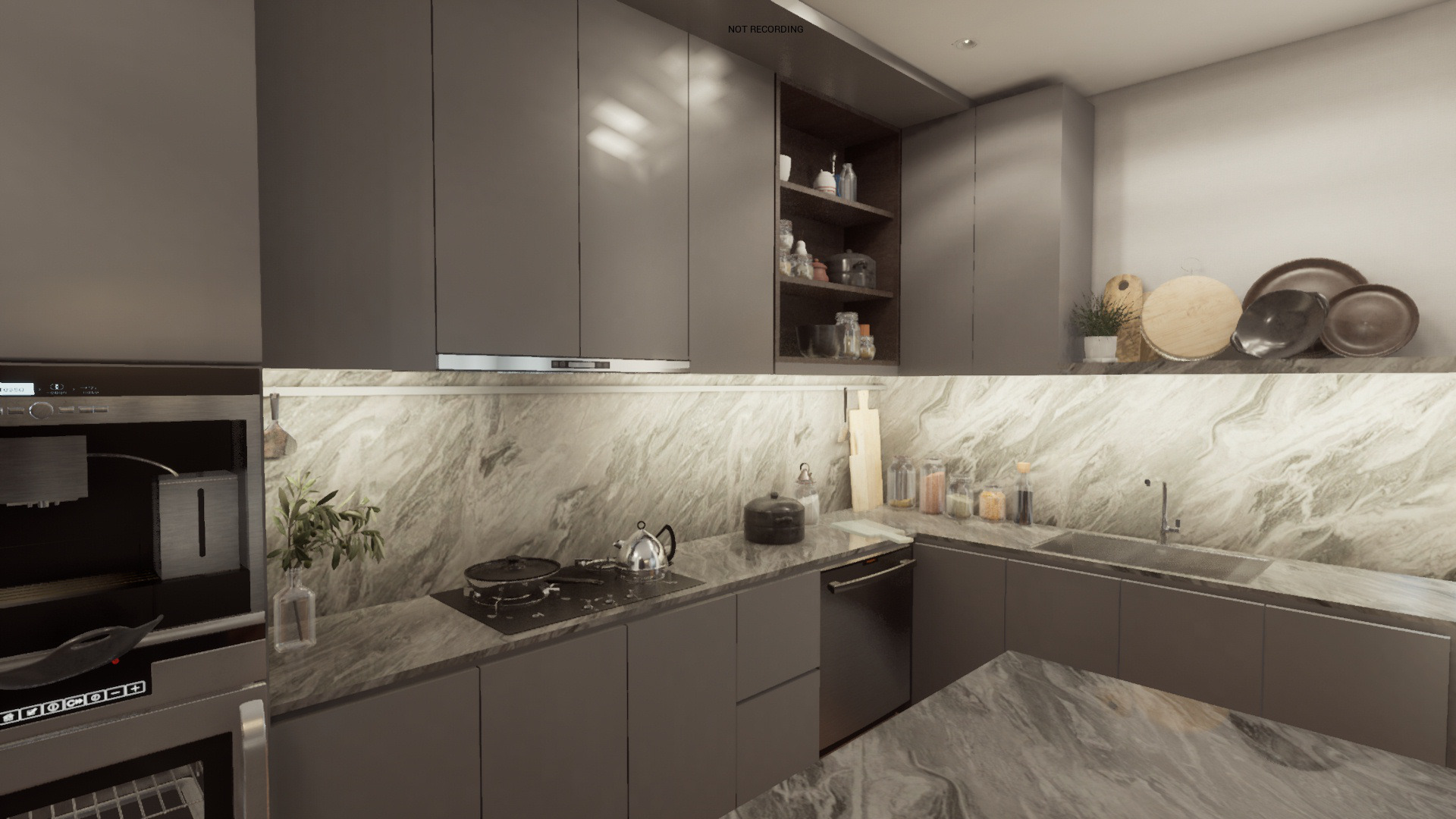}
    \includegraphics[width=0.24\linewidth]{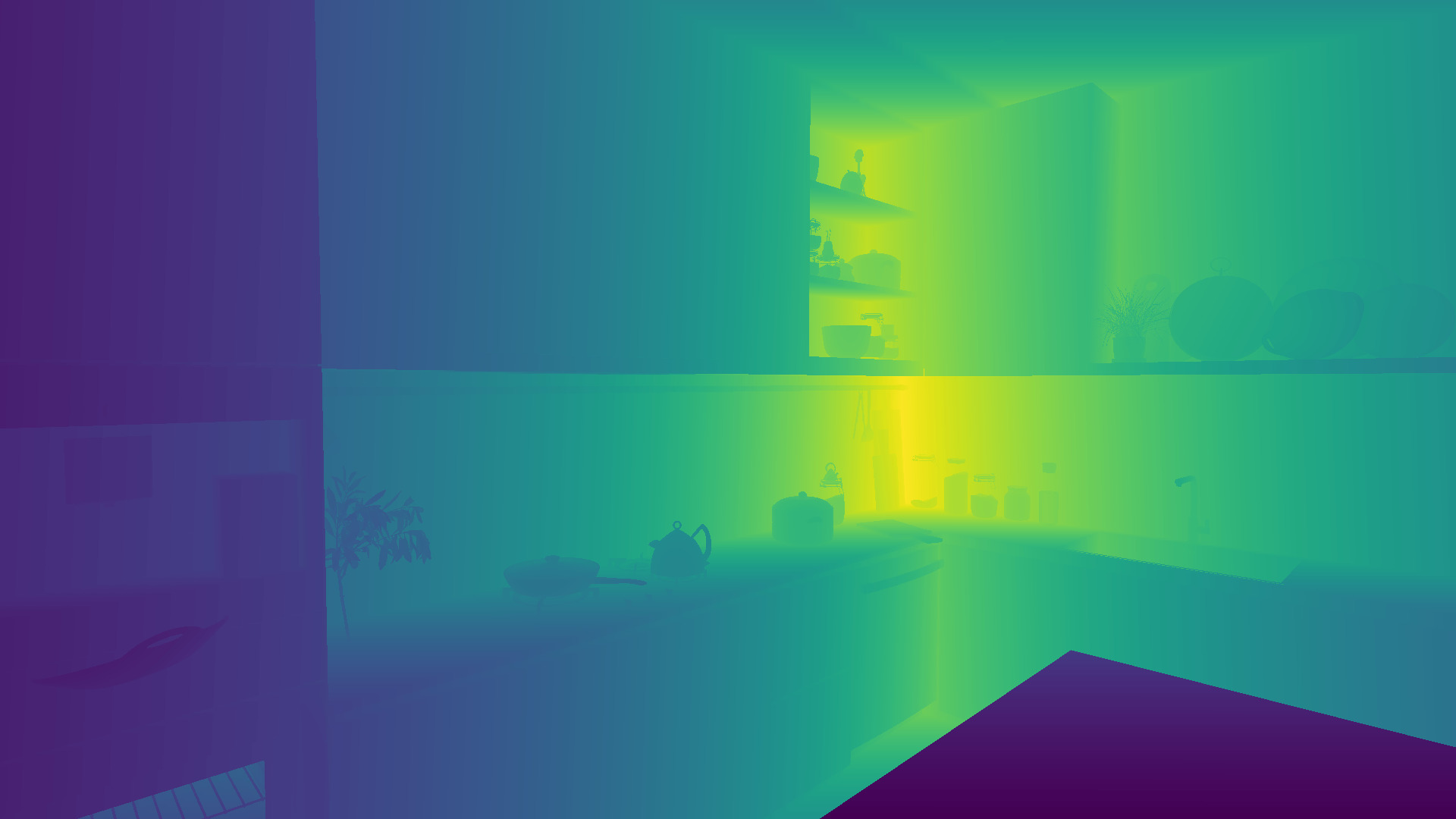}
    \includegraphics[width=0.24\linewidth]{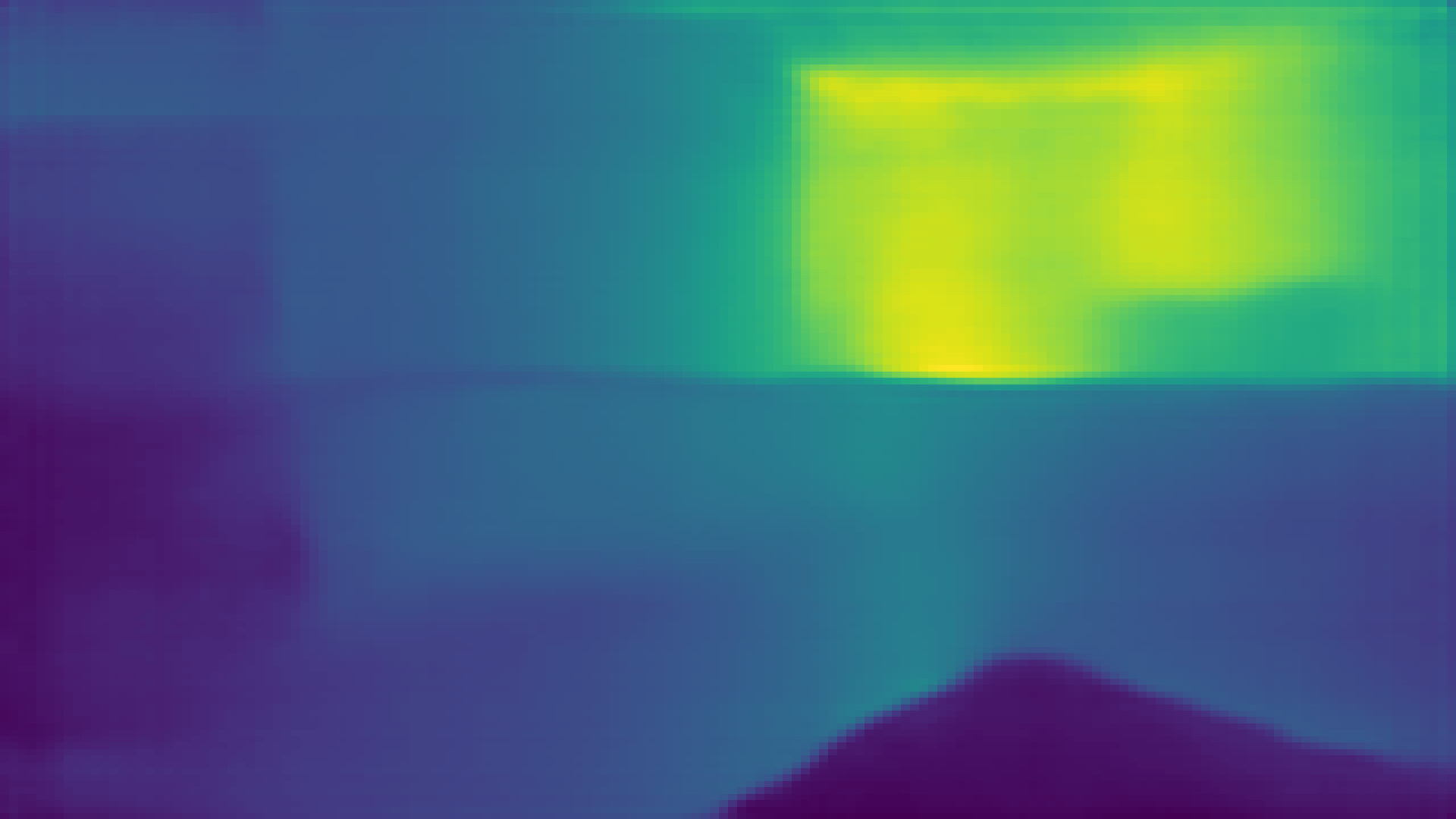}
    \includegraphics[width=0.24\linewidth]{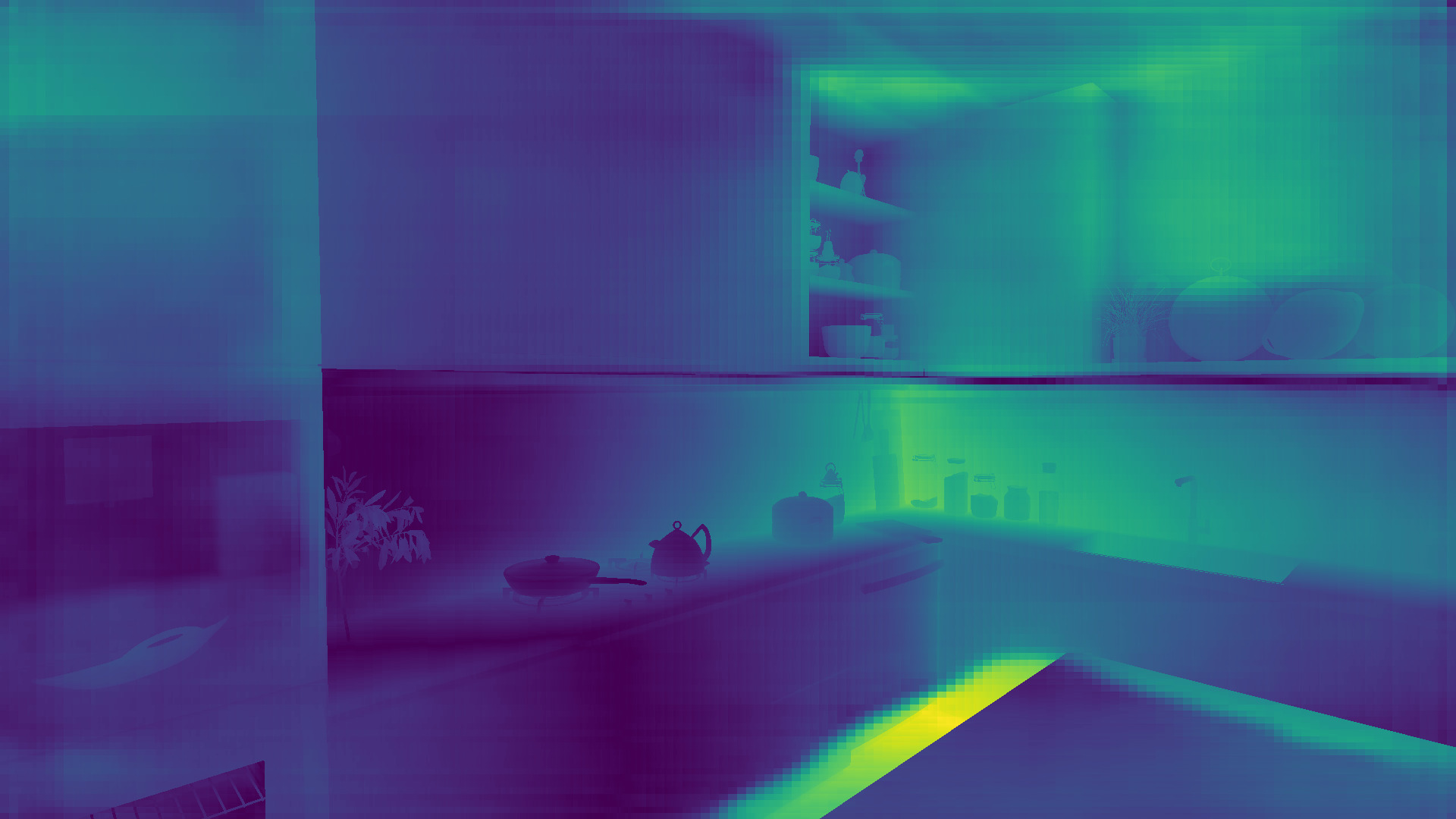}\\
    \smallskip
    \includegraphics[width=0.24\linewidth]{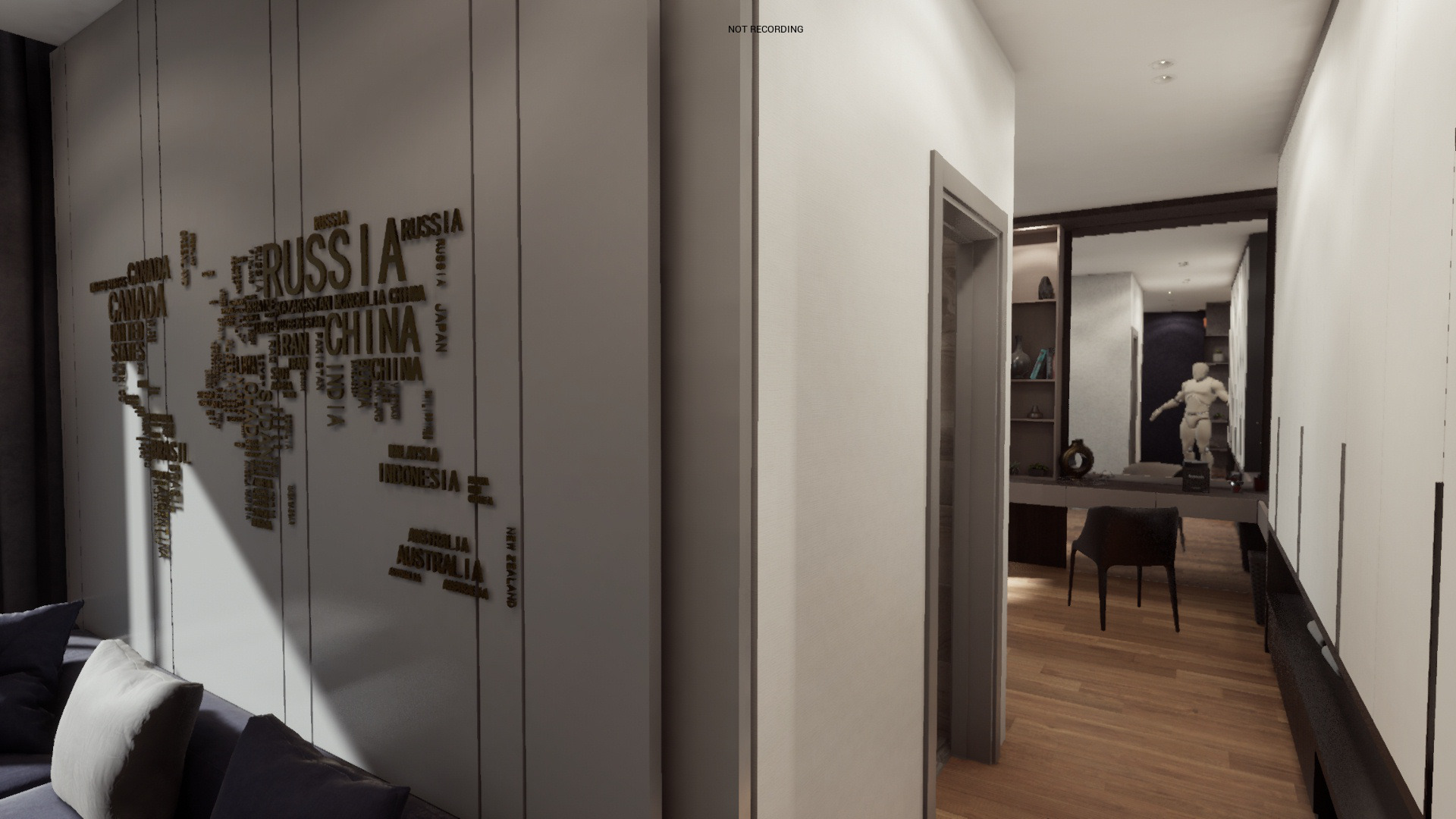}
    \includegraphics[width=0.24\linewidth]{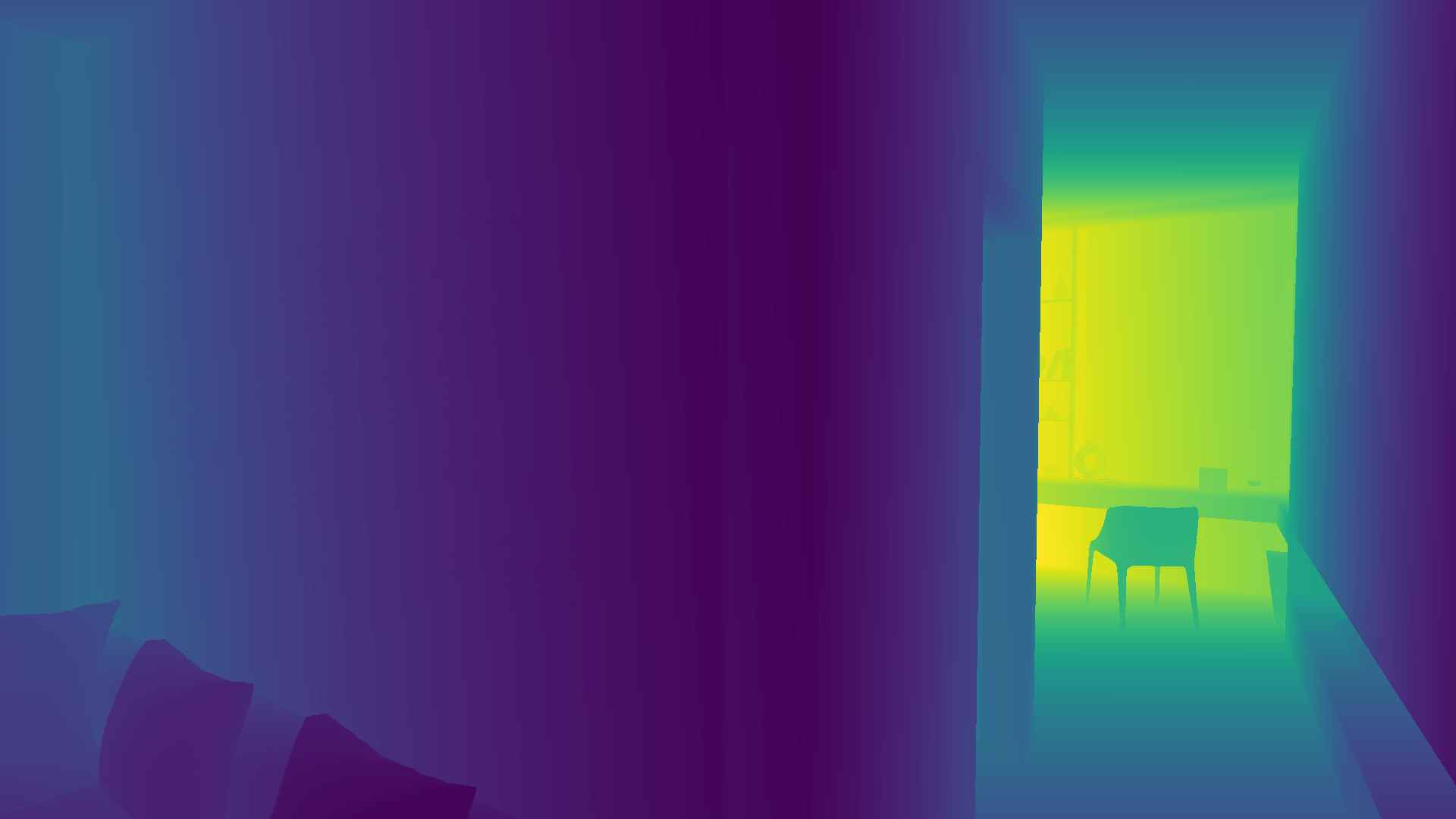}
    \includegraphics[width=0.24\linewidth]{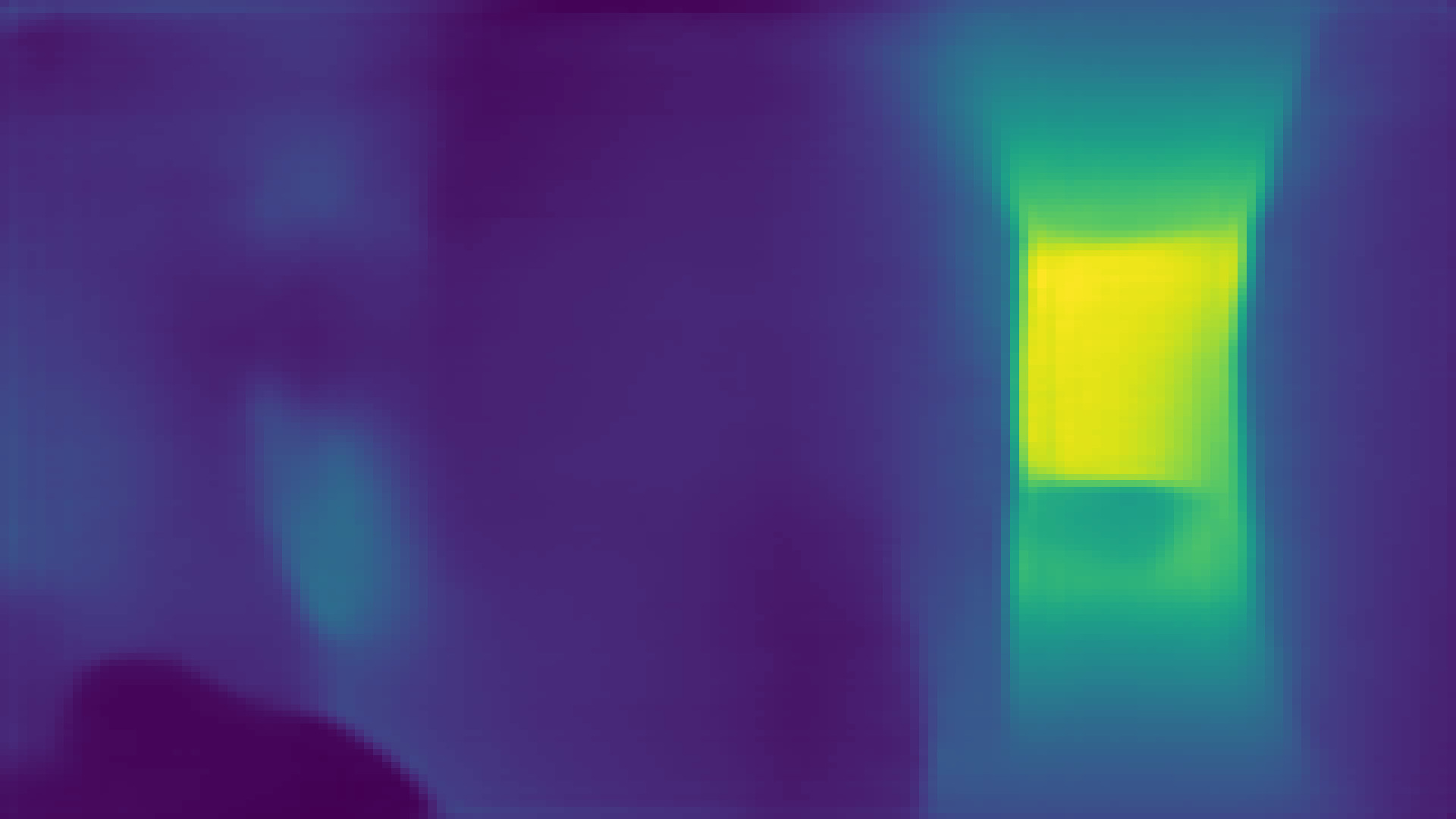}
    \includegraphics[width=0.24\linewidth]{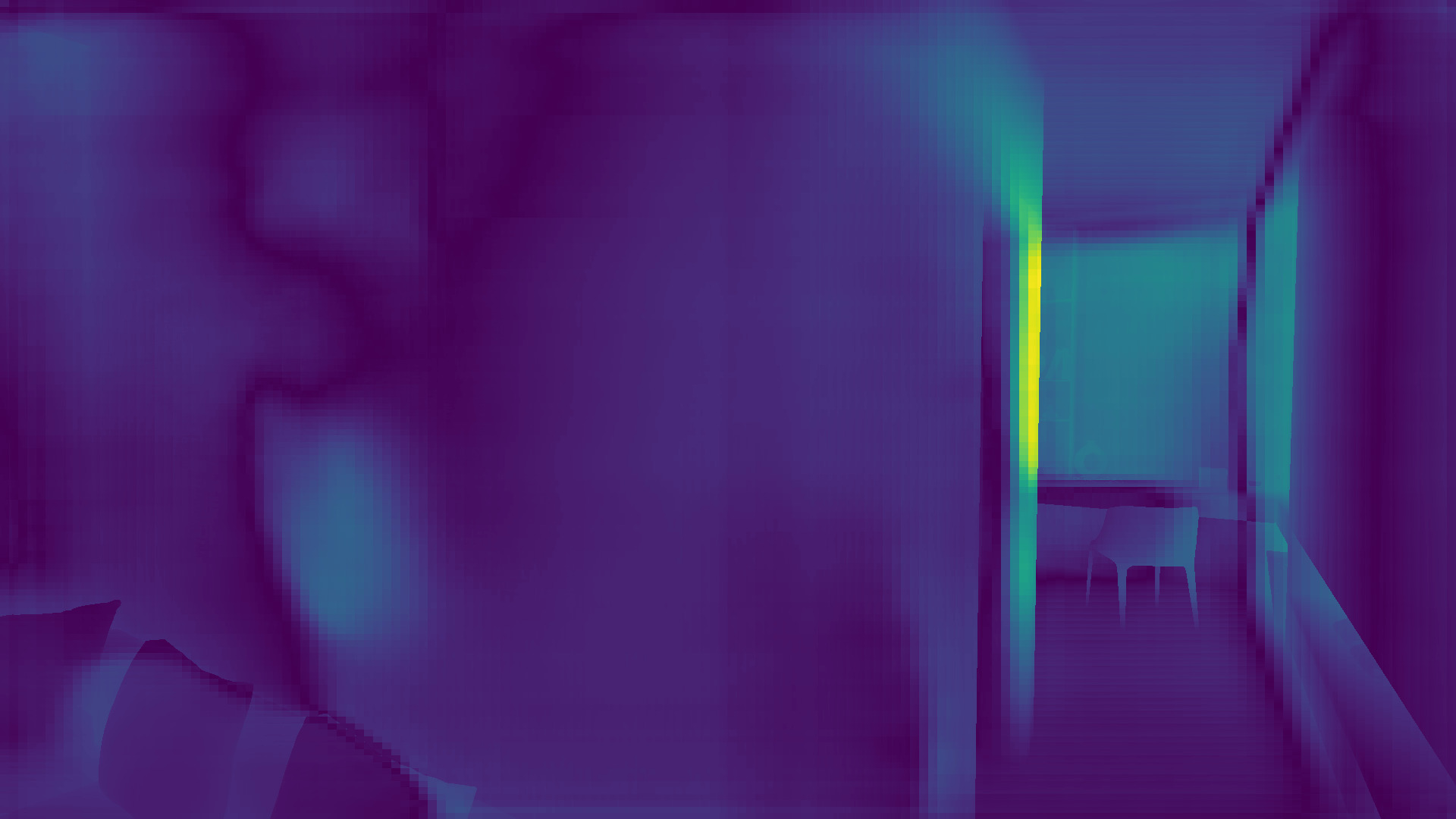}\\
    \includegraphics[width=0.24\linewidth]{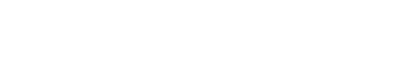}
    \includegraphics[width=0.24\linewidth]{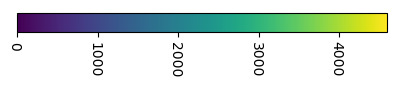}
    \includegraphics[width=0.24\linewidth]{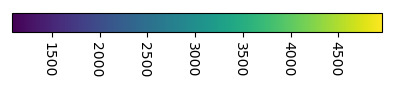}
    \includegraphics[width=0.24\linewidth]{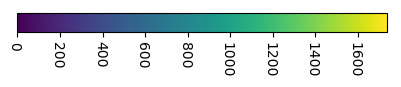}
    \caption{Qualitative visualization of Fully Convolutional Residual Networks for monocular depth estimation on data generated by our simulator. First column shows the RGB images, second column is the depth ground truth, third column shows the corresponding depth predictions, and the last column is the error map between the predicted and the ground truth depth.}
    \label{fig:laina_unrealrox}
\end{figure}

\begin{figure}[!htb]
    \centering
    \includegraphics[width=0.24\linewidth]{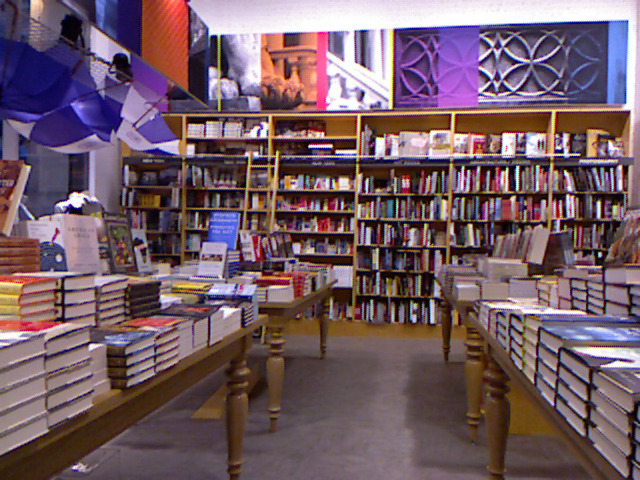}
    \includegraphics[width=0.24\linewidth]{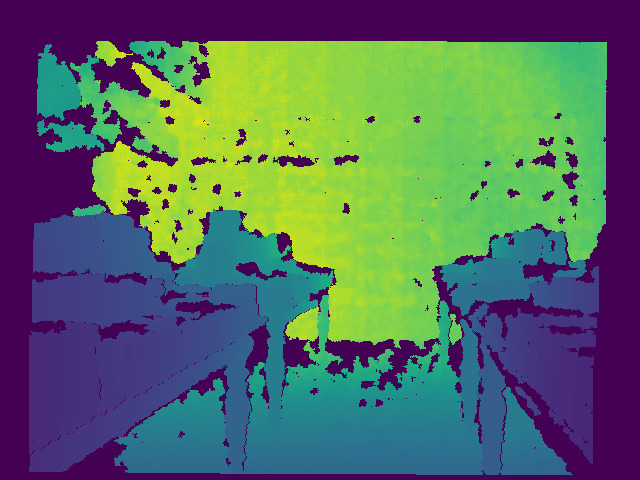}
    \includegraphics[width=0.24\linewidth]{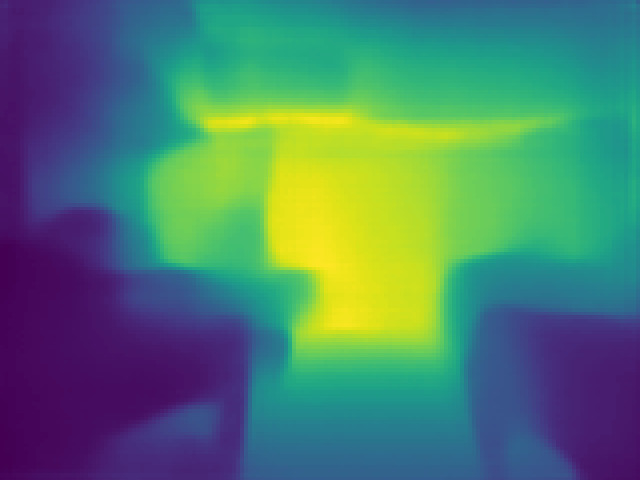}
    \includegraphics[width=0.24\linewidth]{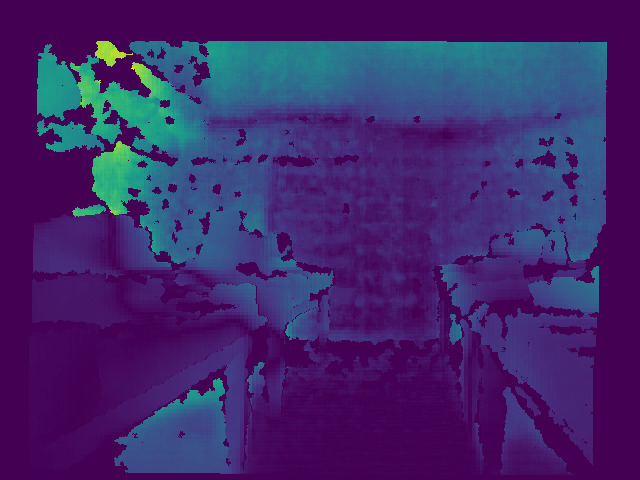}\\
    \includegraphics[width=0.24\linewidth]{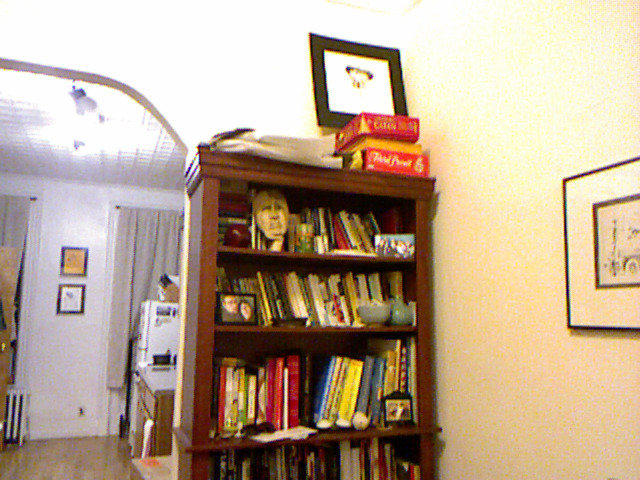}
    \includegraphics[width=0.24\linewidth]{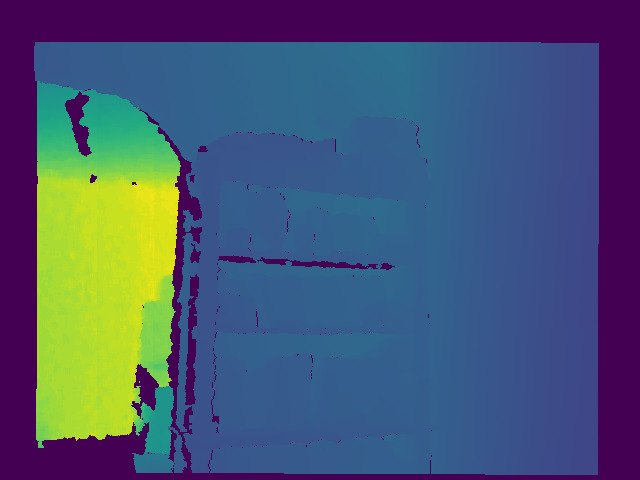}
    \includegraphics[width=0.24\linewidth]{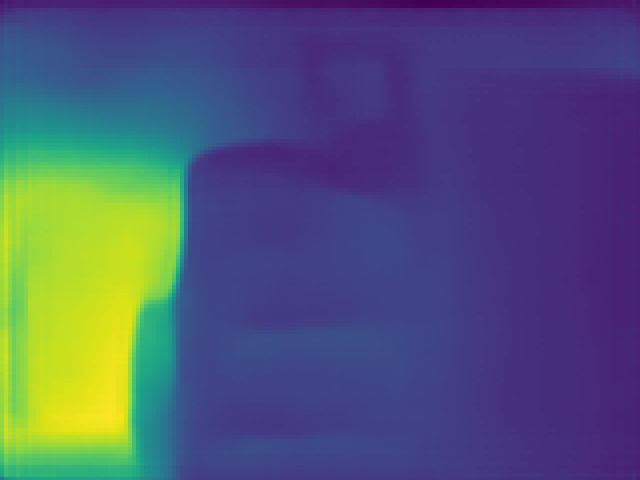}
    \includegraphics[width=0.24\linewidth]{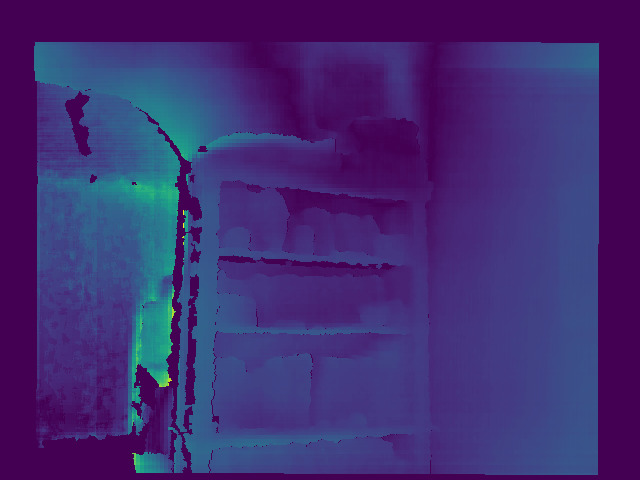}\\
    \includegraphics[width=0.24\linewidth]{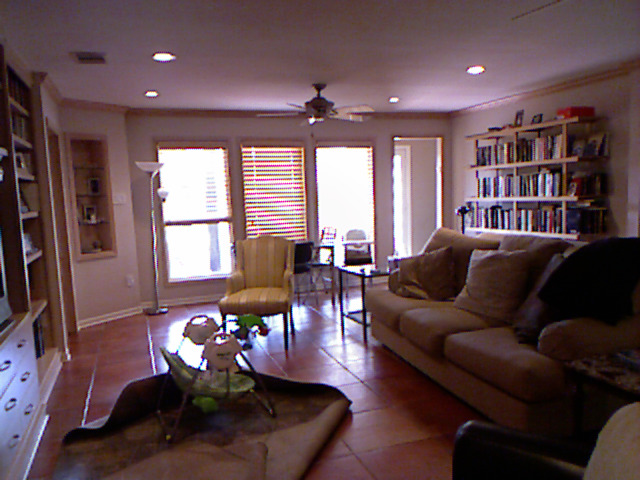}
    \includegraphics[width=0.24\linewidth]{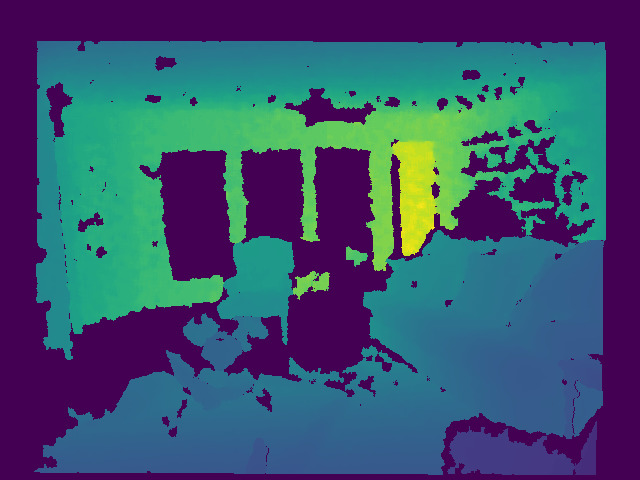}
    \includegraphics[width=0.24\linewidth]{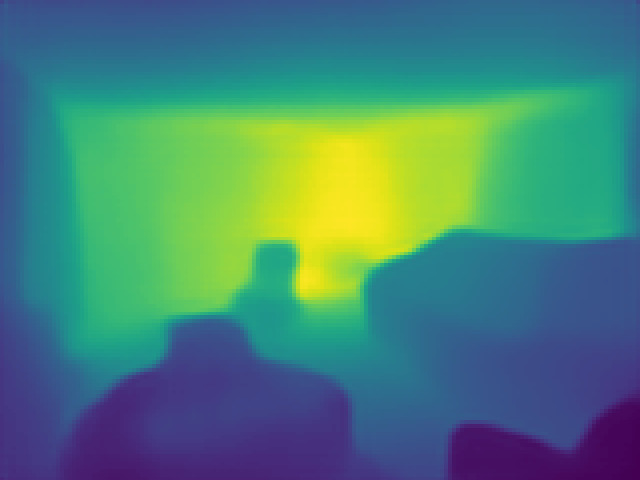}
    \includegraphics[width=0.24\linewidth]{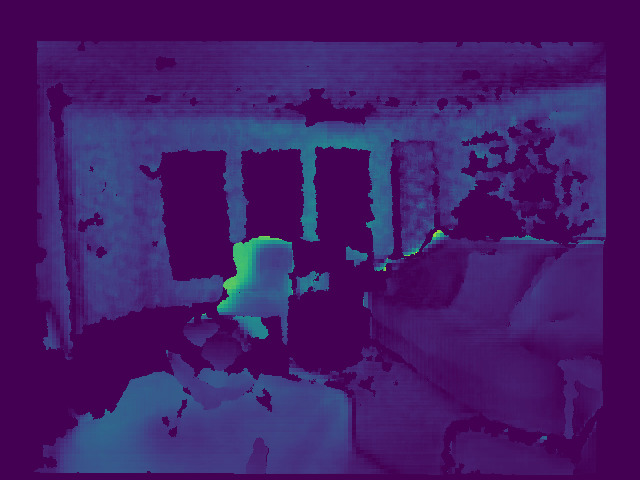}\\
    \includegraphics[width=0.24\linewidth]{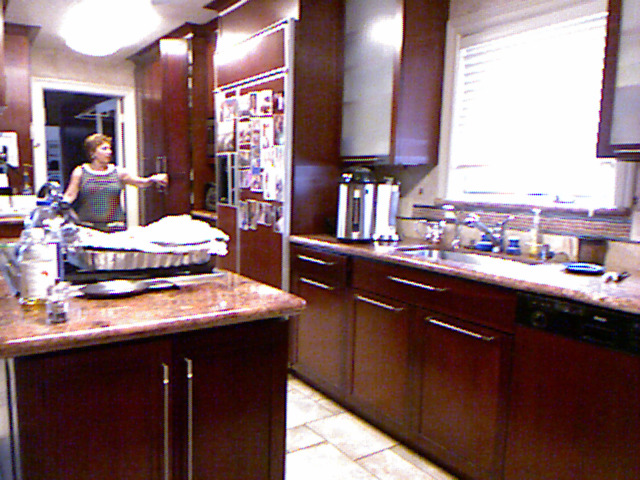}
    \includegraphics[width=0.24\linewidth]{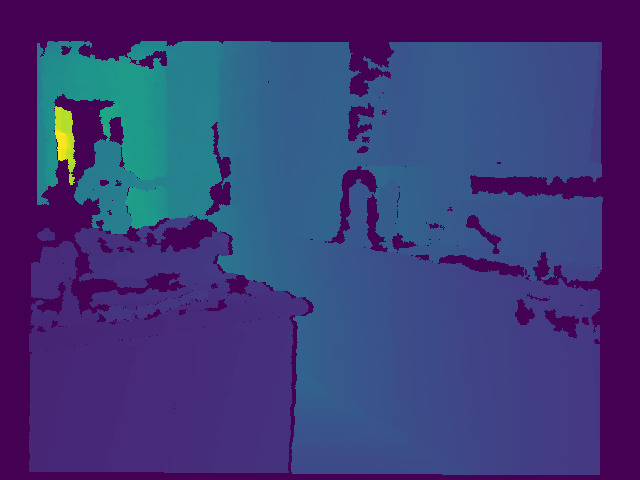}
    \includegraphics[width=0.24\linewidth]{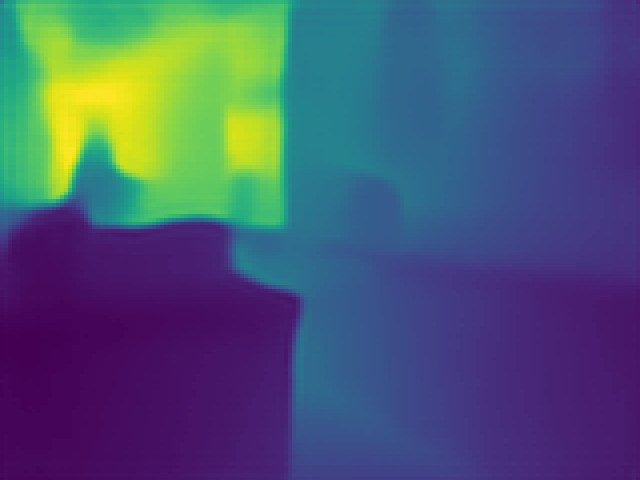}
    \includegraphics[width=0.24\linewidth]{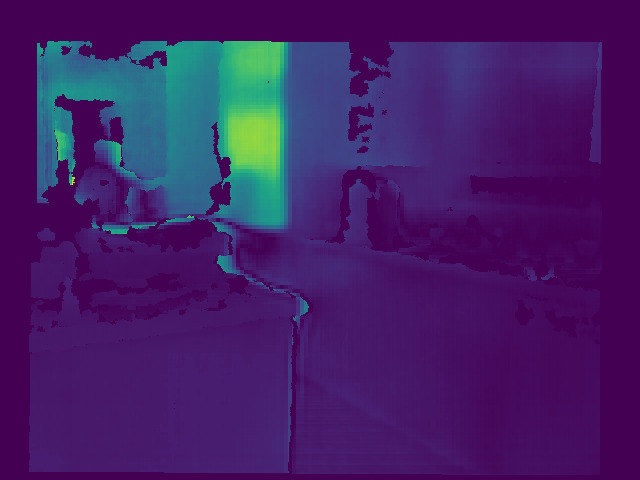}\\
    \includegraphics[width=0.24\linewidth]{color_bar}
    \includegraphics[width=0.24\linewidth]{gt_depth_bar}
    \includegraphics[width=0.24\linewidth]{prediction_bar}
    \includegraphics[width=0.24\linewidth]{error_map_bar}
    \caption{Qualitative evaluation of Fully Convolutional Residual Networks for monocular depth estimation on test data coming from NYUDv2 dataset \cite{Silberman2012}. First column shows the RGB images, second column is the depth ground truth, third column shows the corresponding depth predictions, and the last column is the error map between the predicted and the ground truth depth.}
    \label{fig:laina_nyu}
\end{figure}

The current trend for monocular depth estimation takes advantage of deep architectures, more concretely deep \acp{CNN}, with or without additional post-processing techniques for further refinement \cite{Eigen2014} \cite{Eigen2015} \cite{Xu2018}. Arguably, one of the most successful architecture is the Fully Convolutional Residual Network proposed by Laina \emph{et al.} \cite{Laina2016}. To prove the usefulness of our simulator, we have trained Laina's method using a set of samples coming from our simulator (Figure \ref{fig:laina_unrealrox} shows a random subset of the training images) and then we have tested it on a real-world dataset such as NYUDv2 \cite{Silberman2012} (Figure \ref{fig:laina_nyu} shows a random subset of testing samples for qualitative visualization).

As shown in this qualitative evaluation, knowledge learned using our simulated data can be seamlessly transferred to real-world data with adequate results in terms of accuracy and mean error.

\subsection{6D Object Pose Estimation}

Another widely used technique for which data generated with our tool can be helpful is 6D pose estimation of objects from 2D RGB images. This approach takes the object location problem one step further since it infers 3D rotation of the detected objects besides its location in an image (traditionally represented with a 2D bounding box). As a result, this estimation gives back a 3D bounding box that will estimate both 3D location (centroid) and rotation of the object.

This estimation was usually done through multi-stage algorithms that generated a coarse initial estimation that needed to be refined later. However, newer approaches like the one from Tekin \emph{et al.} \cite{Tekin2018} generate fine estimations which are accurate enough withoutrequiring multiples stages thus making it possible to perform 6D object pose estimation in real time. It is inspired by the YOLO network for semantic segmentation \cite{Redmon2016} \cite{Redmon2017} to estimate projected 3D bounding boxes that, later, will be converted to a 6D pose by leveraging PnP algorithms.

\begin{figure}[!htb]
	\centering
	\includegraphics[width=0.32\linewidth]{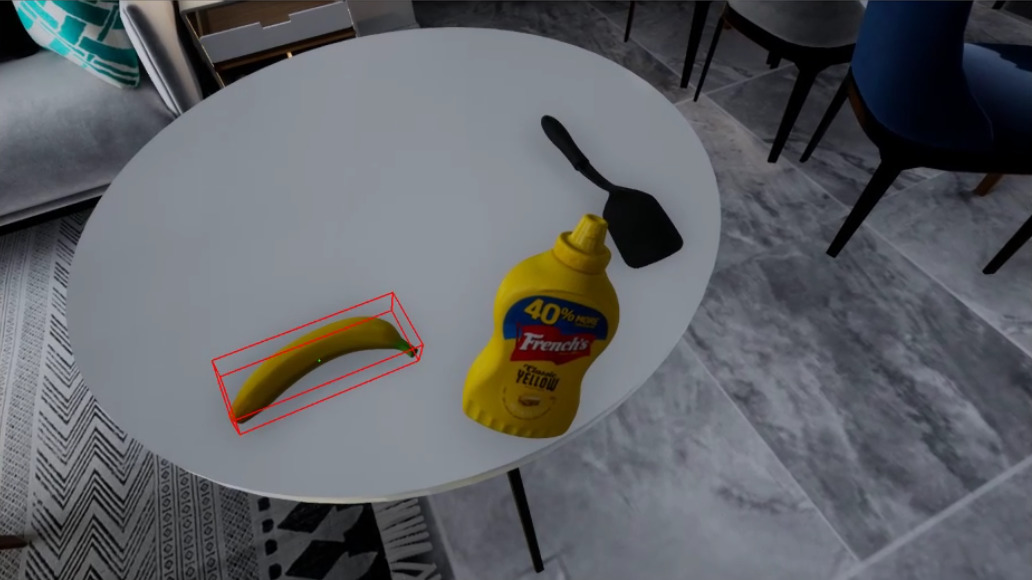}
	\includegraphics[width=0.32\linewidth]{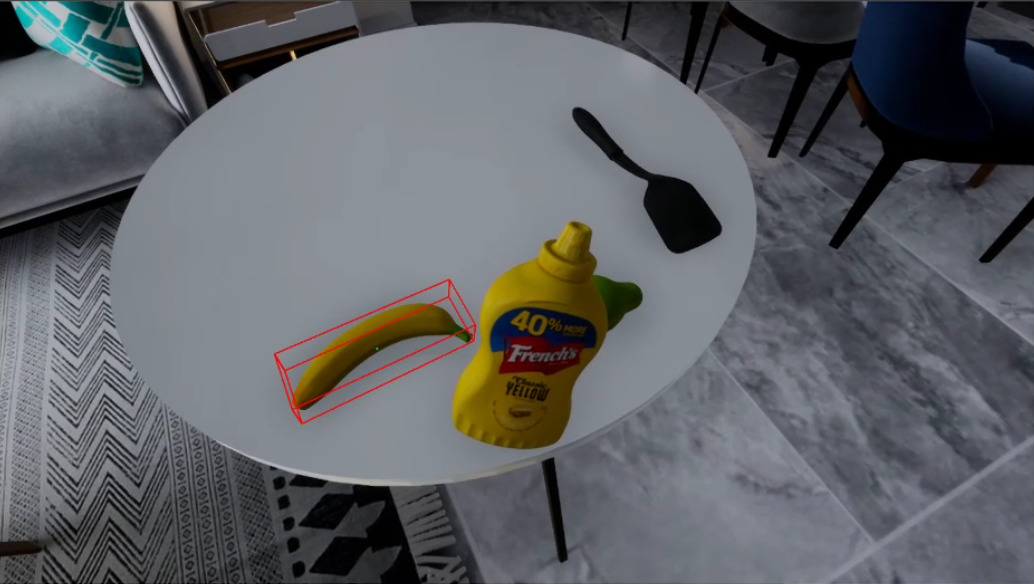}
	\includegraphics[width=0.32\linewidth]{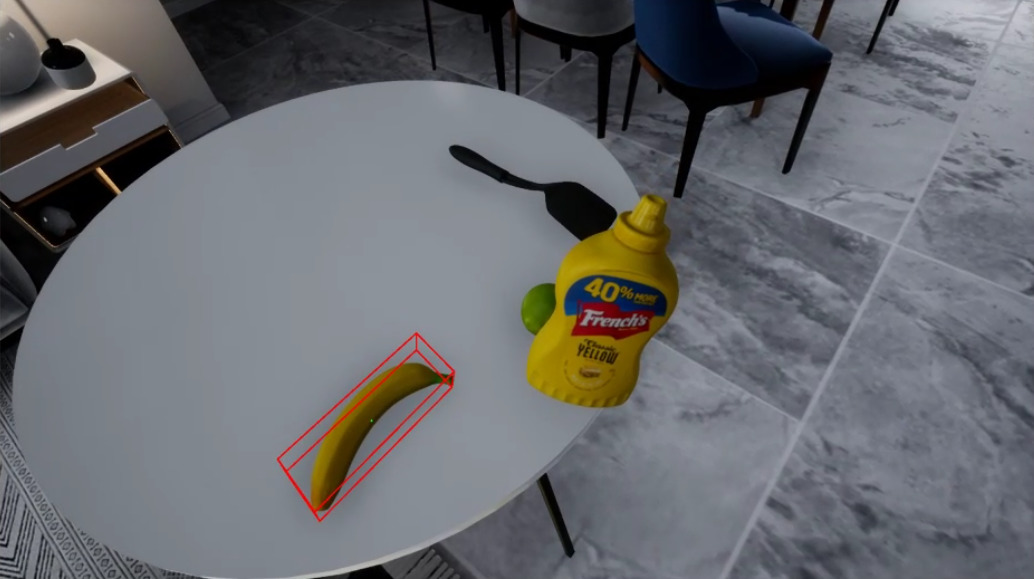}
	\caption{Qualitative evaluation of 6D pose estimation over synthetic data with single shot 6D object pose network \cite{Tekin2018} trained with synthetic data from our simulator.}
	\label{fig:6dpose_synth}
\end{figure}

To prove the usefulness of our generator in this problem, the network by Tekin \emph{et al.} has been trained with our simulated data, and then tested with both synthetic and real images in order to see if it can transfer the knowledge to real-world data. First of all, in Figure \ref{fig:6dpose_synth} we can observe the pose estimation (through its 3D bounding box) of a banana in a sequence of synthetic images from our simulator.

\begin{figure}[!htb]
	\centering
	\includegraphics[width=0.40\linewidth]{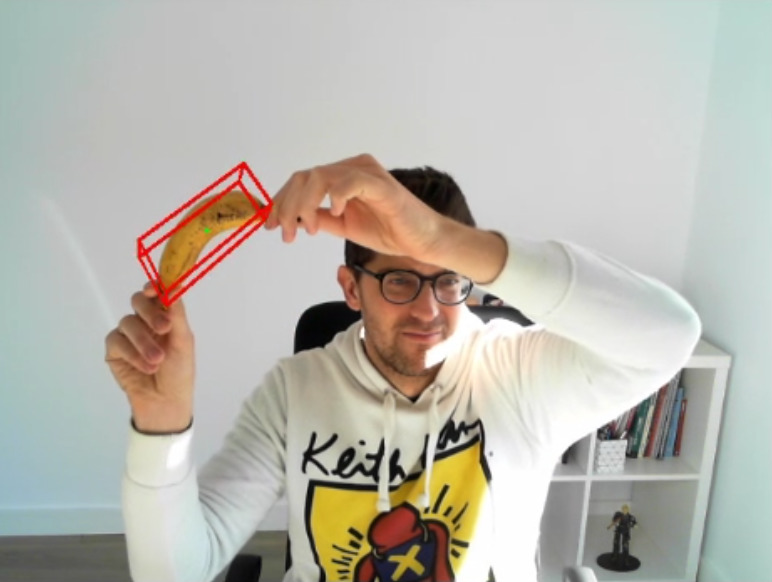}
	\includegraphics[width=0.40\linewidth]{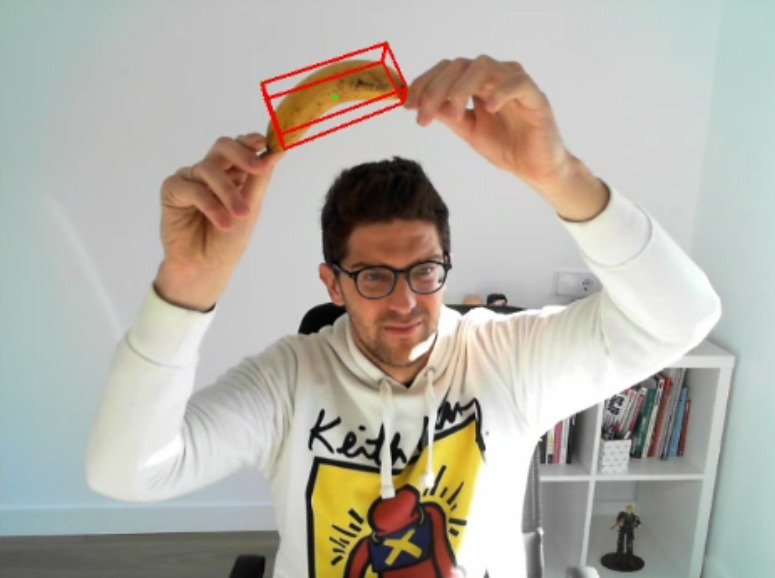}\\
	\includegraphics[width=0.40\linewidth]{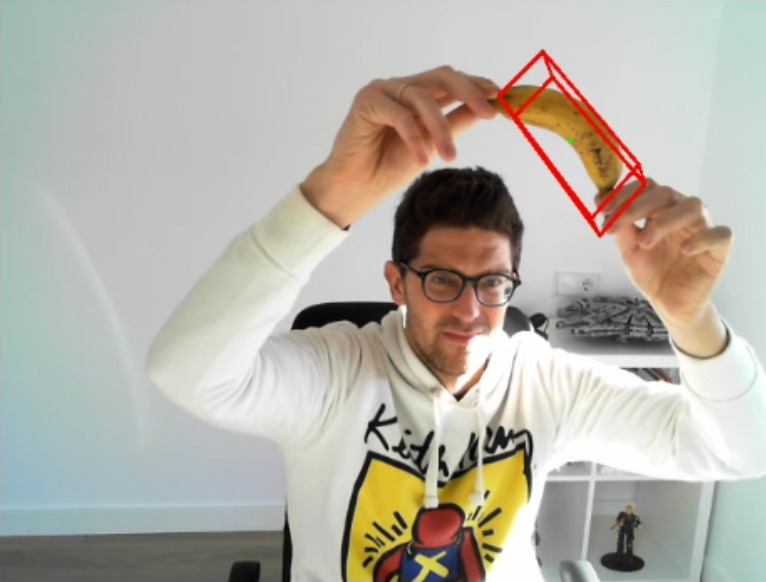}
	\includegraphics[width=0.40\linewidth]{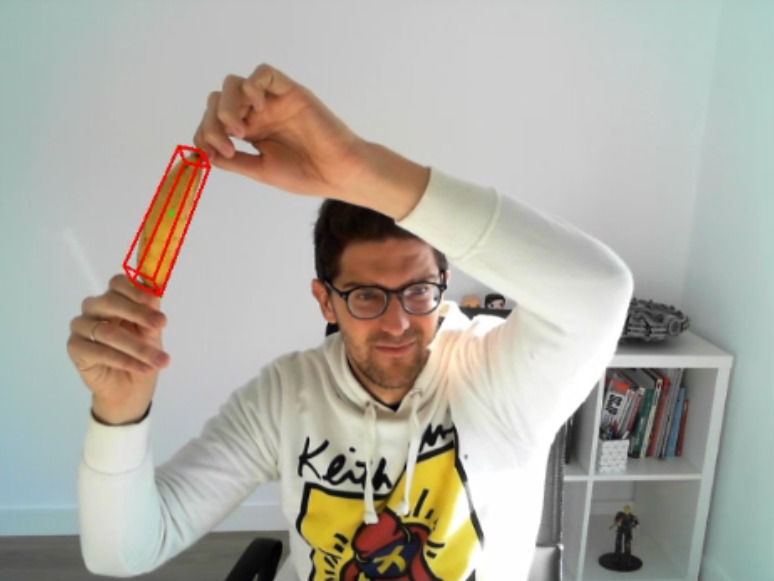}
	\caption{Qualitative evaluation of 6D pose estimation over real data with single shot 6D object pose network \cite{Tekin2018} trained with synthetic data from our simulator.}
	\label{fig:6dpose_real}
\end{figure}

Later, Figure \ref{fig:6dpose_real} shows the same previously trained network trying to estimate the pose of a real banana on a live sequence captured by a camera.

As shown in this evaluation, it follows the same trend as the depth estimation experiments: knowledge learned from our synthetic data generator can be transferred to real-world data with success.

\section{Conclusion}
\label{sec:conclusion}

This paper presented a virtual reality system, in which a human operator is embodied as a robotic agent using \ac{VR} setups such as Oculus Rift or HTC Vive Pro, for generating automatically annotated synthetic data for various robotic vision tasks. This environment leverages photorealism for bridging the reality gap so that models trained on its simulated data can be transferred to a real-world domain while still generalizing properly. The whole project, with all the aforementioned components (recording/playback, multi-camera, HUD, controller, and robotic pawns) is freely available \footnote{\url{https://github.com/3dperceptionlab/unrealrox}} with an open-source license and detailed documentation so that any researcher can use to generate custom data or even extend it to suit their particular needs. That data generation process was engineered and designed with efficiency and easiness in mind and it outperforms other existing solutions such as UnrealCV at object, robot, and camera repositioning, and image generation.

The outcome of this work demonstrates the potential of using \ac{VR} for simulating robotic interactions and generating synthetic data that facilitates training data-driven methods for various applications such as semantic segmentation, depth estimation, or object recognition.

\section{Limitations and Future Works}
\label{sec:limitations}

Currently, the environment still has certain limitations that must be addressed in order to make it applicable to a wider range of robotic vision tasks. One of them is the simulation of non-rigid objects and deformations when grasping such kind of objects. We have limited ourselves to manipulate non-deformable objects in order not to affect realism, since this is a different approach with a non-haptic manipulation and deformations need to be modelled at the object level. We are currently investigating the mechanisms that \ac{UE4} offers to model those transformations. Another important shortcoming is the absence of tactile information when grasping objects. We plan to include simulated tactile sensors to provide force data when fingers collide with objects and grasp them instead of providing only visual information. Furthermore, although not strictly a limitation, we are working on making the system able to process \acp{URDF} to automatically import robots, including their constraints, cinematics, and colliders, in the environment instead of doing that manually for each robot model.

\ifCLASSOPTIONcompsoc
  \section*{Acknowledgments}
\else
  \section*{Acknowledgment}
\fi

This work has been funded by the Spanish Government TIN2016-76515-R grant for the COMBAHO project, supported with Feder funds. This work has also been supported by three Spanish national grants for PhD studies (FPU15/04516, FPU17/00166, and ACIF/2018/197), by the University of Alicante project GRE16-19, and by the Valencian Government project GV/2018/022.
Experiments were made possible by a generous hardware donation from NVIDIA.
We would also like to thank Zuria Bauer for her collaboration in the depth estimation experiments.

\bibliographystyle{plain}
\bibliography{bare_jrnl_compsoc}   % name your BibTeX data base

\ifCLASSOPTIONcaptionsoff
  \newpage
\fi

\end{document}